\newif\ifdraft
\newif\ifanon
\drafttrue
\anonfalse

\documentclass[11pt]{article}

\ifanon
\usepackage[review]{acl}
\else
\usepackage{acl}
\fi
\usepackage{times}
\usepackage{latexsym}
\usepackage[T1]{fontenc}
\usepackage[utf8]{inputenc}
\usepackage{microtype}
\usepackage{inconsolata}

\usepackage{hyperref}
\usepackage{url}
\usepackage{booktabs}
\usepackage{graphicx}
\usepackage{makecell}
\usepackage{listings}

\usepackage{cleveref}
\usepackage{subcaption}

\usepackage{courier}

\usepackage{paralist}
\usepackage{enumitem}

\usepackage[normalem]{ulem}

\usepackage{tikz}
\usetikzlibrary{tikzmark}
\usetikzlibrary{arrows}
\usetikzmarklibrary{listings}
\usetikzlibrary{positioning}

\usepackage{wrapfig}

\usepackage{siunitx}
\newtheorem{definition}{Definition}

\usepackage{subfiles}

\Crefname{section}{\S}{\S\S}
\crefformat{section}{\S#2#1#3}
\Crefname{figure}{Figure}{Figures}
\crefformat{figure}{Figure #2#1#3}
\Crefname{Figure}{Figure}{Figures}
\Crefname{Table}{Table}{Tables}
\crefformat{table}{Table #2#1#3}

\ifdraft
\usepackage{background}
\backgroundsetup{contents={}}
\fi

\lstdefinelanguage{python}{
  sensitive=false,
  morekeywords={if,def,return,class,self},
  keywordstyle=\ttfamily\bfseries,
  identifierstyle=\ttfamily,
  comment=[l]{//},
  commentstyle=\ttfamily,
  string=[d]{"},
  stringstyle=\ttfamily,
  mathescape=true,
  literate={STAR}{{$\star$}}1,
  escapechar=\#,
  extendedchars=true,
basicstyle=\footnotesize\ttfamily,
  showstringspaces=false,
  numbers=none,
  firstnumber=0,
  numberstyle=\small,
  stepnumber=5,
  numbersep=5pt,
  upquote=true,
  columns=fixed,
  texcl=false,
  flexiblecolumns=true}

\lstdefinelanguage{TypeScript}{
  sensitive=false,
  morekeywords={if,function,return},
  keywordstyle=\ttfamily\bfseries,
  identifierstyle=\ttfamily,
  comment=[l]{//},
  commentstyle=\ttfamily,
  string=[d]{"},
  stringstyle=\ttfamily,
  mathescape=true,
  literate={STAR}{{$\star$}}1,
  escapechar=\#,
  extendedchars=true,
  basicstyle=\footnotesize\ttfamily,
  showstringspaces=false,
  numbers=none,
  firstnumber=0,
  numberstyle=\small,
  stepnumber=5,
  numbersep=5pt,
  upquote=true,
  columns=fixed,
  texcl=false,
  flexiblecolumns=true}

\title{Understanding How CodeLLMs (Mis)Predict Types with \\ Activation Steering}

\author{Francesca Lucchetti \& Arjun Guha \\
Khoury College of Computer Sciences \\
Northeastern University \\
Boston, MA 02115 \\
\texttt{\{lucchetti.f,a.guha\}@northeastern.edu}
}

\begin{document}

\maketitle

\begin{abstract}







Large Language Models (LLMs) are widely used by software engineers for programming tasks. However, research shows that LLMs often lack a deep understanding of program semantics. Even minor changes to syntax, such as renaming variables, can significantly degrade performance across various tasks. In this work, we examine the task of \emph{type prediction}: given a partially typed program, can a model predict a missing type annotations such that the resulting program is more typed? We construct a dataset of adversarial examples where models initially predict the correct types, but begin to fail after semantically irrelevant edits. This is problematic, as models should ideally generalize across different syntactic forms of semantically equivalent code. This lack of robustness suggests that models may have a shallow understanding of code semantics.

Despite this, we provide evidence that LLMs do, in fact, learn robust mechanisms for type prediction—though these mechanisms often fail to activate in adversarial scenarios. By using \emph{activation steering}, a method that manipulates a model’s internal activations to guide it toward using latent knowledge, we restore accurate predictions on adversarial inputs. We show that steering successfully activates a type prediction mechanism that is shared by both Python and TypeScript, and is more effective than prompting with in-context examples. Across five different models, our comprehensive evaluation demonstrates that LLMs can learn generalizable representations of code semantics that transfer across programming languages.

\end{abstract}

\section{Introduction}
\label{intro}

Large Language Models (LLMs) are widely used by software engineers on many programming tasks. Despite their impressive capabilities, research has shown that they are not robust to semantically irrelevant features of programs: syntactic changes such as reordering conditions or renaming variables can significantly impact LLM performance on programming tasks~\citep{pmlr-v235-hooda24a}. This raises a fundamental question: do contemporary LLMs learn to reason about program semantics, or do they merely learn textual features such as the associations between variable names and their types? For example, predicting that a variable named $n$  has type \emph{int}, regardless of how it is used.

Reasoning about programs involves a number of different tasks~\citep{pmlr-v235-gu24c}. In this paper, we focus on the \emph{type prediction} task for gradually typed programming languages, specifically Python and TypeScript, defined as follows.
\begin{definition}[Type Prediction]
Given a partially typed program $p$, choose an untyped variable binding $\mathit{var} \in p$, predict a type annotation $\mathit{var} : T$, and insert the annotation back into the program to get a new program $p'$ that also passes the type-checker. 
\end{definition}

Types are fundamental to programming languages. Reliably predicting types requires understanding control flow and data flow in a program, and gradual type prediction is particularly challenging. Unlike type inference (e.g., in Haskell or OCaml), where classical algorithms work, gradual type prediction is undecidable~\citep{migeed:decidable}. Moreover, it is always possible to predict the $\mathit{any}$ type, which is imprecise, but sometimes necessary in very dynamic code. The challenge is to predict a type that is both precise and consistent with program semantics~\citep{phipps-costin:typewhich}, and classical algorithms so far do no scale to modern programming languages (\cref{background-and-related-work}). 


LLMs are remarkably good at type prediction for Python and TypeScript~\citep{yee:typeweaver,fried:incoder}. However, as we show in this paper, when a model successfully predicts the type $T$ of a variable $\mathit{var} \in p^{+}$, we can often construct a variation $p^{-}$ with minimal syntactic changes that make the model mispredict the type. The question we ask is, \emph{why do these type mispredictions occur?}


In this paper, we give evidence that models learn a robust internal mechanism for type prediction in hidden layers $\ell$. However, this mechanism can fail to activate when the input program $p^-$ has adversarial syntactic features that mislead prediction (e.g. unreliable variable names). We show that we can correct such mispredictions by editing model layers $\ell$ with targeted \emph{steering vectors} $\mathbf{v}^{\ell}$. This allows us to demonstrate that:
\begin{enumerate}[itemsep=0pt,parsep=0pt]

\item Adding $\mathbf{v}^{\ell}$ to layers $\ell$ activates the mechanism and significantly improves type prediction performance (\cref{section:steering-all-models});

\item $\mathbf{v}^{\ell}$ is shared across languages; we can improve Python type prediction with $\mathbf{v}^{\ell}$ computed from TypeScript and vice versa (\cref{section:lang-transfer}); and

\item $\mathbf{v}^{\ell}$ enables \emph{prediction} but does not control \emph{precision} of types. In other words, when a model predicts a type such as \emph{any}, $\mathbf{v}^{\ell}$ does not make the prediction more precise (\cref{section:error-analysis}).

\end{enumerate}
We also show that this internal type prediction mechanism is hard to access without directly adding $\mathbf{v}^{\ell}$ to the model. Specifically, in-context learning has a negligible impact on accuracy of type prediction for problems where a direct model edit is successful (\cref{section:steering-baselines}).

Our extensive evaluation shows that results generalize across five different LLMs from four model families~\citep{hui2024qwen2,qwen2,dubey2024llama,roziere2023code,li:starcoder}. These include both pretrained and instruction-tuned LLMs, LLMs trained exclusively on code, and general-purpose LLMs trained on code and data.

\begin{figure}
\begin{subfigure}{\columnwidth}
\begin{lstlisting}[language=python]
def is_palindrome(s: #\colorbox{yellow}{[FILL]}#):
    s = s.lower()
    return s[::-1] == s
\end{lstlisting}
\caption{The abstract type prediction task.}
\label{task-example}
\end{subfigure}

\begin{subfigure}{\columnwidth}
\begin{lstlisting}[language=python]
<fim_prefix>
def is_palindrome(s: <fim_suffix>):
    s = s.lower()
    return s[::-1] == s<fim_middle>
\end{lstlisting}
\caption{A fill-in-the-middle prompt for the task.}
\label{fim-example}
\end{subfigure}

\begin{subfigure}{\columnwidth}
\begin{lstlisting}[language=python]
[USER] Continue this program with the
correct substitution for <FILL>:
def is_palindrome(s: <FILL>):
    s = s.lower()
    return s[::-1]==s
[ASSISTANT] def is_palindrome(s:
\end{lstlisting}
\caption{A prompt for an instruction-tuned model.}
\label{fim-chat-example}
\end{subfigure}

\caption{An example type prediction task, formulated for each type of model.}

\end{figure}

\section{Background and Related Work}
\label{background-and-related-work}

\paragraph{Classical type prediction and type inference}

Type prediction is distinct from type inference as found in languages such as OCaml and Haskell. In those languages, every variable is typed, even if the types are implicit~\citep{harper:xml}. In contrast, a gradually typed programming language allows programs to freely mix typed and untyped code, giving programmers more flexibility than traditional static typing affords~\citep{siek:gtlc,th:migration}. However, untyped code still needs to type-correct for the program to run correctly. With omitted or weak type annotations, type errors may not be caught until program execution.

There is prior work on rule-based type prediction algorithms~\citep{phipps-costin:typewhich,rastogi:gti,siek:gti,campora:migrating,henglein:scheme-to-ml,wright:soft-typing}. But, these papers present algorithms for variations of the lambda calculus or simple functional languages such as Scheme, and have not been scaled to more complex, modern programming languages.

\paragraph{Neural type prediction}

Over the past decade, prior work has explored leveraging neural networks, including LLMs, for type prediction~\citep{hellendoorn:dlti,jesse:diversetyper,jesse:typebert,pandi_probabilistic_2021,wei:lambdanet}. Unlike classical approaches that target idealized programming languages, these works attempt to predict types for widely-used programming languages like TypeScript and Python. A practical approach to automated type prediction would be significant. Airbnb, Dropbox, Slack, Netflix, and many others have each taken several years to manually add type annotations to their multi-million line gradually typed codebases~\citep{airbnb:ts-migrate,dropbox:mypy,slack:ts,heap:ts,netflix:ts,abacus:ts,quip:ts,stripe:sorbet}.

\begin{figure*}[t]
\begin{subfigure}{0.48\textwidth}  
\begin{lstlisting}[language=Python]
class Point#\tikzmark{a-src}#:
  def __init__(self, x, y):
    self.x = x
    self.y = y

def delta_x(p: Point#\tikzmark{c-src}#, #\tikzmark{b-src}#x: #\underline{float}#):
  p.x = p.x + x
\end{lstlisting}
\caption{The original program.}
\label{fig:steering_pair}
\end{subfigure}
~\vrule~
\begin{subfigure}{0.48\textwidth}  
\begin{lstlisting}[language=Python]
class #\tikzmark{a-dst}#Type0:
  def __init__(self, x, y):
    self.x = x
    self.y = y

def delta_x(p: #\tikzmark{a-other}#Type0, x: #\underline{float}#):
  p.x = p.x + x
\end{lstlisting}
\caption{Type renaming.}
\end{subfigure}
\hrule
\begin{subfigure}{0.48\textwidth}  
\begin{lstlisting}[language=Python]
class Point:
  def __init__(self, x, y):
    self.x = x
    self.y = y

def delta_x(p: Point, #\tikzmark{b-dst}#tmp: #\underline{float}#):
  p.x = p.x + tmp#\tikzmark{b-other}#
\end{lstlisting}
\caption{Variable renaming.}
\label{var-rename-example}
\end{subfigure}
~\vrule~
\begin{subfigure}{0.48\textwidth}  
\begin{lstlisting}[language=Python]
class Point:
  def __init__(self, x, y):
    self.x = x
    self.y = y

def delta_x(p#\tikzmark{c-dst}#, x: #\underline{float}#):
  p.x = p.x + x
\end{lstlisting}
\caption{Type annotation removal.}
\end{subfigure}
\begin{tikzpicture}[remember picture,overlay]
\draw[gray,-latex]
  ([yshift=5,xshift=3]pic cs:a-src)
  to [bend left=10]
  ([yshift=5,xshift=0]pic cs:a-dst);
\draw[gray,-latex]
  ([yshift=-1,xshift=3]pic cs:b-src)
  to [bend left=60]
  ([yshift=5,xshift=15]pic cs:b-dst);
\draw[gray,-latex]
  ([yshift=0,xshift=28]pic cs:a-dst)
  to [bend left=10]
  ([yshift=6,xshift=15]pic cs:a-other);
\draw[gray,-latex]
  ([yshift=-1,xshift=10]pic cs:b-dst)
  to [bend left=5]
  ([yshift=0,xshift=0]pic cs:b-other);
\draw[gray,dotted,line width=0.5mm,-latex]
  ([yshift=-2,xshift=-6]pic cs:c-src)
  to [bend right=10]
  ([yshift=5,xshift=-2]pic cs:c-dst);
\end{tikzpicture}

\caption{Examples of three semantics-preserving edits. The type prediction site is \texttt{\underline{float}}. We ensure that each edit is internally consistent. E.g., in (\ref{var-rename-example}), when we rename the binding \texttt{x} to \texttt{tmp}, we rename references to the binding. }
\label{mutation-examples}
\end{figure*}

\begin{figure*}
\begin{lstlisting}[language=python]
class KafkaAvroBackend(RepositoryBackend):
    def __init__(
        self, #\sout{config}##\colorbox{yellow}{\_\_tmp0}#: #\underline{dict}#, producer=AvroProducer, loader=AvroMessageLoader,
        value_serializer: Callable = to_message_from_dto,
        get_producer_config: Callable = get_producer_config,
        get_loader_config: Callable = get_loader_config
    ) -> None:
        producer_config = get_producer_config(#\sout{config}##\colorbox{yellow}{\_\_tmp0}#)
\end{lstlisting}

\caption{A fragment of a Python steering pair. The original code is 70 lines of text. The \underline{dict} is the expected prediction. But, renaming \texttt{config} to \texttt{\_\_tmp0} makes the model mispredict \texttt{Repository}, which is a hallucination.}
\label{real-world-py-steering-pair}

\end{figure*}

\paragraph{Mutation testing and program transformations}

In our experiments, we construct type prediction prompts by renaming variables to arbitrary names, or deleting some type annotations in the context. We construct our edits such that they do not break program syntax, and all the information necessary for type prediction is still present in the program. To do so, we take inspiration from \emph{mutation testing}~\citep{demillo:test-data-selection}. The goal of mutation testing is to test a program's test suite. To do so, a mutator injects small bugs that alter the semantics of a program, such as changing a $0$ to a $1$ or turning $x > y$ into $x < y$. The hypothesis is that a good test suite should be able to catch these artificial bugs, and there is a substantial evidence that the ability to catch both artificial and real-world bugs is strongly correlated~\citep{just:mutants}. 

Our technique differs from mutation testing in a key way: we make program edits that would not affect test cases, but affect LLM predictions. We make minimal, semantics-preserving edits that lead to type mispredictions for a given LLM. The nature of code allows us to construct these edits in a sound and scalable way (\cref{building_prompts}). 

\paragraph{Activation Steering}

It is well known that even the most capable LLMs are sensitive to small variations in prompts. Prior work uses a black-box approach to study these phenomena by looking at model performance on programming tasks~\citep{hooda2024large, tambon2024bugs}. In contrast, we investigate type prediction with a whitebox approach. We use activation steering to query what a model's inner activations on code prompts reveal about its understanding of type systems. 

Activation steering is an inference-time model editing technique used to control model behavior. Research has shown that steering can moderate negative qualities like deceitfulness and sycophancy in model outputs~\citep{rimsky-etal-2024-steering, li2024inference}. Steering uses targeted steering vectors computed from model activations over positive and negative outputs. The intuition is that by quantifying the difference between positive and negative outputs, we can edit (steer) prediction away from the negative. Steering can be used to interpret the causal features behind model predictions by verifying that the structure of model internal representations is consistent with how language works. For example, steering has been used to verify that models encode faithful representations of English grammar and verbs ~\citep{ravfogel2021counterfactual}. Similarly, we use steering to show that models have a robust understanding of code and type systems.





\section{Methodology}

\subsection{Adversarial Type Prediction Tasks}
\label{building_prompts}

Our goal is to build a dataset of type prediction tasks that models fail to solve correctly, but have known working solutions. Different models fail and succeed at different tasks, so the datasets will be model-dependent. 

We present a variation of mutation testing that constructs minimal, semantics-preserving edits that trigger mispredictions. These edits are automated and applied randomly to programs from GitHub, allowing us to build challenging type prediction tasks at scale. Our edits produce programs that have unconventional syntax, but have the structure and behavior of real code.


\paragraph{Type Prediction Prompt Format}

We build datasets for both LLMs pretrained on code and instruction-tuned models.

Contemporary LLMs trained on code typically preprocess their training data to \emph{fill-in-the-middle} (FIM)~\citep{bavarian2022efficient,fried:incoder}. FIM training (1)~splits $\approx50\%$ of training items into three chunks---prefix, middle, and suffix---of random lengths; (2)~adds special tokens to the start of each chunk; and (3)~reorders the middle chunk to appear last. The language modeling training objective remains unchanged. At inference time, this allows models to generate the middle chunk, conditioned on the prefix and the suffix using a decoder-only LLM. \Cref{task-example} shows an example type prediction task, where we want the model to predict the type annotation for the argument $s$, which is in the middle of the program. To do so, we construct a prompt that marks the prefix and suffix with the model-specific FIM tokens (\cref{fim-example}).

In contrast, for instruction-tuned models, we formulate type prediction as a two-turn conversation between the user and assistant using the model-specific chat template (\cref{fim-chat-example}). The prompt includes the instruction to fill in the target type annotation site \texttt{<FILL>}. We include the prefix in the model's answer so that the model produces a well-formed program.

\paragraph{Semantics-preserving Code Edits}
\label{mutation-edits}

For each model $M$, we first build a dataset of ``easy'' type prediction tasks that $M$ solves correctly.\footnote{These files are in the training corpora for most models and we find that models easily predict types.} For Python, we use  ManyTypes4Py~\citep{mir2021manytypes4py}, a dataset of code from 5,382 Python projects with Python type annotations that successfully type-check. For TypeScript, we filter The Stack~\citep{kocetkov:the-stack} to find 1.1M TypeScript files that type-check. This ensures that the expected gold labels for type annotations are correct. Every program $p$ in the dataset may have several type annotations $\mathit{var}: t \in p$, and each of these annotations is a potential type annotation task. From these files, we build a large set of type prediction prompts $(p^+,t)$ where $M$ succeeds at type prediction. This dataset is potentially class-imbalanced, since models are unsurprisingly are better at predicting builtin types than user-defined types. We make sure to balance the distribution of types for our experiments \Cref{section:test-sets}.

Secondly, for each model $M$, we build a dataset of ``hard'' type prediction tasks that $M$ cannot solve. We select an easy task from the previous dataset, $(p^+,t)$ and incrementally apply the following semantics-preserving program edits at random.
\begin{inparaenum}[1)]
\item \emph{Rename variable:} We select a function/method argument and rename it to an arbitrary name that does not conflict with other variables.
\item \emph{Remove type annotation:} We select a type annotation (excluding the target $t$) and delete it. In a gradually typed language, removing or relaxing an annotation does not alter program semantics.
\item \emph{Rename user-defined type:} We select an arbitrary type definition (e.g., a class name or a type alias) and rename it to an arbitrary name that does not conflict with other names in the program.
\item \emph{Rename builtin type:} We introduce a type alias for a builtin type.
\end{inparaenum}
\Cref{mutation-examples} illustrates several separate edits to a program. 

The aforementioned edits do not change the type structure of the program. They make $p^+$ look different, but the target type $t$ remains unchanged. After applying each edit, we prompt $M$ to predict the type annotation. If $M$ mispredicts, we stop and use the current program as a failing type prediction task $(p^-, t)$. By construction, this is an adversarial type prediction task that $M$ fails to solve due to syntactic changes.

If $p^+$ is particularly simple, we may fail to construct $(p^-, t)$. In practice, we get several thousand challenging examples for each model, even in ablations where we restrict set of edits that we perform. \Cref{real-world-py-steering-pair} illustrates a real example from our dataset that makes a model mispredict. Note that a single edit often alters $p^+$ at several points.

We automatically construct $p^-$  by manipulating the concrete syntax tree of TypeScript and Python using TreeSitter-based parsers. This allows us to build these edits correctly and at scale.


\begin{figure*}[!t]
    \centering
    \includegraphics[width=\linewidth]{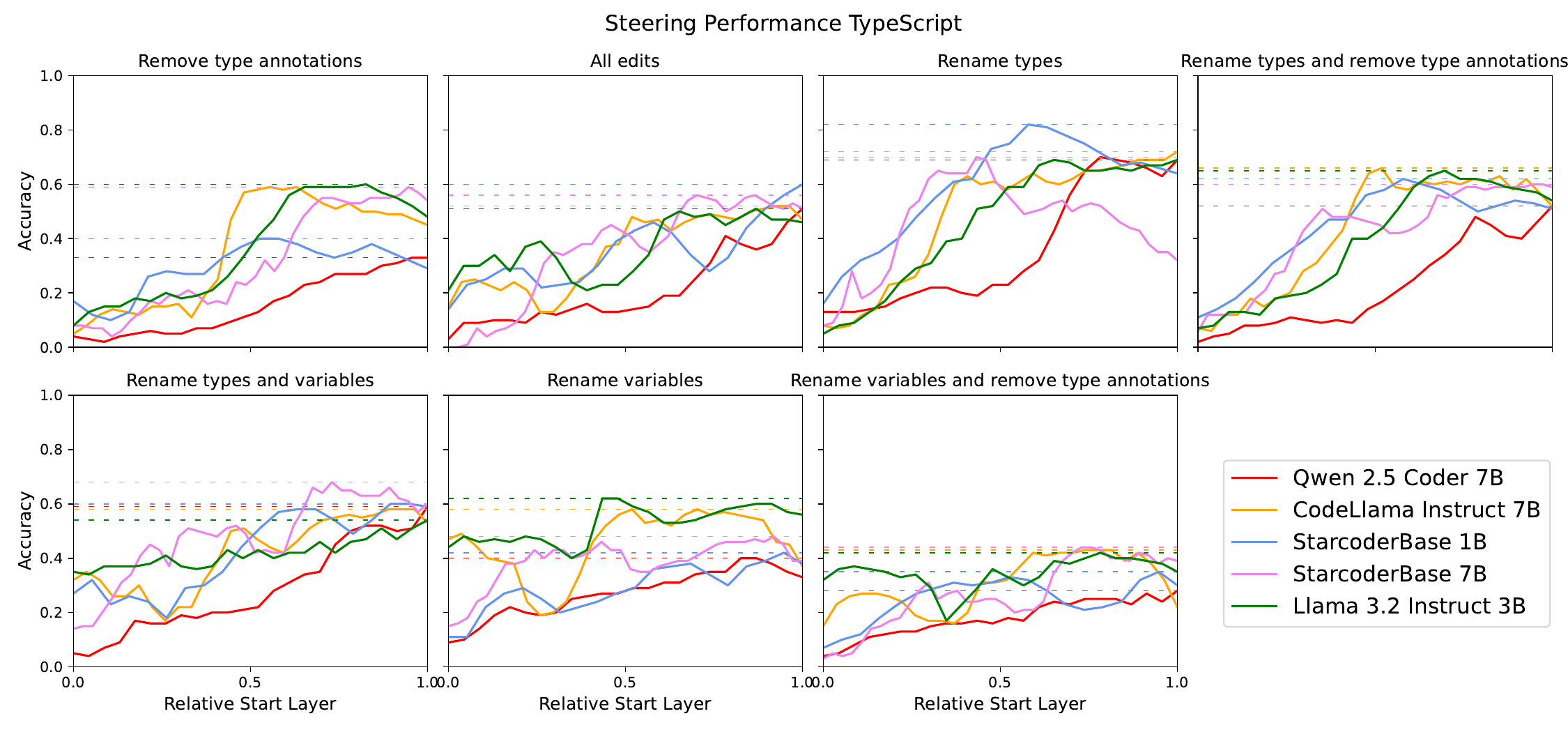}
    \caption{Steering accuracy for all models on the TypeScript test set, with steering on five consecutive layers. The models have a varying number of layers, so the $x$-axis is normalized: for a model with $n$ layers, $x=0$ indicates steering on the first five layers, and $x=1$ indicates steering on the last five layers.}
    \label{fig:all-models}
\end{figure*}

\paragraph{Test sets and class balance}
\label{section:test-sets}

For each model, we build test sets of 100 type prediction tasks $(p^-,t)$ that the model gets wrong. The natural distribution of type annotations is heavily skewed toward built-in and primitive types, thus we class-balance the test set to ensure that no target type $t$ occurs more than four times. Each test set has a mix of both built-in and user-defined types. This ensures that our evaluation is not skewed by reporting success on the most common types. We use the same class-balancing approach to construct the steering dataset for activation steering vectors, described below.

\subsection{Finding the Type Prediction Mechanism}

Why might a model fail to solve a type prediction task $(p^-,t)$, when it succeeded at the original task $(p^+,t)$? 
Note that since $p^+$ is sourced from GitHub-based datasets, the model was trained on these programs, whereas for the edited program $p^-$, by construction the model has likely never been trained on similar syntax.
There are two hypotheses:
\begin{inparaenum}
    \item the model has \emph{not} learned a robust mechanism for type prediction that generalizes outside of training data and resists adversarial prompts, basing its prediction on text features rather than program semantics;
    \item the model has a robust mechanism for type prediction, but it does not activate on adversarial prompts.
\end{inparaenum}
We argue that hypothesis 2 is correct. Using activation steering, we build steering vectors $\mathbf{v}^\ell$ that, when added to layer $\ell$, can activate robust type prediction on adversarial prompts. We present how we construct $\mathbf{v}^\ell$ below.

\paragraph{Constructing Steering Vectors}


For a given model $M$, we construct a dataset of triples $(p_i^+,p_i^-,t_i) \in \mathcal{D}$ where $p_i^-$ is an edited version of $p_i^+$, the maximum likelihood generation is $M(p_i^+)=t_i$, and $M(p^-) \ne t$. We apply forward passes $M(p_i^+), M(p_i^-)$ and save model activations of the last token before the type prediction token. Concretely, this involves pausing the model's forward pass at a layer $\ell_j$ of the transformer and saving the output of that layer, before it gets fed to subsequent layers. We write $A_{\ell}(x)$ to denote the activation vector at layer $\ell$ for prompt $x$. We compute steering vectors $\mathbf{v}_{\ell}$---one for each layer---as the mean difference between positive and negative activations at that layer:
\begin{equation}
\mathbf{v}_{\ell} = \frac{1}{|\mathcal{D}|}\sum_{(p_i^+,p_i^-,t) \in \mathcal{D}}\left(A_{\ell}(p_i^+) - A_{\ell}(p_i^-)\right)
\label{eq:sub}
\end{equation}
We compute steering tensors using hundreds of positive and negative prompt pairs for each of our edits, described previously \Cref{mutation-edits}. 

The intuition behind \cref{eq:sub} is that the resulting vector represents a transformation in activation space that separates the model's incorrect predictions from correct ones. Thus adding $\mathbf{v}_\ell$ to layer $\ell$ should prompt the model to shift to an internal mechanism not usually enabled on the adversarial prompts. We determine the layer $\ell$ experimentally, and also consider steering at up to five adjacent layers.

\begin{figure*}[t]
    \centering
    \includegraphics[width=\linewidth]{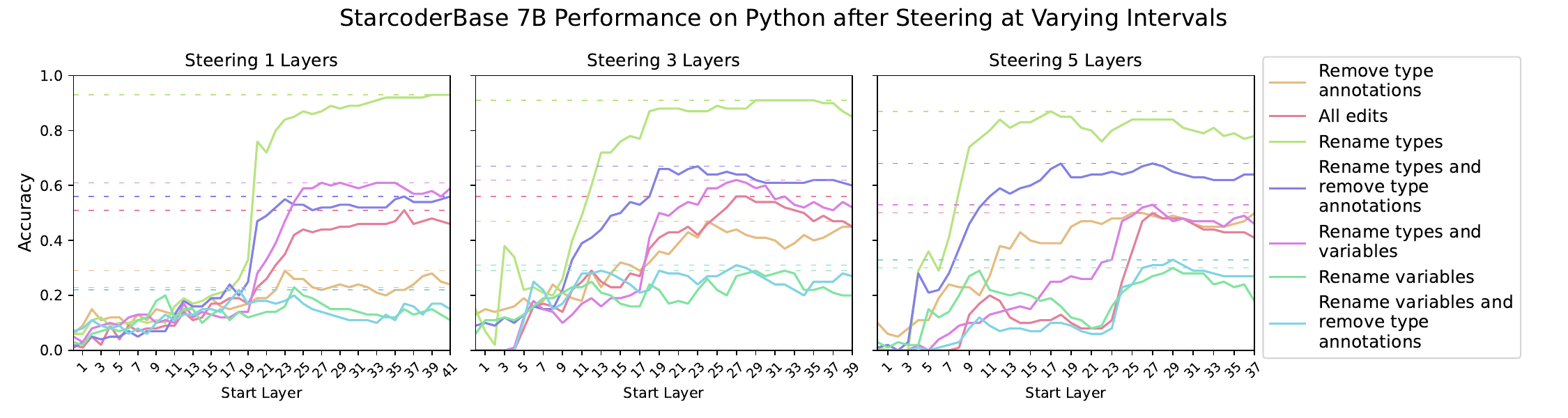}
    \caption{Steering accuracy for StarCoderBase 7B on Python. Each plot show steers in one, three, and five consecutive layers respectively.}
    \label{fig:layer_intervals}
\end{figure*}

\begin{figure*}
    \centering
    \includegraphics[width=\linewidth]{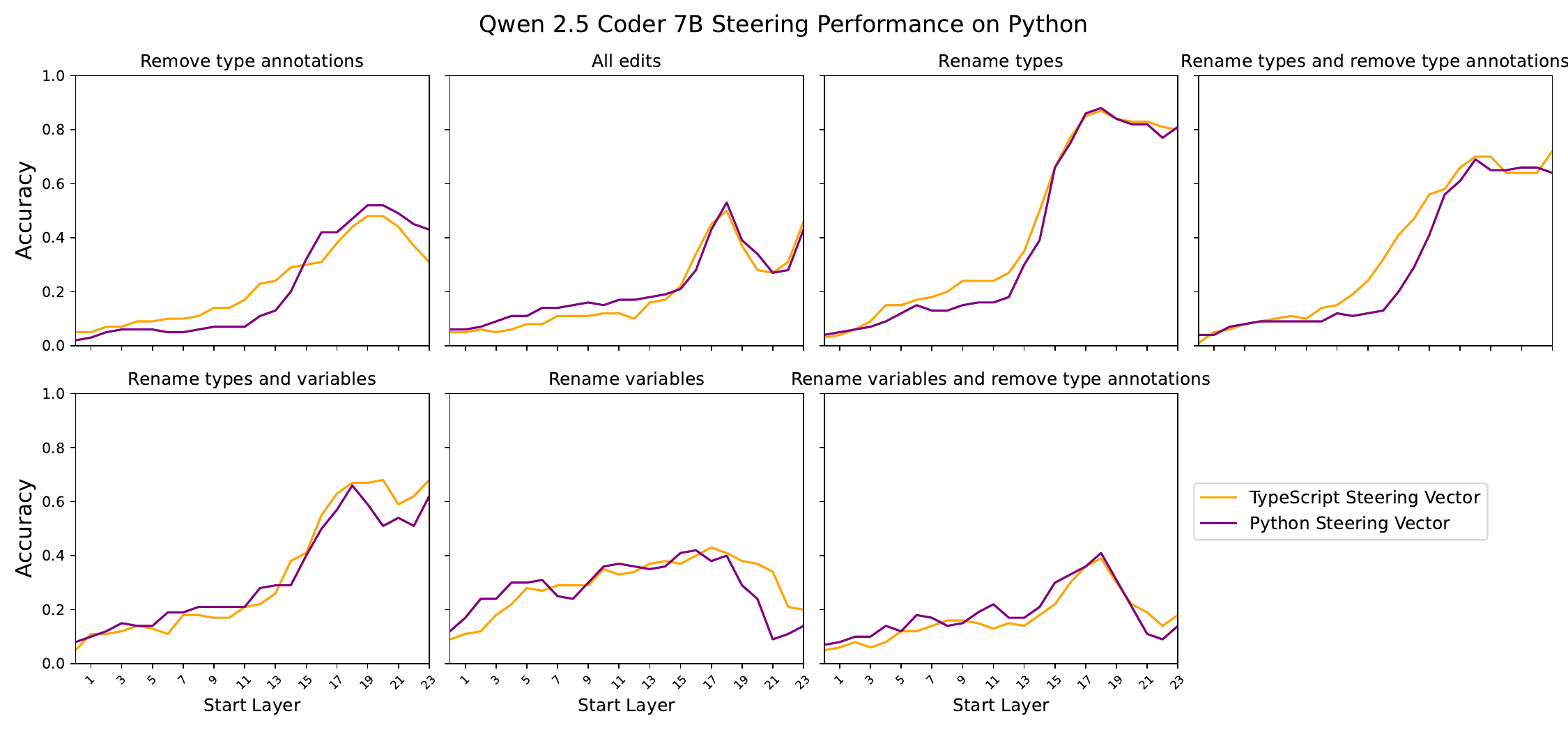}
    \caption{For Qwen 2.5 Coder 7B, we plot the performance of TypeScript steering vectors on the Python test set. We compare with the performance of steering vectors constructed from Python programs and find that the two achieve comparable accuracy.  In the appendix we report similar results for all other models (\cref{appendix:lang-transfer}).}
    \label{fig:lang_transfer}
\end{figure*}

\section{Results}

\subsection{Steering Improves Type Prediction on Out-of-Distribution Tasks}
\label{section:steering-all-models}

\Cref{fig:all-models} shows TypeScript test-set accuracy on every model with steering. Each subfigure ablates the set of edit operations used to construct the type prediction tasks so that we can see the effectiveness of steering on different edits. The $x$-axis indicates the relative position of layer $\ell$ where we apply $\textbf{v}^\ell$. 
($x=0$ indicates that $\ell$ is the first layer and $x=1$ indicates that it is the last layer.) In these experiments we apply $\textbf{v}^\ell$ to five adjacent layers $\ell \cdots \ell+4$, which we find is more effective than steering fewer layers (\Cref{section:layer-ablation}).

The figures show that steering is most effective in the later middle layers of every model, which suggests that this is where the type prediction mechanism lies. Recall that every type prediction task in the test sets are tasks that the model gets wrong without steering, thus the baseline accuracy is zero. When we construct $p^-$ using all possible edits, steering in the middle layers corrects mispredicted types on 50\%-60\% of the test set (varying by model). Steering is most effective when we construct $p^-$ by just renaming types, and corrects mispredictions on up to 80\% of the test set. We discuss steering performance in more depth in \Cref{section:error-analysis}.

\begin{figure*}[t]
    \centering
    \includegraphics[width=\linewidth]{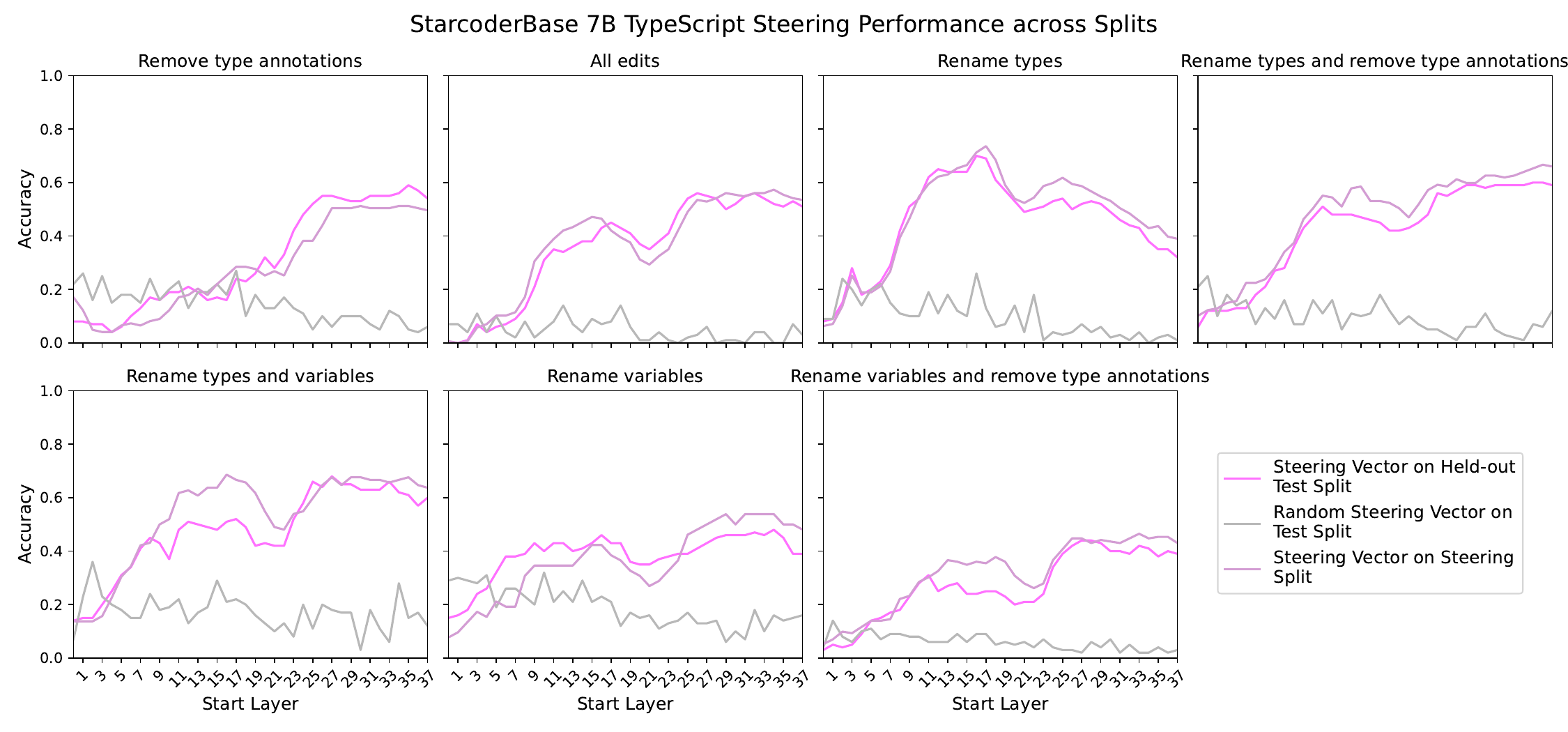}
    \caption{Steering accuracy for StarCoderBase 7B on TypeScript prompts on the test set, the steering set itself, and a random steering vector.  Random performs poorly; the test and steering sets have similar performance.}
\label{fig:steering-splits}
\end{figure*}

Overall, results indicate that we can find a $\mathbf{v}^\ell$ for each model that enables a robust type prediction even for adversarial type prediction tasks.
While \cref{fig:all-models} shows results for TypeScript, we have similar results for Python in the appendix, where steering is even more effective for certain edits (\Cref{fig:all_models-lang_py-interval_5}).

\subsection{Types Are Predicted Over Several Layers}
\label{section:layer-ablation}

The type prediction mechanism may span several layers. Therefore, we consider steering at one, three, and five adjacent layers. \Cref{fig:layer_intervals} shows the effect of this ablation on StarCoderBase-7B with Python: the $x$-axis indicates the start layer for steering and the $y$-axis is test-set accuracy. We find that steering on five layers is most effective. The appendix has similar results for TypeScript and the other models (\cref{appendix:model-results}, \cref{appendix:interval-results}).

\subsection{The Type Prediction Mechanism Is Shared Between Languages}
\label{section:lang-transfer}

Python and TypeScript are syntactically distinct, but their semantics have a lot in common~\cite{politz:lambda-py,bierman:ts}. Both languages are gradually typed. So, could it be that LLMs learn a type prediction mechanism that is language agnostic? To test this hypothesis, we evaluate if steering vectors built on TypeScript data can improve the accuracy of Python type prediction, and vice versa. We conduct this experiment with each of our datasets: we steer a model using vectors from language $A$ but evaluate on the corresponding held-out test set from language $B$. \Cref{fig:lang_transfer} shows that this is nearly as effective as steering prediction on the same language.

This result suggests that models learn similar representations of types across languages. The interchangeable nature of steering vectors suggests that models store shared concepts (e.g., types) in similar vector subspaces across languages. This provides some insight on how models internalize shared concepts across languages through consistent structures in activation space.


\subsection{Steering Outperforms Other Baselines}
\label{section:steering-baselines}

\paragraph{Random baseline}

A competing hypothesis to the one that we advance is the following: adding $\mathbf{v}^\ell$ is just adding noise, and steering is effectively just resampling from the output distribution. To refute this, we also steer with with a random vector and find that the computed steering vectors significantly outperform the random baseline (\Cref{fig:steering-splits}). This indicates that the steering vectors we compute perform true, localized transformations \emph{towards the correct type prediction task} in activation space.

\Cref{fig:steering-splits} also shows the performance of steering on the prompts $p^-$ from the steering set. We find that test-set and steering-set accuracy are approximately the same. This suggests that steering tensors can generalize outside the specific types and programs they were built from. We report results for this experiment for all our models in \Cref{appendix:splits}.

\begin{figure*}[t]
    \centering
    \includegraphics[width=\linewidth]{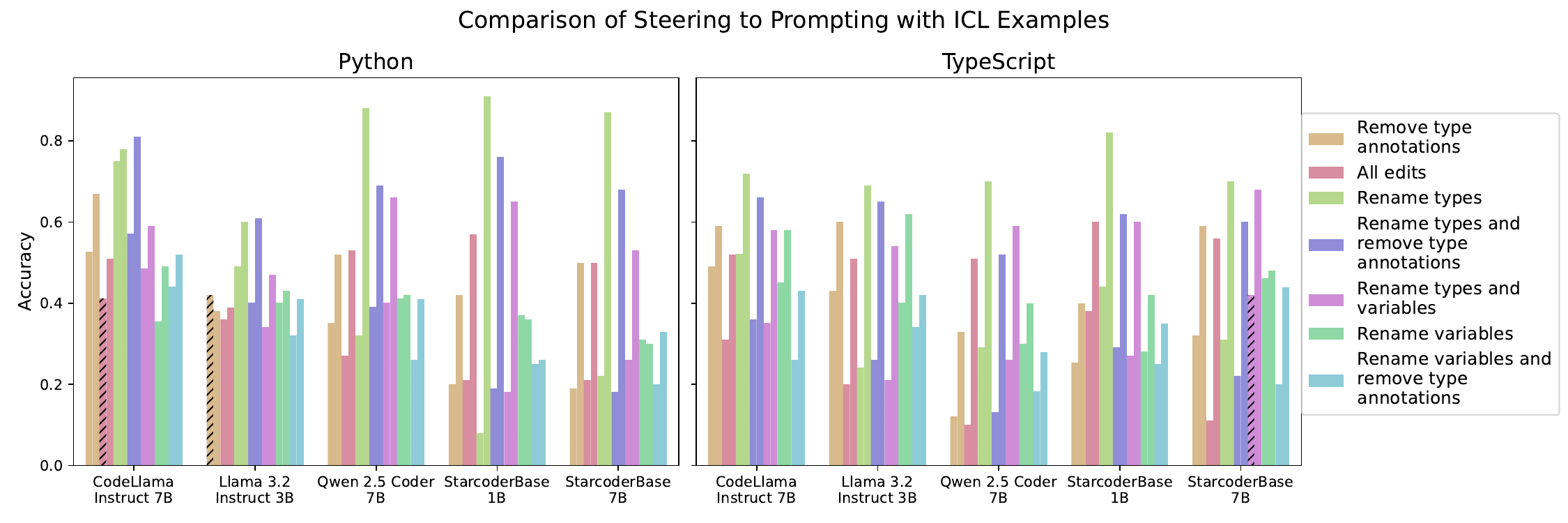}
    \caption{For each model, language and edit, we plot the best performance of steering vectors against in-context prompting (hatched bars).}
    \label{fig:icl-comparison}
\end{figure*}

\paragraph{In-context learning}

The usual way to instruct an LLM towards the correct task is with in-context examples (ICL). We perform an experiment where instead of steering, we prompt the model with two examples of adversarial type prediction tasks $(p^-,t)$. We find that prompting almost always underperforms steering (\cref{fig:icl-comparison}). This indicates that directly calculating the steering vector is a more robust way to enable the model's type prediction mechanism on adversarial programs.

\subsection{Steering Enables Type Prediction But Does Not Improve Type Precision}
\label{section:error-analysis}

\begin{figure}[t]
    \centering
    \includegraphics[width=\linewidth]{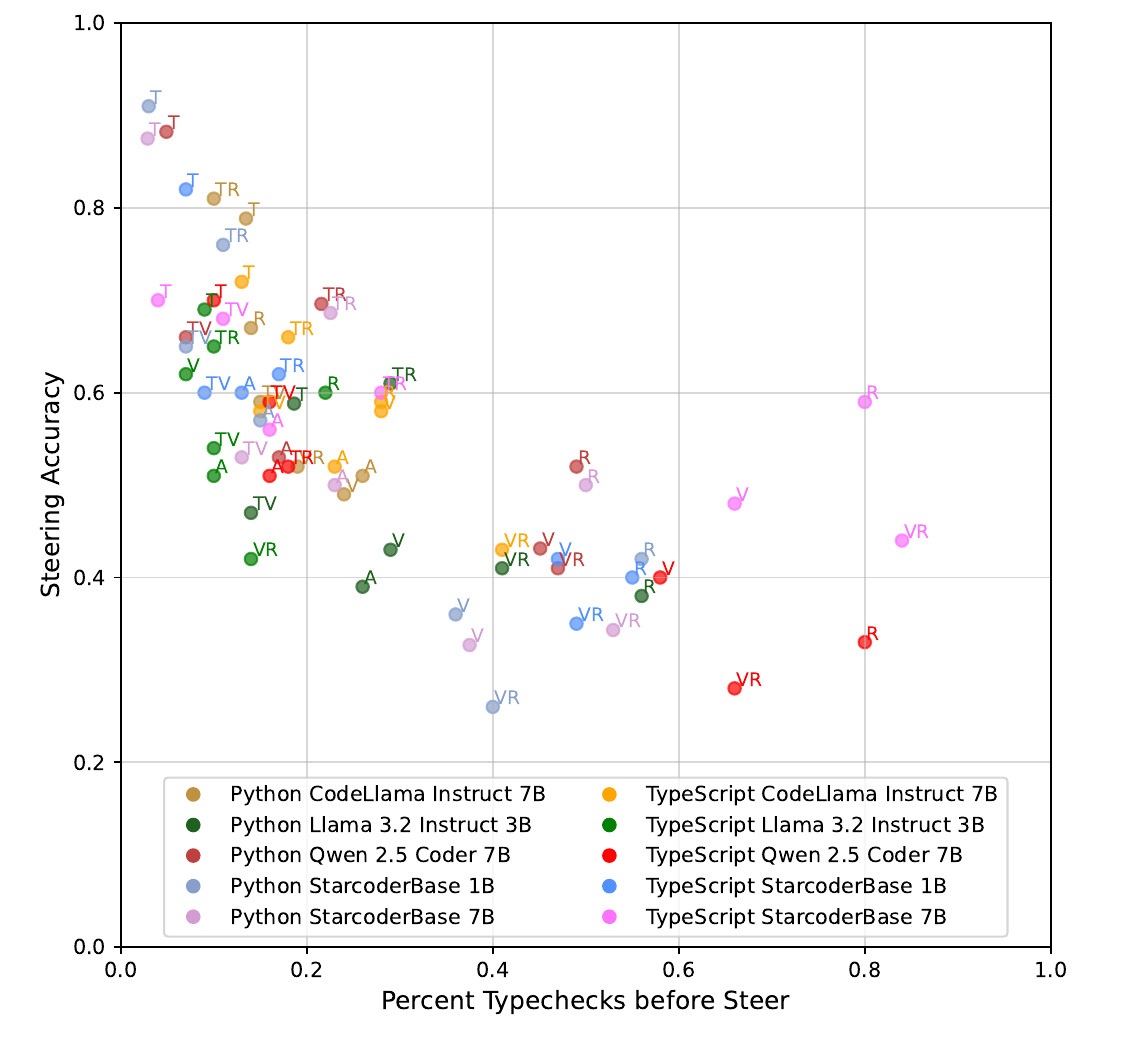}
    \caption{
    For every combination of model and edit-type, we plot the accuracy of type prediction on steering vs. the percentage of programs that are type-correct with the original, mispredicted type. The labels are: $V$ for renaming variables; $T$ for renaming types; $R$ for removing type annotations, and combinations of these.}
    \label{fig:correlation}
\end{figure}

Why doesn't steering always correct mispredictions? A complication of type prediction is that there may be several solutions to a type prediction problem that are type-correct, though some solutions are more precise than others. Therefore, if a model $M$ fails a task $(p^-,t)$ and predicts $M(p^-) = t'$, where $t' \ne t$, it may be the case that $t'$ is still a type-correct prediction. In \cref{fig:correlation}, we plot the accuracy of every combination of model and type of edit. On the $y$-axis we report steering accuracy and on the $x$-axis we report the fraction of programs where $p^-$ with the mispredicted type $t'$ is still type-correct (i.e., passes the type-checker).
We find a strong negative correlation ($r(68) = -0.687, p < 5.16 \times 10^{-11}$) between steering accuracy and type-correctness \emph{before} steering. When the model predicts a type that introduces a type error, steering is able to correct it. However, when the model merely predicts a unexpected type, steering is not as effective at directing the model to the expected answer.

Qualitatively, looking at these results, we find that most of these unexpected types are imprecise types, such as \emph{any}, or \emph{dict} instead of \emph{Config}. Overall, this experiment shows that we have identified the mechanism that enables the type prediction task, but not a mechanism that allows us to control the degree of type precision. Whether or not it is possible to identify such a mechanism in LLMs is a topic for future work.

\section{Conclusion}

Collectively, our results indicate that steering vectors steer the model toward a mechanism for type prediction that
\begin{inparaenum}[1)]
\item generalizes across different source codes;
\item is less sensitive to semantically irrelevant features; and
\item generalizes across the languages we study.
\end{inparaenum}

Given these observations, we conclude that there exists a robust mechanism for type prediction in LLMs which, when activated through activation steering, is more robust against adversarial programs. Furthermore, this mechanism is difficult to activate with prompting. This finding shows that it is insufficient to make conclusions about model's learned capabilities based on outputs alone. 

Whether a model is capable of performing robust and generalizable type prediction is a question of correctly aligning the model to the task. Activation steering is capable of performing this alignment for localized edits. Fine-tuning directly on edits could improve performance, but this defeats the purpose of studying behavior on adversarial or unseen prompts. In order to effectively use the information learned by LLMs, further research into how this information is organized, stored and retrieved is necessary.



\section{Limitations}

Our findings shed light on how CodeLLMs display robust type prediction for TypeScript and Python. Both these languages are well represented in CodeLLM training corpora. However, our findings may not extend to low-resource gradually typed languages, e.g., Typed Racket~\cite{th:typed-scheme} or Luau~\cite{brown:luau-part-two} since the performance of base models on these languages is very poor. Future work will include implementing semantics-preserving edits for other languages.

Our investigation focuses on type prediction to understand whether models learn program semantics along with syntax. The reduced scope allows us to conduct an in depth evaluation of models and steering vectors. Future research may focus on studying learned representations of other code concepts such as control flow, data races and vulnerabilities. 

We apply automatically generated edits to prompts as a scalable way to approximate real code with arbitrary syntax. To ensure diverse and comprehensive test sets, we use hundreds of real programs for each model, varying the source code, target types, and programming languages. However, we note that these automatically generated edits may not fully capture the complete variance possible in code.

\section{Ethics Statement}

The purpose of this work is to understand whether LLMs perform type prediction using robust mechanisms. It is our view that interpreting LLMs is necessary for understanding whether models approach programming in a principled way. As LLMs become more integrated into developers' workflows, model errors could compromise the security of entire systems. For this reason, we make a first investigation into understanding the mechanisms behind model prediction.

We take care to use publicly available code for our experiments. Our TypeScript dataset is derived from a subset of The Stack v1.2, which contains permissively licensed data with personal identifying information (PII) filtered. The ManyTypes4Py dataset is funded by the European Commission, which follows data privacy laws under the EU General Data Protection Regulation (GDPR). These datasets are intended for LLMs, which this paper investigates.

\ifanon
\else
\section*{Acknowledgments}

Portions of this work are implemented with NNSight and NDIF~\cite{nnsight-ndif}.
We thank Ming-Ho Yee for help with the TypeScript dataset that we use in this work~\cite{yee:dissertation}. 
This material is based upon
work supported by the U.S. Department of Energy, Office of Science under Award Number DESC0025613.

We thank Northeastern Research Computing for support with the Northeastern University Explorer cluster. This work used the Delta cluster at the National Center for Supercomputing Applications (NCSA) through allocation CIS230213 from the Advanced Cyberinfrastructure Coordination Ecosystem: Services \& Support (ACCESS) program, which is supported by U.S. National Science Foundation grants 2138259, 2138286, 2138307, 2137603, and 2138296.

\emph{Disclaimer}: This report was prepared as an account of work sponsored by an agency of the United
States Government. Neither the United States Government nor any agency thereof, nor any of their
employees, makes any warranty, express or implied, or assumes any legal liability or responsibility
for the accuracy, completeness, or usefulness of any information, apparatus, product, or process
disclosed, or represents that its use would not infringe privately owned rights. Reference herein to
any specific commercial product, process, or service by trade name, trademark, manufacturer, or
otherwise does not necessarily constitute or imply its endorsement, recommendation, or favoring by
the United States Government or any agency thereof. The views and opinions of authors expressed
herein do not necessarily state or reflect those of the United States Government or any agency
thereof.

\fi

\bibliography{main,arjun}

\begin{thebibliography}{48}
\providecommand{\natexlab}[1]{#1}

\bibitem[{Abacus(2019)}]{abacus:ts}
Abacus. 2019.
\newblock \href {https://blog.abacus.com/how-we-completed-a-partial-typescript-migration-in-six-months/} {How {We} {Completed} a ({Partial}) {TypeScript} {Migration} {In} {Six} {Months}}.
\newblock Section: Developing In Real Time.

\bibitem[{Bavarian et~al.(2022)Bavarian, Jun, Tezak, Schulman, McLeavey, Tworek, and Chen}]{bavarian2022efficient}
Mohammad Bavarian, Heewoo Jun, Nikolas Tezak, John Schulman, Christine McLeavey, Jerry Tworek, and Mark Chen. 2022.
\newblock Efficient training of language models to fill in the middle.
\newblock \emph{arXiv preprint arXiv:2207.14255}.

\bibitem[{Bierman et~al.(2014)Bierman, Abadi, and Torgersen}]{bierman:ts}
Gavin Bierman, Martín Abadi, and Mads Torgersen. 2014.
\newblock \href {https://doi.org/10.1007/978-3-662-44202-9_11} {Understanding {TypeScript}}.
\newblock In \emph{{ECOOP} 2014 – {Object}-{Oriented} {Programming}}, Lecture {Notes} in {Computer} {Science}, pages 257--281, Berlin, Heidelberg. Springer.

\bibitem[{Campora et~al.(2018)Campora, Chen, Erwig, and Walkingshaw}]{campora:migrating}
John~Peter Campora, Sheng Chen, Martin Erwig, and Eric Walkingshaw. 2018.
\newblock Migrating {Gradual} {Types}.
\newblock \emph{Proceedings of the ACM on Programming Languages (PACMPL)}, 2(POPL).

\bibitem[{Cartwright and Fagan(1991)}]{wright:soft-typing}
Robert Cartwright and Mike Fagan. 1991.
\newblock Soft typing.
\newblock In \emph{{ACM} {SIGPLAN} {Conference} on {Programming} {Language} {Design} and {Implementation} ({PLDI})}.

\bibitem[{DeMillo et~al.(1978)DeMillo, Lipton, and Sayward}]{demillo:test-data-selection}
R.A. DeMillo, R.J. Lipton, and F.G. Sayward. 1978.
\newblock \href {https://doi.org/10.1109/C-M.1978.218136} {Hints on {Test} {Data} {Selection}: {Help} for the {Practicing} {Programmer}}.
\newblock \emph{Computer}, 11(4):34--41.
\newblock Conference Name: Computer.

\bibitem[{Dubey et~al.(2024)Dubey, Jauhri, Pandey, Kadian, Al-Dahle, Letman, Mathur, Schelten, Yang, Fan et~al.}]{dubey2024llama}
Abhimanyu Dubey, Abhinav Jauhri, Abhinav Pandey, Abhishek Kadian, Ahmad Al-Dahle, Aiesha Letman, Akhil Mathur, Alan Schelten, Amy Yang, Angela Fan, et~al. 2024.
\newblock The llama 3 herd of models.
\newblock \emph{arXiv preprint arXiv:2407.21783}.

\bibitem[{{Felix Rieseberg}(2017)}]{slack:ts}
{Felix Rieseberg}. 2017.
\newblock \href {https://slack.engineering/typescript-at-slack/} {{TypeScript} at {Slack}}.
\newblock Section: Uncategorized.

\bibitem[{Fiotto-Kaufman et~al.(2024)Fiotto-Kaufman, Loftus, Todd, Brinkmann, Juang, Pal, Rager, Mueller, Marks, Sharma, Lucchetti, Ripa, Belfki, Prakash, Multani, Brodley, Guha, Bell, Wallace, and Bau}]{nnsight-ndif}
Jaden Fiotto-Kaufman, Alexander~R Loftus, Eric Todd, Jannik Brinkmann, Caden Juang, Koyena Pal, Can Rager, Aaron Mueller, Samuel Marks, Arnab~Sen Sharma, Francesca Lucchetti, Michael Ripa, Adam Belfki, Nikhil Prakash, Sumeet Multani, Carla Brodley, Arjun Guha, Jonathan Bell, Byron Wallace, and David Bau. 2024.
\newblock \href {https://arxiv.org/abs/2407.14561} {Nnsight and ndif: Democratizing access to foundation model internals}.
\newblock \emph{Preprint}, arXiv:2407.14561.

\bibitem[{Fried et~al.(2023)Fried, Aghajanyan, Lin, Wang, Wallace, Shi, Zhong, Yih, Zettlemoyer, and Lewis}]{fried:incoder}
Daniel Fried, Armen Aghajanyan, Jessy Lin, Sida Wang, Eric Wallace, Freda Shi, Ruiqi Zhong, Wen-tau Yih, Luke Zettlemoyer, and Mike Lewis. 2023.
\newblock {InCoder}: {A} {Generative} {Model} for {Code} {Infilling} and {Synthesis}.
\newblock In \emph{International {Conference} on {Learning} {Representations} ({ICLR})}.

\bibitem[{Gu et~al.(2024)Gu, Roziere, Leather, Solar-Lezama, Synnaeve, and Wang}]{pmlr-v235-gu24c}
Alex Gu, Baptiste Roziere, Hugh~James Leather, Armando Solar-Lezama, Gabriel Synnaeve, and Sida Wang. 2024.
\newblock \href {https://proceedings.mlr.press/v235/gu24c.html} {{CRUXE}val: A benchmark for code reasoning, understanding and execution}.
\newblock In \emph{Proceedings of the 41st International Conference on Machine Learning}, volume 235 of \emph{Proceedings of Machine Learning Research}, pages 16568--16621. PMLR.

\bibitem[{Harper and Mitchell(1993)}]{harper:xml}
Robert Harper and John~C. Mitchell. 1993.
\newblock \href {https://doi.org/10.1145/169701.169696} {On the type structure of standard {ML}}.
\newblock \emph{ACM Transactions on Programming Languages and Systems}, 15(2):211--252.

\bibitem[{Hellendoorn et~al.(2018)Hellendoorn, Bird, Barr, and Allamanis}]{hellendoorn:dlti}
Vincent~J. Hellendoorn, Christian Bird, Earl~T. Barr, and Miltiadis Allamanis. 2018.
\newblock Deep {Learning} {Type} {Inference}.
\newblock In \emph{{ACM} {Joint} {Meeting} on {European} {Software} {Engineering} {Conference} and {Symposium} on the {Foundations} of {Software} {Engineering} ({ESEC}/{FSE})}.

\bibitem[{Henglein and Rehof(1995)}]{henglein:scheme-to-ml}
Fritz Henglein and Jakob Rehof. 1995.
\newblock Safe polymorphic type inference for a dynamically typed language: {Translating} {Scheme} to {ML}.
\newblock In \emph{International {Conference} on {Functional} {Programming} {Languages} and {Computer} {Architecture} ({FPCA})}.

\bibitem[{Hooda et~al.(2024{\natexlab{a}})Hooda, Christodorescu, Allamanis, Wilson, Fawaz, and Jha}]{pmlr-v235-hooda24a}
Ashish Hooda, Mihai Christodorescu, Miltiadis Allamanis, Aaron Wilson, Kassem Fawaz, and Somesh Jha. 2024{\natexlab{a}}.
\newblock \href {https://proceedings.mlr.press/v235/hooda24a.html} {Do large code models understand programming concepts? {C}ounterfactual analysis for code predicates}.
\newblock In \emph{Proceedings of the 41st International Conference on Machine Learning}, volume 235 of \emph{Proceedings of Machine Learning Research}, pages 18738--18748. PMLR.

\bibitem[{Hooda et~al.(2024{\natexlab{b}})Hooda, Christodorescu, Allamanis, Wilson, Fawaz, and Jha}]{hooda2024large}
Ashish Hooda, Mihai Christodorescu, Miltos Allamanis, Aaron Wilson, Kassem Fawaz, and Somesh Jha. 2024{\natexlab{b}}.
\newblock Do large code models understand programming concepts? a black-box approach.
\newblock \emph{arXiv preprint arXiv:2402.05980}.

\bibitem[{Hui et~al.(2024)Hui, Yang, Cui, Yang, Liu, Zhang, Liu, Zhang, Yu, Dang et~al.}]{hui2024qwen2}
Binyuan Hui, Jian Yang, Zeyu Cui, Jiaxi Yang, Dayiheng Liu, Lei Zhang, Tianyu Liu, Jiajun Zhang, Bowen Yu, Kai Dang, et~al. 2024.
\newblock Qwen2. 5-coder technical report.
\newblock \emph{arXiv preprint arXiv:2409.12186}.

\bibitem[{{Jake Zimmerman}(2022)}]{stripe:sorbet}
{Jake Zimmerman}. 2022.
\newblock \href {https://stripe.com/blog/sorbet-stripes-type-checker-for-ruby} {Sorbet: {Stripe}’s type checker for {Ruby}}.

\bibitem[{Jesse et~al.(2022)Jesse, Devanbu, and Sawant}]{jesse:diversetyper}
Kevin Jesse, Premkumar Devanbu, and Anand~Ashok Sawant. 2022.
\newblock \href {https://doi.org/10.1109/TSE.2022.3178945} {Learning {To} {Predict} {User}-{Defined} {Types}}.
\newblock \emph{IEEE Transactions on Software Engineering}, pages 1--1.

\bibitem[{Jesse et~al.(2021)Jesse, Devanbu, and Ahmed}]{jesse:typebert}
Kevin Jesse, Premkumar~T. Devanbu, and Toufique Ahmed. 2021.
\newblock \href {https://doi.org/10.1145/3468264.3473135} {Learning type annotation: is big data enough?}
\newblock In \emph{Proceedings of the 29th {ACM} {Joint} {Meeting} on {European} {Software} {Engineering} {Conference} and {Symposium} on the {Foundations} of {Software} {Engineering}}, pages 1483--1486, Athens Greece. ACM.

\bibitem[{Just et~al.(2014)Just, Jalali, Inozemtseva, Ernst, Holmes, and Fraser}]{just:mutants}
René Just, Darioush Jalali, Laura Inozemtseva, Michael~D. Ernst, Reid Holmes, and Gordon Fraser. 2014.
\newblock \href {https://doi.org/10.1145/2635868.2635929} {Are mutants a valid substitute for real faults in software testing?}
\newblock In \emph{Proceedings of the 22nd {ACM} {SIGSOFT} {International} {Symposium} on {Foundations} of {Software} {Engineering}}, {FSE} 2014, pages 654--665, New York, NY, USA. Association for Computing Machinery.

\bibitem[{Kocetkov et~al.(2023)Kocetkov, Li, Allal, Li, Mou, Ferrandis, Jernite, Mitchell, Hughes, Wolf, Bahdanau, von Werra, and de~Vries}]{kocetkov:the-stack}
Denis Kocetkov, Raymond Li, Loubna~Ben Allal, Jia Li, Chenghao Mou, Carlos~Muñoz Ferrandis, Yacine Jernite, Margaret Mitchell, Sean Hughes, Thomas Wolf, Dzmitry Bahdanau, Leandro von Werra, and Harm de~Vries. 2023.
\newblock \href {http://arxiv.org/abs/2211.15533} {The {Stack}: 3 {TB} of permissively licensed source code}.
\newblock In \emph{Deep {Learning} for {Code} {Workshop} ({DL4C})}.

\bibitem[{Lehtosalo(2019)}]{dropbox:mypy}
Jukka Lehtosalo. 2019.
\newblock \href {https://dropbox.tech/application/our-journey-to-type-checking-4-million-lines-of-python} {Our journey to type checking 4 million lines of {Python}}.

\bibitem[{Li et~al.(2024)Li, Patel, Vi{\'e}gas, Pfister, and Wattenberg}]{li2024inference}
Kenneth Li, Oam Patel, Fernanda Vi{\'e}gas, Hanspeter Pfister, and Martin Wattenberg. 2024.
\newblock Inference-time intervention: Eliciting truthful answers from a language model.
\newblock \emph{Advances in Neural Information Processing Systems}, 36.

\bibitem[{Li et~al.(2023)Li, Allal, Zi, Muennighoff, Kocetkov, Mou, Marone, Akiki, Li, Chim, Liu, Zheltonozhskii, Zhuo, Wang, Dehaene, Davaadorj, Lamy-Poirier, Monteiro, Shliazhko, Gontier, Meade, Zebaze, Yee, Umapathi, Zhu, Lipkin, Oblokulov, Wang, Murthy, Stillerman, Patel, Abulkhanov, Zocca, Dey, Zhang, Fahmy, Bhattacharyya, Yu, Singh, Luccioni, Villegas, Kunakov, Zhdanov, Romero, Lee, Timor, Ding, Schlesinger, Schoelkopf, Ebert, Dao, Mishra, Gu, Robinson, Anderson, Dolan-Gavitt, Contractor, Reddy, Fried, Bahdanau, Jernite, Ferrandis, Hughes, Wolf, Guha, von Werra, and de~Vries}]{li:starcoder}
Raymond Li, Loubna~Ben Allal, Yangtian Zi, Niklas Muennighoff, Denis Kocetkov, Chenghao Mou, Marc Marone, Christopher Akiki, Jia Li, Jenny Chim, Qian Liu, Evgenii Zheltonozhskii, Terry~Yue Zhuo, Thomas Wang, Olivier Dehaene, Mishig Davaadorj, Joel Lamy-Poirier, João Monteiro, Oleh Shliazhko, Nicolas Gontier, Nicholas Meade, Armel Zebaze, Ming-Ho Yee, Logesh~Kumar Umapathi, Jian Zhu, Benjamin Lipkin, Muhtasham Oblokulov, Zhiruo Wang, Rudra Murthy, Jason Stillerman, Siva~Sankalp Patel, Dmitry Abulkhanov, Marco Zocca, Manan Dey, Zhihan Zhang, Nour Fahmy, Urvashi Bhattacharyya, Wenhao Yu, Swayam Singh, Sasha Luccioni, Paulo Villegas, Maxim Kunakov, Fedor Zhdanov, Manuel Romero, Tony Lee, Nadav Timor, Jennifer Ding, Claire Schlesinger, Hailey Schoelkopf, Jan Ebert, Tri Dao, Mayank Mishra, Alex Gu, Jennifer Robinson, Carolyn~Jane Anderson, Brendan Dolan-Gavitt, Danish Contractor, Siva Reddy, Daniel Fried, Dzmitry Bahdanau, Yacine Jernite, Carlos~Muñoz Ferrandis, Sean Hughes, Thomas Wolf, Arjun Guha, Leandro von
  Werra, and Harm de~Vries. 2023.
\newblock {StarCoder}: may the source be with you!
\newblock \emph{Transactions of Machine Learning Research (TMLR)}.

\bibitem[{{Lily Brown} et~al.(2023){Lily Brown}, {Andy Friesen}, and {Alan Jeffery}}]{brown:luau-part-two}
{Lily Brown}, {Andy Friesen}, and {Alan Jeffery}. 2023.
\newblock Goals of the {Luau} {Type} {System}, {Two} {Years} {On}.
\newblock ACM.

\bibitem[{{Luke Autry}()}]{heap:ts}
{Luke Autry}.
\newblock \href {https://heap.io/blog/migrating-to-typescript} {How we failed, then succeeded, at migrating to {TypeScript}}.

\bibitem[{Migeed and Palsberg(2020)}]{migeed:decidable}
Zeina Migeed and Jens Palsberg. 2020.
\newblock What is {Decidable} about {Gradual} {Types}?
\newblock \emph{Proceedings of the ACM on Programming Languages (PACMPL)}, 4(POPL).

\bibitem[{{Mihai Parparita}(2020)}]{quip:ts}
{Mihai Parparita}. 2020.
\newblock \href {https://quip.com/blog/the-road-to-typescript-at-quip-part-two} {The {Road} to {TypeScript} at {Quip}, {Part} {Two}}.

\bibitem[{Mir et~al.(2021)Mir, Lato{\v{s}}kinas, and Gousios}]{mir2021manytypes4py}
Amir~M Mir, Evaldas Lato{\v{s}}kinas, and Georgios Gousios. 2021.
\newblock Manytypes4py: A benchmark python dataset for machine learning-based type inference.
\newblock In \emph{2021 IEEE/ACM 18th International Conference on Mining Software Repositories (MSR)}, pages 585--589. IEEE.

\bibitem[{Pandi et~al.(2021)Pandi, Barr, Gordon, and Sutton}]{pandi_probabilistic_2021}
Irene~Vlassi Pandi, Earl~T. Barr, Andrew~D. Gordon, and Charles Sutton. 2021.
\newblock \href {https://arxiv.org/abs/2004.00348v3} {Probabilistic {Type} {Inference} by {Optimising} {Logical} and {Natural} {Constraints}}.

\bibitem[{Phipps-Costin et~al.(2021)Phipps-Costin, Anderson, Greenberg, and Guha}]{phipps-costin:typewhich}
Luna Phipps-Costin, Carolyn~Jane Anderson, Michael Greenberg, and Arjun Guha. 2021.
\newblock \href {https://doi.org/10.1145/3485488} {Solver-based {Gradual} {Type} {Migration}}.
\newblock \emph{Proceedings of the ACM on Programming Languages (PACMPL)}, 5(OOPSLA).

\bibitem[{Politz et~al.(2013)Politz, Martinez, Milano, Warren, Patterson, Li, Chitipothu, and Krishnamurthi}]{politz:lambda-py}
Joe~Gibbs Politz, Alejandro Martinez, Mae Milano, Sumner Warren, Daniel Patterson, Junsong Li, Anand Chitipothu, and Shriram Krishnamurthi. 2013.
\newblock \href {https://doi.org/10.1145/2509136.2509536} {Python: the full monty}.
\newblock In \emph{{ACM} {SIGPLAN} {Conference} on {Object} {Oriented} {Programmingm}, {Systems}, {Languages} and {Applications} ({OOPSLA})}, pages 217--232, Indianapolis, IN, USA. ACM.

\bibitem[{Rastogi et~al.(2012)Rastogi, Chaudhuri, and Hosmer}]{rastogi:gti}
Aseem Rastogi, Avik Chaudhuri, and Basil Hosmer. 2012.
\newblock The {Ins} and {Outs} of {Gradual} {Type} {Inference}.
\newblock In \emph{{ACM} {SIGPLAN}-{SIGACT} {Symposium} on {Principles} of {Programming} {Languages} ({POPL})}.

\bibitem[{Ravfogel et~al.(2021)Ravfogel, Prasad, Linzen, and Goldberg}]{ravfogel2021counterfactual}
Shauli Ravfogel, Grusha Prasad, Tal Linzen, and Yoav Goldberg. 2021.
\newblock Counterfactual interventions reveal the causal effect of relative clause representations on agreement prediction.
\newblock In \emph{Proceedings of the 25th Conference on Computational Natural Language Learning}, pages 194--209.

\bibitem[{Rimsky et~al.(2024)Rimsky, Gabrieli, Schulz, Tong, Hubinger, and Turner}]{rimsky-etal-2024-steering}
Nina Rimsky, Nick Gabrieli, Julian Schulz, Meg Tong, Evan Hubinger, and Alexander Turner. 2024.
\newblock \href {https://doi.org/10.18653/v1/2024.acl-long.828} {Steering llama 2 via contrastive activation addition}.
\newblock In \emph{Proceedings of the 62nd Annual Meeting of the Association for Computational Linguistics (Volume 1: Long Papers)}, pages 15504--15522, Bangkok, Thailand. Association for Computational Linguistics.

\bibitem[{Roziere et~al.(2023)Roziere, Gehring, Gloeckle, Sootla, Gat, Tan, Adi, Liu, Remez, Rapin et~al.}]{roziere2023code}
Baptiste Roziere, Jonas Gehring, Fabian Gloeckle, Sten Sootla, Itai Gat, Xiaoqing~Ellen Tan, Yossi Adi, Jingyu Liu, Tal Remez, J{\'e}r{\'e}my Rapin, et~al. 2023.
\newblock Code llama: Open foundation models for code.
\newblock \emph{arXiv preprint arXiv:2308.12950}.

\bibitem[{Rudenko(2020)}]{airbnb:ts-migrate}
Sergii Rudenko. 2020.
\newblock \href {https://medium.com/airbnb-engineering/ts-migrate-a-tool-for-migrating-to-typescript-at-scale-cd23bfeb5cc} {ts-migrate: {A} {Tool} for {Migrating} to {TypeScript} at {Scale}}.

\bibitem[{Siek and Taha(2006)}]{siek:gtlc}
Jeremy~G. Siek and Walid Taha. 2006.
\newblock Gradual {Typing} for {Functional} {Languages}.
\newblock In \emph{Scheme {Workshop}}.

\bibitem[{Siek and Vachharajani(2008)}]{siek:gti}
Jeremy~G. Siek and Manish Vachharajani. 2008.
\newblock Gradual {Typing} with {Unification}-based {Inference}.
\newblock In \emph{{ACM} {SIGPLAN} {International} {Symposium} on {Dynamic} {Languages} ({DLS})}.

\bibitem[{{Sumana Mohan} et~al.(2022){Sumana Mohan}, {Joe King}, {Ryan Burgess}, {Jem Young}, and {Stacy London}}]{netflix:ts}
{Sumana Mohan}, {Joe King}, {Ryan Burgess}, {Jem Young}, and {Stacy London}. 2022.
\newblock \href {https://frontendhappyhour.com/episodes/typescript-migration-strict-type-of-cocktails} {{TypeScript} migration - {Strict} type of cocktails - {Front} {End} {Happy} {Hour}}.

\bibitem[{Tambon et~al.(2024)Tambon, Dakhel, Nikanjam, Khomh, Desmarais, and Antoniol}]{tambon2024bugs}
Florian Tambon, Arghavan~Moradi Dakhel, Amin Nikanjam, Foutse Khomh, Michel~C Desmarais, and Giuliano Antoniol. 2024.
\newblock Bugs in large language models generated code.
\newblock \emph{arXiv preprint arXiv:2403.08937}.

\bibitem[{Tobin-Hochstadt and Felleisen(2006)}]{th:migration}
Sam Tobin-Hochstadt and Matthias Felleisen. 2006.
\newblock Interlanguage {Migration}: {From} {Scripts} to {Programs}.
\newblock In \emph{{ACM} {SIGPLAN} {International} {Symposium} on {Dynamic} {Languages} ({DLS})}.

\bibitem[{Tobin-Hochstadt and Felleisen(2008)}]{th:typed-scheme}
Sam Tobin-Hochstadt and Matthias Felleisen. 2008.
\newblock The {Design} and {Implementation} of {Typed} {Scheme}.
\newblock In \emph{{ACM} {SIGPLAN}-{SIGACT} {Symposium} on {Principles} of {Programming} {Languages} ({POPL})}.

\bibitem[{Wei et~al.(2020)Wei, Goyal, Durrett, and Dillig}]{wei:lambdanet}
Jiayi Wei, Maruth Goyal, Greg Durrett, and Isil Dillig. 2020.
\newblock {LambdaNet}: {Probabilistic} {Type} {Inference} using {Graph} {Neural} {Networks}.
\newblock In \emph{International {Conference} on {Learning} {Representations} ({ICLR})}.

\bibitem[{Yang et~al.(2024)Yang, Yang, Hui, Zheng, Yu, Zhou, Li, Li, Liu, Huang, Dong, Wei, Lin, Tang, Wang, Yang, Tu, Zhang, Ma, Xu, Zhou, Bai, He, Lin, Dang, Lu, Chen, Yang, Li, Xue, Ni, Zhang, Wang, Peng, Men, Gao, Lin, Wang, Bai, Tan, Zhu, Li, Liu, Ge, Deng, Zhou, Ren, Zhang, Wei, Ren, Fan, Yao, Zhang, Wan, Chu, Liu, Cui, Zhang, and Fan}]{qwen2}
An~Yang, Baosong Yang, Binyuan Hui, Bo~Zheng, Bowen Yu, Chang Zhou, Chengpeng Li, Chengyuan Li, Dayiheng Liu, Fei Huang, Guanting Dong, Haoran Wei, Huan Lin, Jialong Tang, Jialin Wang, Jian Yang, Jianhong Tu, Jianwei Zhang, Jianxin Ma, Jin Xu, Jingren Zhou, Jinze Bai, Jinzheng He, Junyang Lin, Kai Dang, Keming Lu, Keqin Chen, Kexin Yang, Mei Li, Mingfeng Xue, Na~Ni, Pei Zhang, Peng Wang, Ru~Peng, Rui Men, Ruize Gao, Runji Lin, Shijie Wang, Shuai Bai, Sinan Tan, Tianhang Zhu, Tianhao Li, Tianyu Liu, Wenbin Ge, Xiaodong Deng, Xiaohuan Zhou, Xingzhang Ren, Xinyu Zhang, Xipin Wei, Xuancheng Ren, Yang Fan, Yang Yao, Yichang Zhang, Yu~Wan, Yunfei Chu, Yuqiong Liu, Zeyu Cui, Zhenru Zhang, and Zhihao Fan. 2024.
\newblock Qwen2 technical report.
\newblock \emph{arXiv preprint arXiv:2407.10671}.

\bibitem[{Yee(2024)}]{yee:dissertation}
Ming-Ho Yee. 2024.
\newblock \href {https://repository.library.northeastern.edu/files/neu:4f241c784} {\emph{Predicting {typeScript} type annotations and definitions with machine learning}}.
\newblock Ph.D. thesis.

\bibitem[{Yee and Guha(2023)}]{yee:typeweaver}
Ming-Ho Yee and Arjun Guha. 2023.
\newblock Do {Machine} {Learning} {Models} {Produce} {TypeScript} {Types} that {Type} {Check}?
\newblock In \emph{European {Conference} on {Object} {Oriented} {Programming} ({ECOOP})}.

\end{thebibliography}

\clearpage


\appendix
\onecolumn

\section{Use of AI Assistants}

Some of the code for this paper was written with AI assistants enabled.

\section{Model Results}
\label{appendix:model-results}

\begin{figure*}[t]
    \centering
    \includegraphics[width=\textwidth]{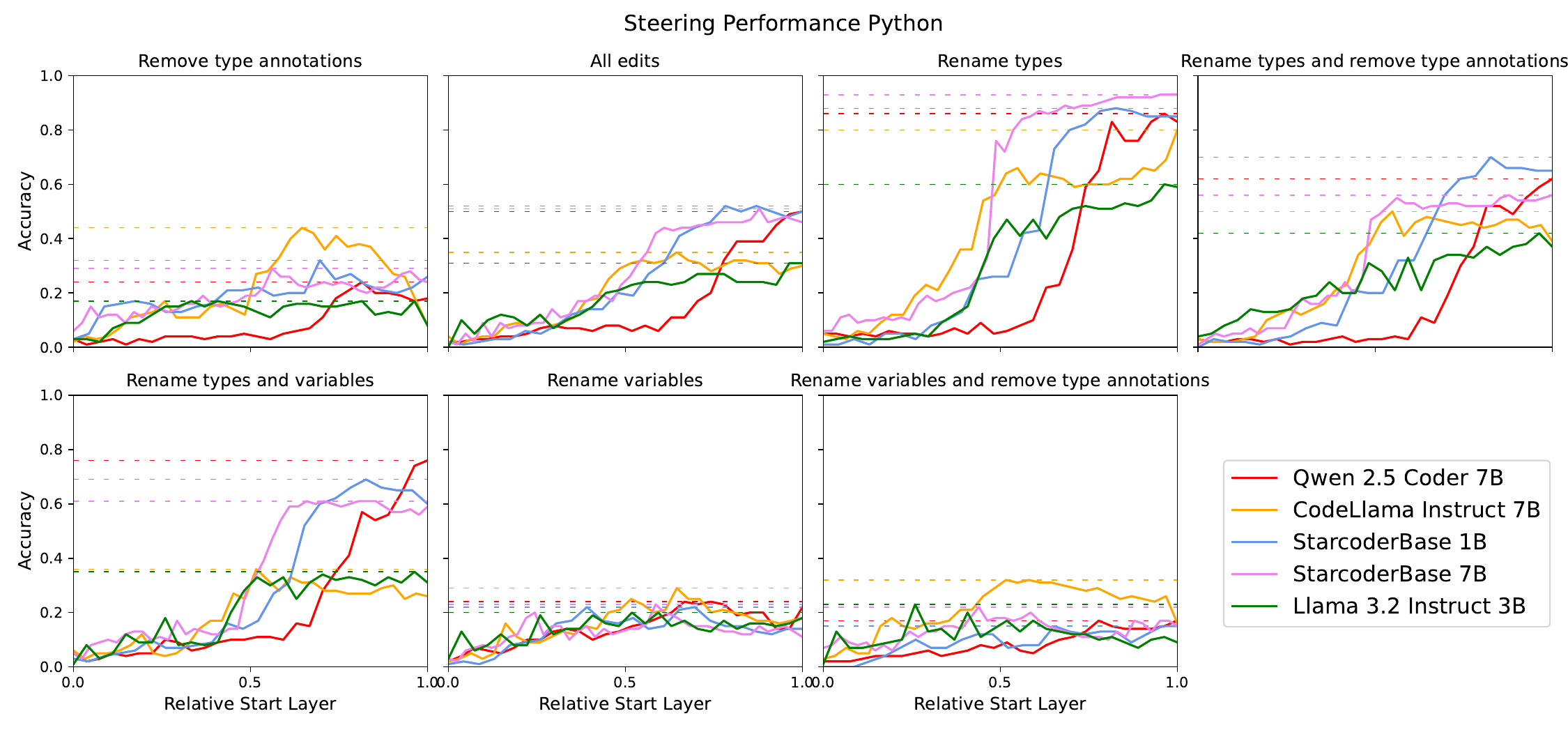}
    \caption{Steering performance for all models on Python data, steering 1 adjacent layers.}
    \label{fig:all_models-lang_py-interval_1}
\end{figure*}

\begin{figure*}
    \centering
    \includegraphics[width=\textwidth]{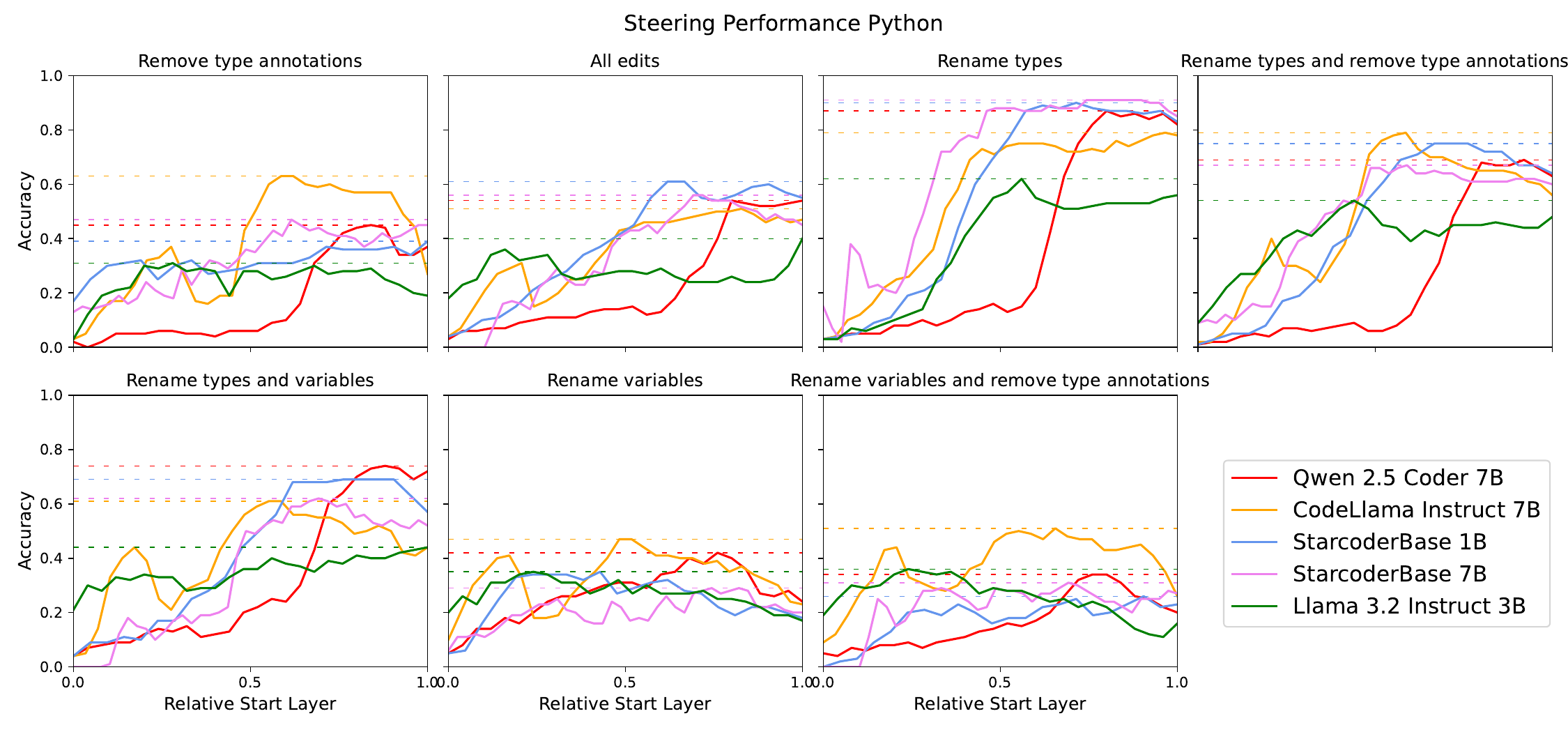}
    \caption{Steering performance for all models on Python data, steering 3 adjacent layers.}
    \label{fig:all_models-lang_py-interval_3}
\end{figure*}

\begin{figure*}
    \centering
    \includegraphics[width=\textwidth]{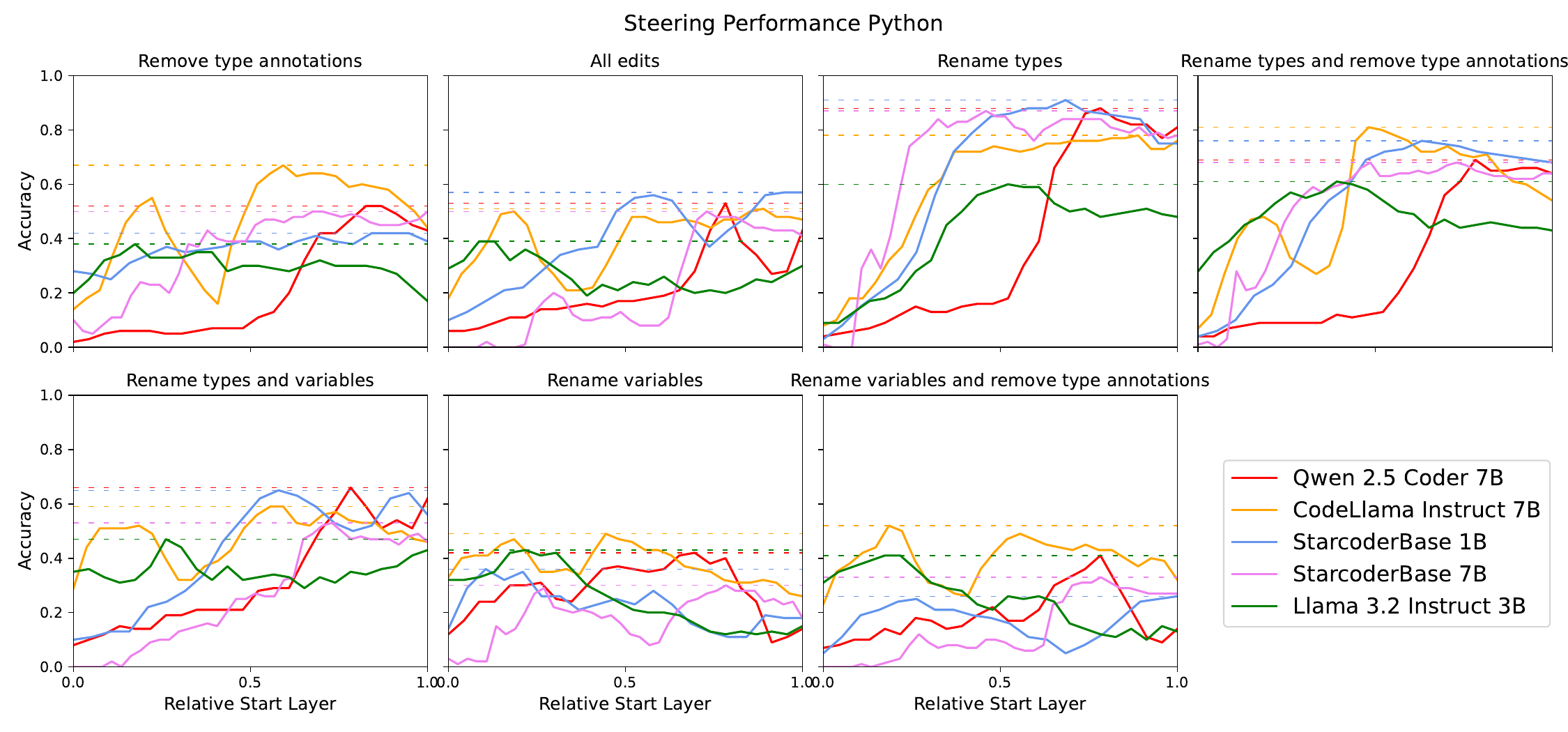}
    \caption{Steering performance for all models on Python data, steering 5 adjacent layers.}
    \label{fig:all_models-lang_py-interval_5}
\end{figure*}

\begin{figure*}
    \centering
    \includegraphics[width=\textwidth]{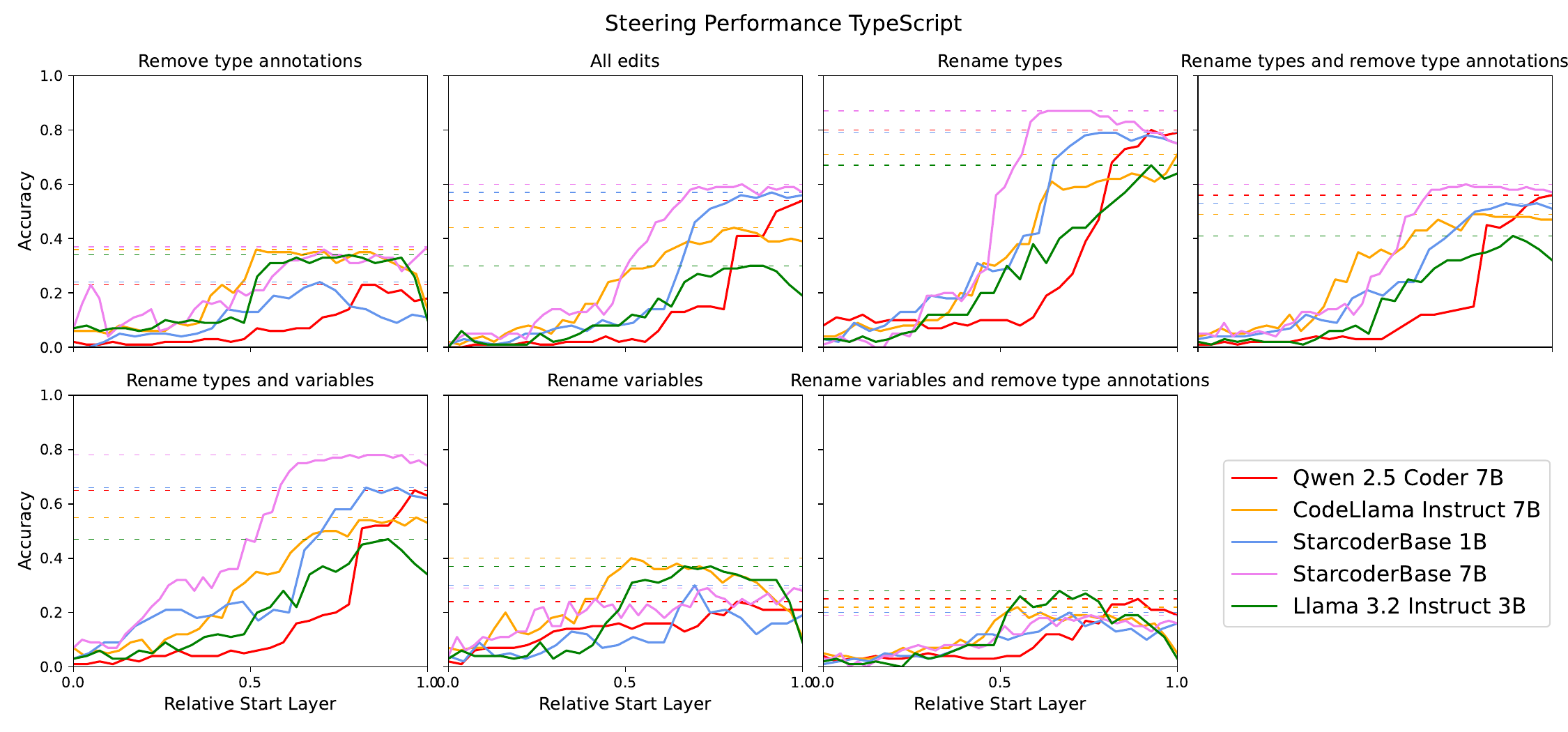}
    \caption{Steering performance for all models on TypeScript data, steering 1 adjacent layers.}
    \label{fig:all_models-lang_ts-interval_1}
\end{figure*}

\begin{figure*}
    \centering
    \includegraphics[width=\textwidth]{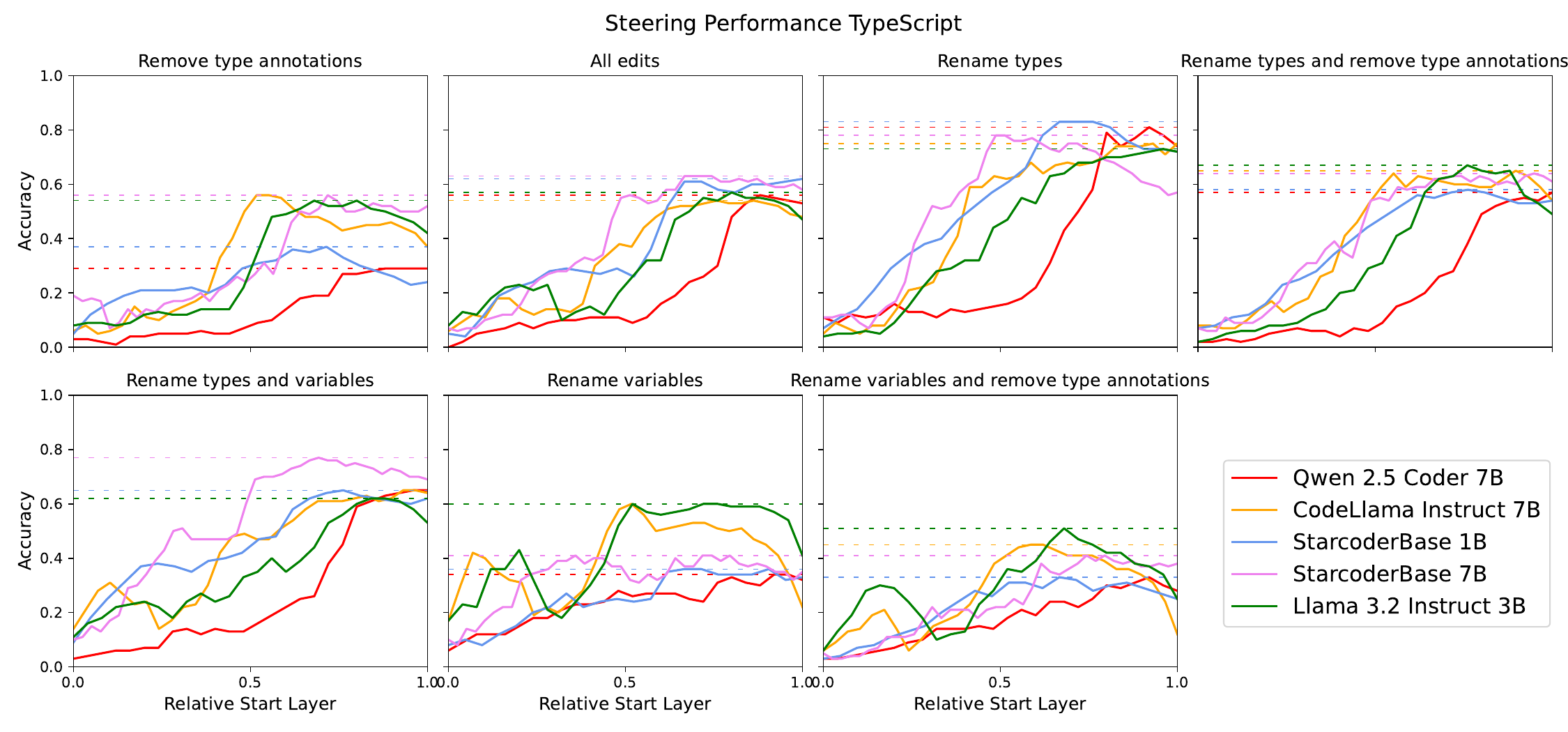}
    \caption{Steering performance for all models on TypeScript data, steering 3 adjacent layers.}
    \label{fig:all_models-lang_ts-interval_3}
\end{figure*}

\begin{figure*}
    \centering
    \includegraphics[width=\textwidth]{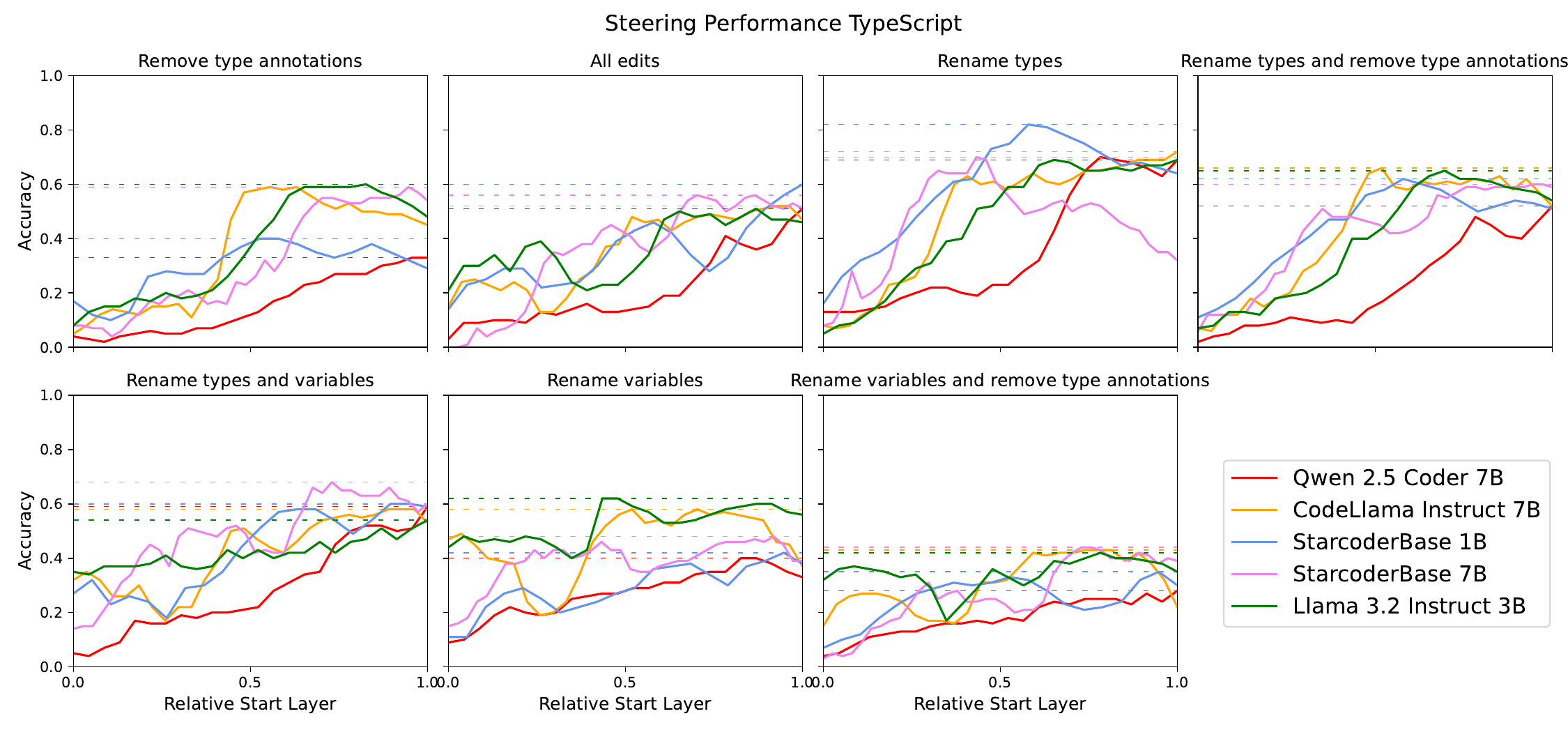}
    \caption{Steering performance for all models on TypeScript data, steering 5 adjacent layers.}
    \label{fig:all_models-lang_ts-interval_5}
\end{figure*}

\clearpage

\section{Interval Ablations Results}
\label{appendix:interval-results}

\begin{figure*}
    \centering
    \includegraphics[width=\textwidth]{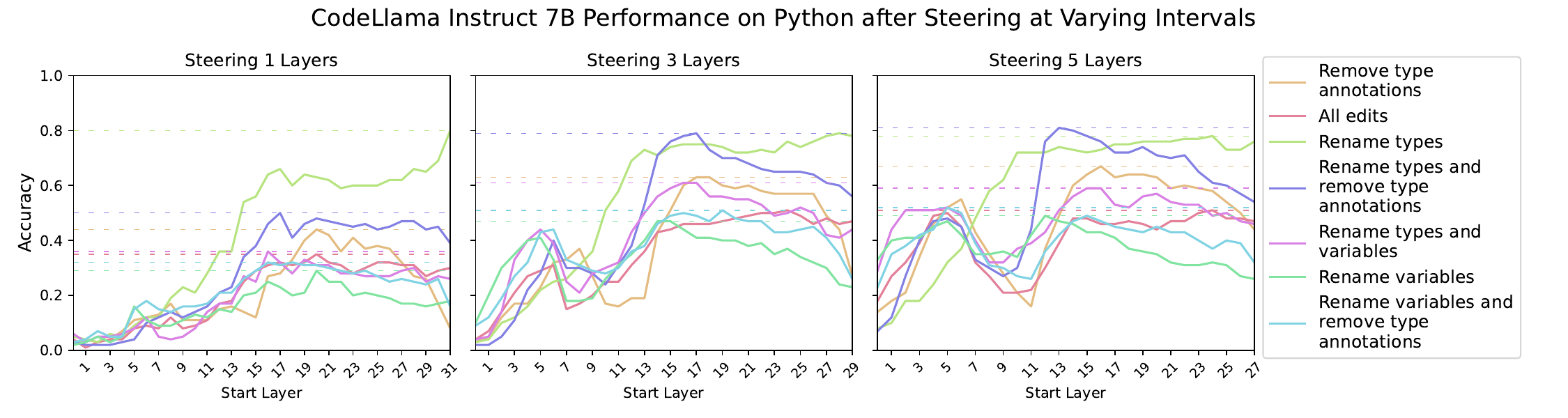}
    \caption{Steering CodeLlama Instruct 7B across different layer intervals}
    \label{fig:intervals-CodeLlama-7b-Instruct-hf-py}
\end{figure*}

\begin{figure*}
    \centering
    \includegraphics[width=\textwidth]{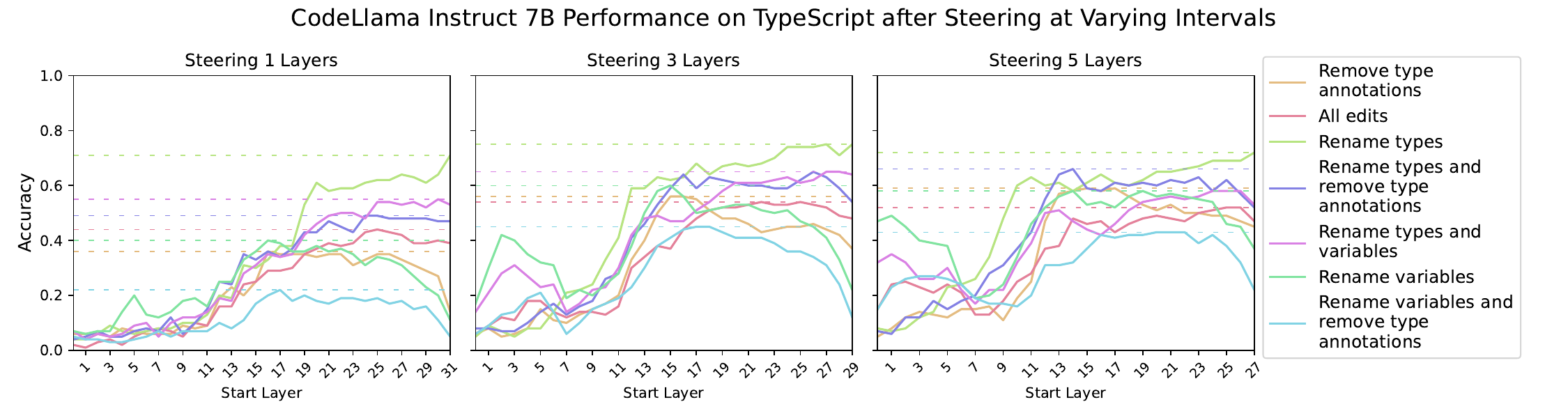}
    \caption{Steering CodeLlama Instruct 7B across different layer intervals}
    \label{fig:intervals-CodeLlama-7b-Instruct-hf-ts}
\end{figure*}

\begin{figure*}
    \centering
    \includegraphics[width=\textwidth]{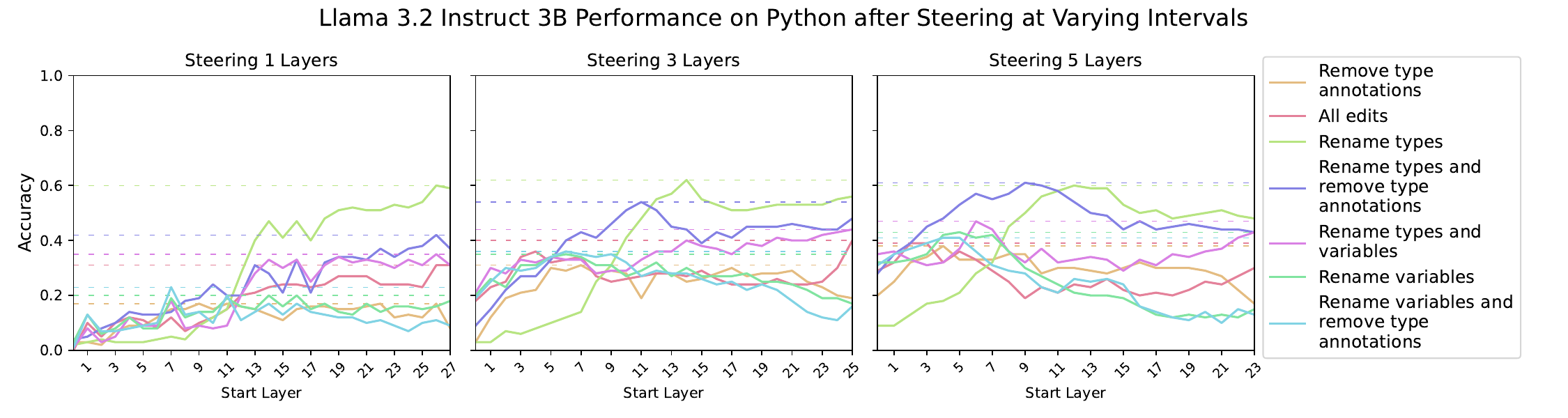}
    \caption{Steering Llama 3.2 Instruct 3B across different layer intervals}
    \label{fig:intervals-Llama-3.2-3B-Instruct-py}
\end{figure*}

\begin{figure*}
    \centering
    \includegraphics[width=\textwidth]{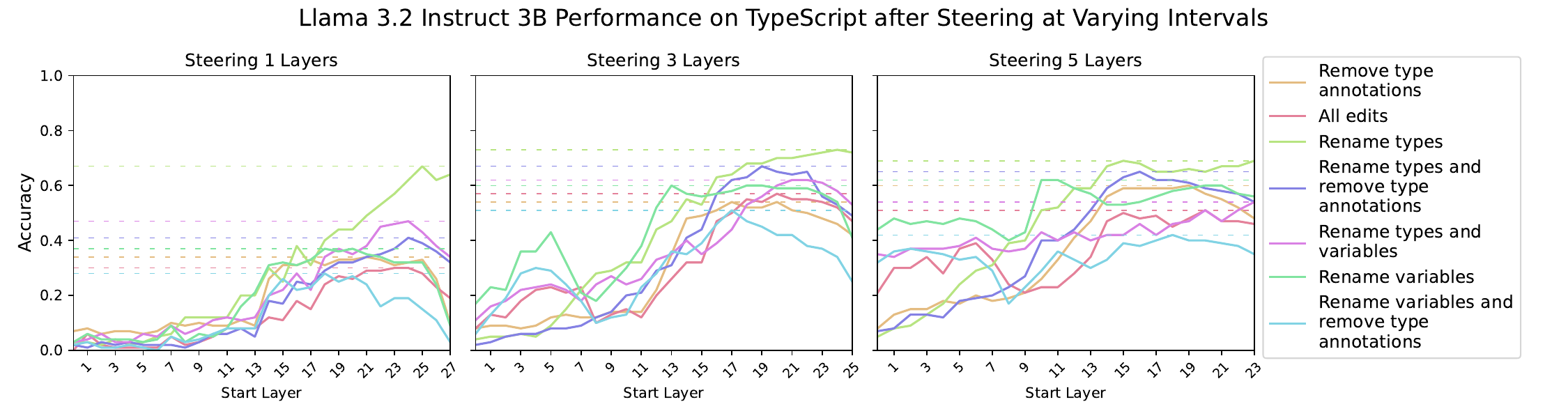}
    \caption{Steering Llama 3.2 Instruct 3B across different layer intervals}
    \label{fig:intervals-Llama-3.2-3B-Instruct-ts}
\end{figure*}

\begin{figure*}
    \centering
    \includegraphics[width=\textwidth]{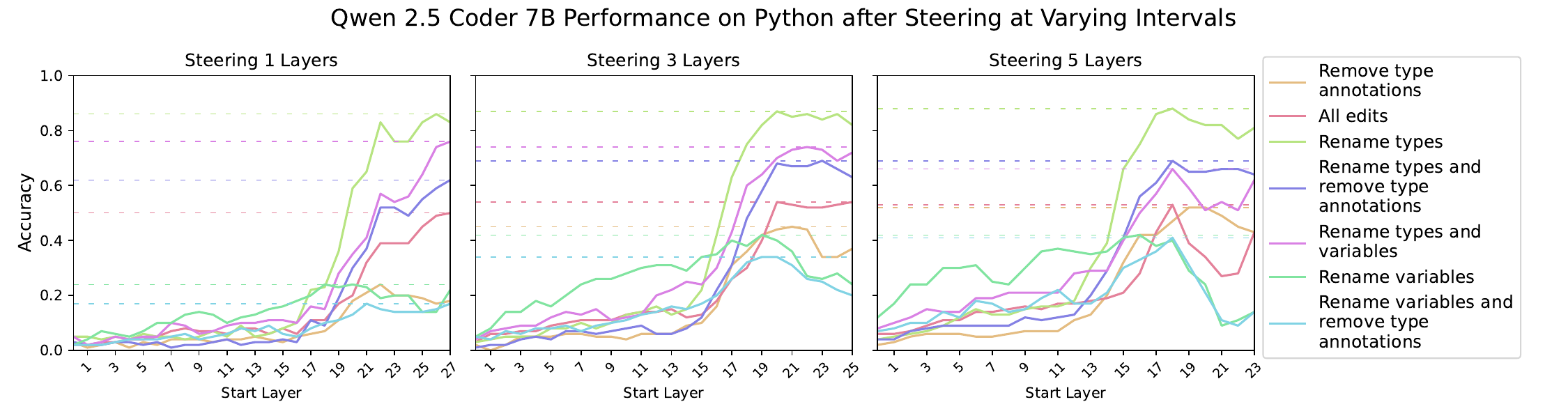}
    \caption{Steering Qwen 2.5 Coder 7B across different layer intervals}
    \label{fig:intervals-qwen2p5_coder_7b_base-py}
\end{figure*}

\begin{figure*}
    \centering
    \includegraphics[width=\textwidth]{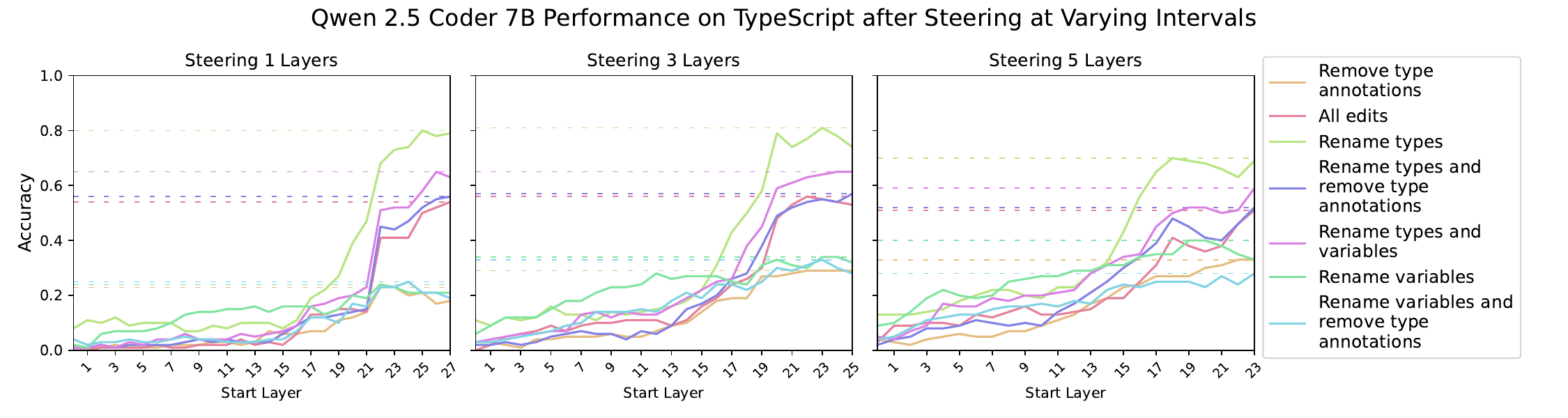}
    \caption{Steering Qwen 2.5 Coder 7B across different layer intervals}
    \label{fig:intervals-qwen2p5_coder_7b_base-ts}
\end{figure*}

\begin{figure*}
    \centering
    \includegraphics[width=\textwidth]{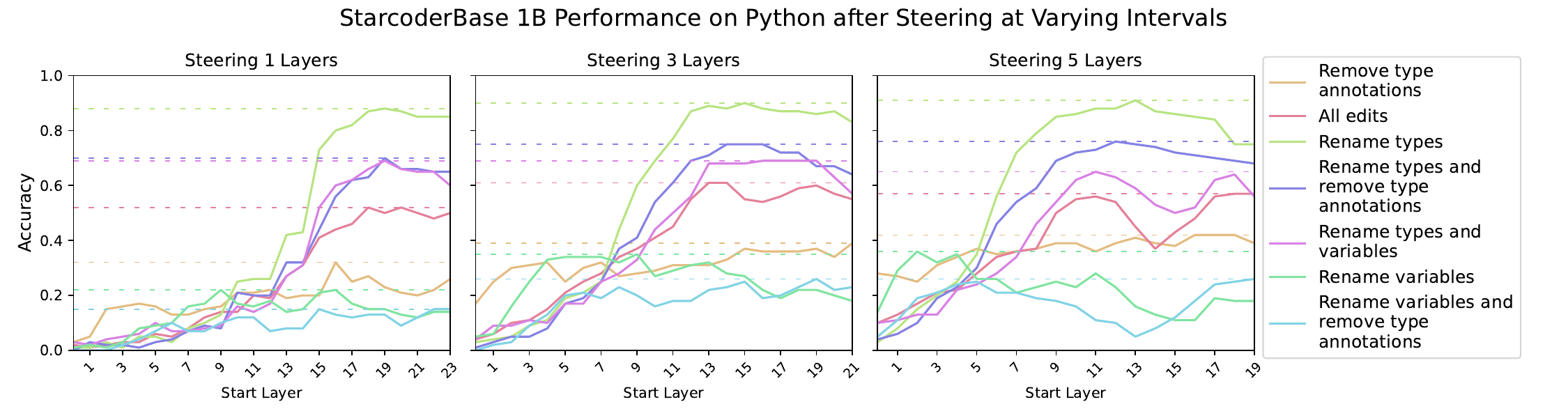}
    \caption{Steering StarcoderBase 1B across different layer intervals}
    \label{fig:intervals-starcoderbase-1b-py}
\end{figure*}

\begin{figure*}
    \centering
    \includegraphics[width=\textwidth]{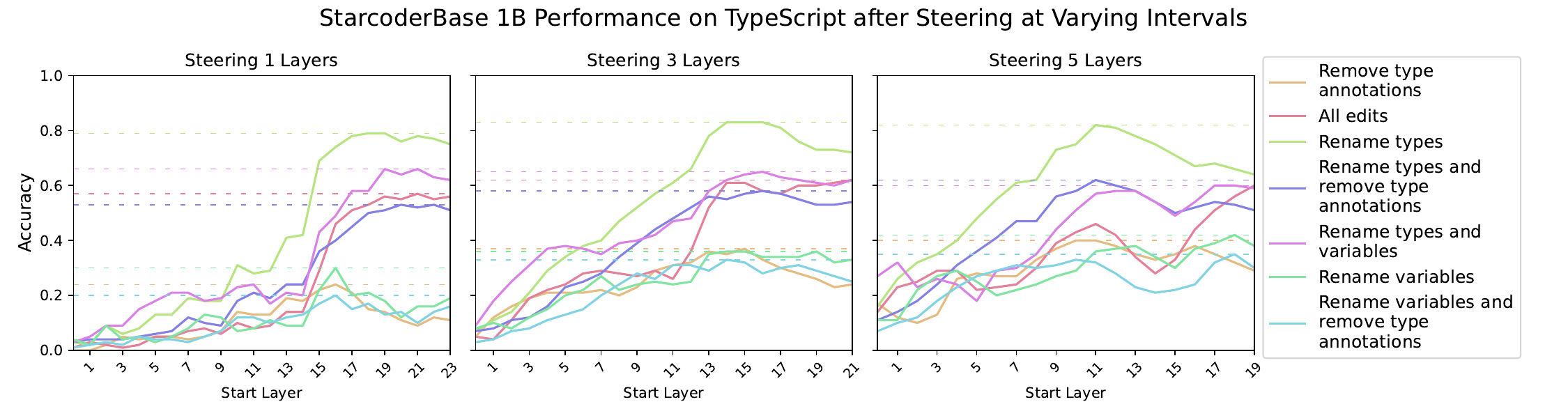}
    \caption{Steering StarcoderBase 1B across different layer intervals}
    \label{fig:intervals-starcoderbase-1b-ts}
\end{figure*}

\begin{figure*}
    \centering
    \includegraphics[width=\textwidth]{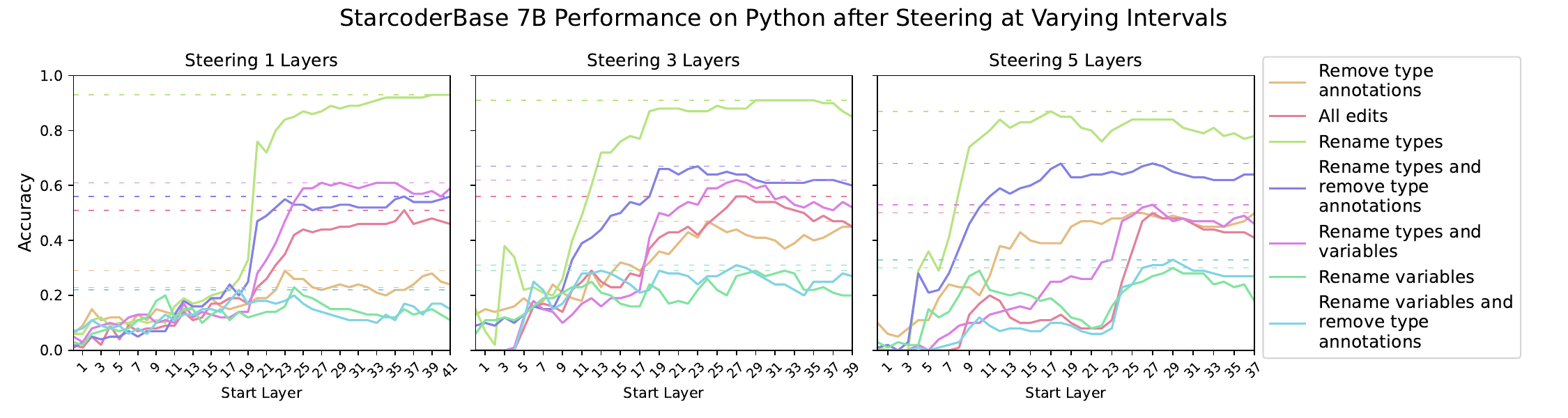}
    \caption{Steering StarcoderBase 7B across different layer intervals}
    \label{fig:intervals-starcoderbase-7b-py}
\end{figure*}

\begin{figure*}
    \centering
    \includegraphics[width=\textwidth]{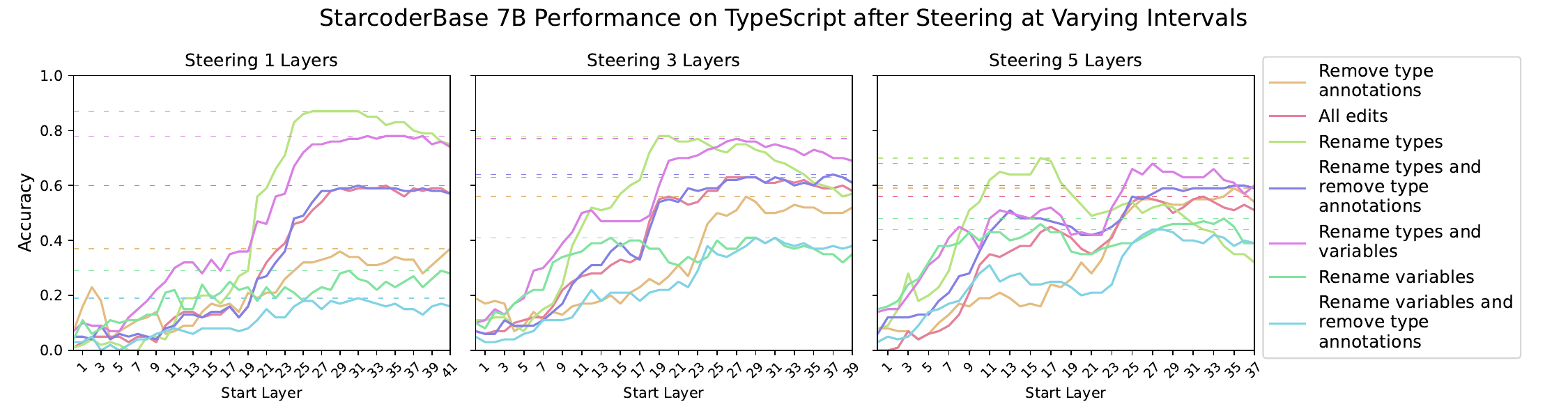}
    \caption{Steering StarcoderBase 7B across different layer intervals}
    \label{fig:intervals-starcoderbase-7b-ts}
\end{figure*}

\clearpage
\section{Language Transfer Results}
\label{appendix:lang-transfer}

\begin{figure*}
    \centering
    \includegraphics[width=\textwidth]{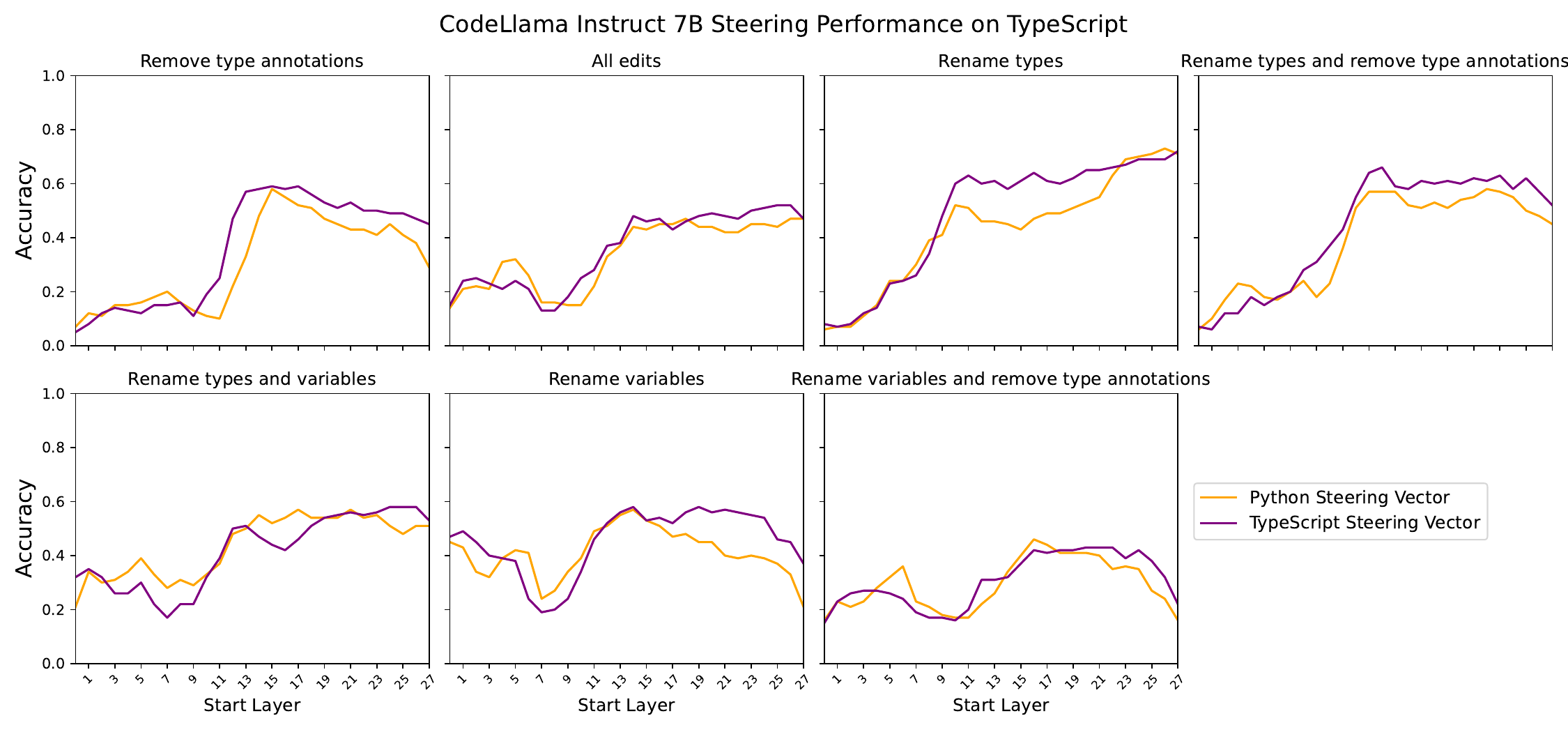}
    \caption{Steering performance for CodeLlama Instruct 7B on TypeScript test set using TypeScript and Python steering vectors. We steer 5 adjacent layers.}
    \label{fig:lang_transfer-CodeLlama-7b-Instruct-hf-py_onto_ts-interval_5}
\end{figure*}

\begin{figure*}
    \centering
    \includegraphics[width=\textwidth]{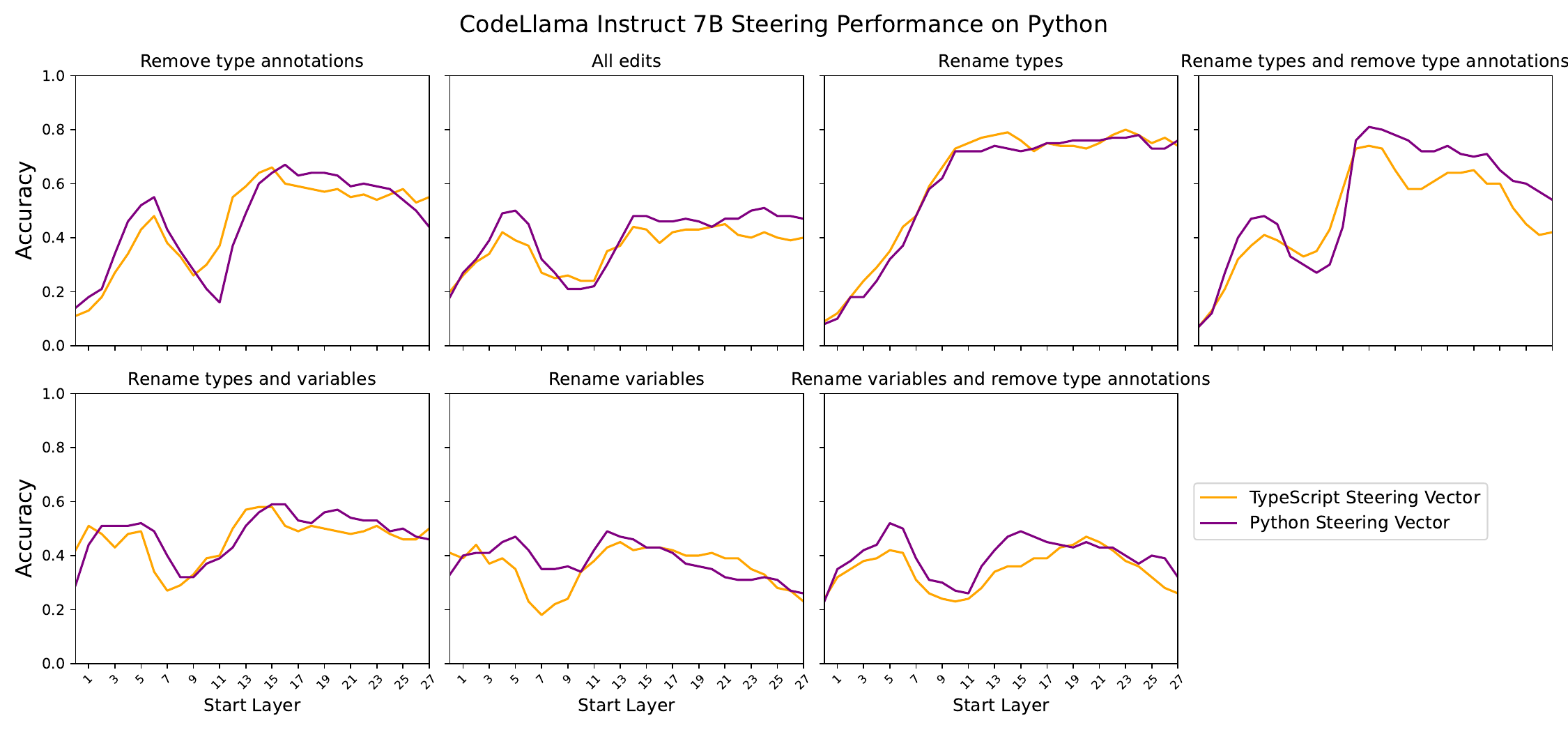}
    \caption{Steering performance for CodeLlama Instruct 7B on Python test set using Python and TypeScript steering vectors. We steer 5 adjacent layers.}
    \label{fig:lang_transfer-CodeLlama-7b-Instruct-hf-ts_onto_py-interval_5}
\end{figure*}

\begin{figure*}
    \centering
    \includegraphics[width=\textwidth]{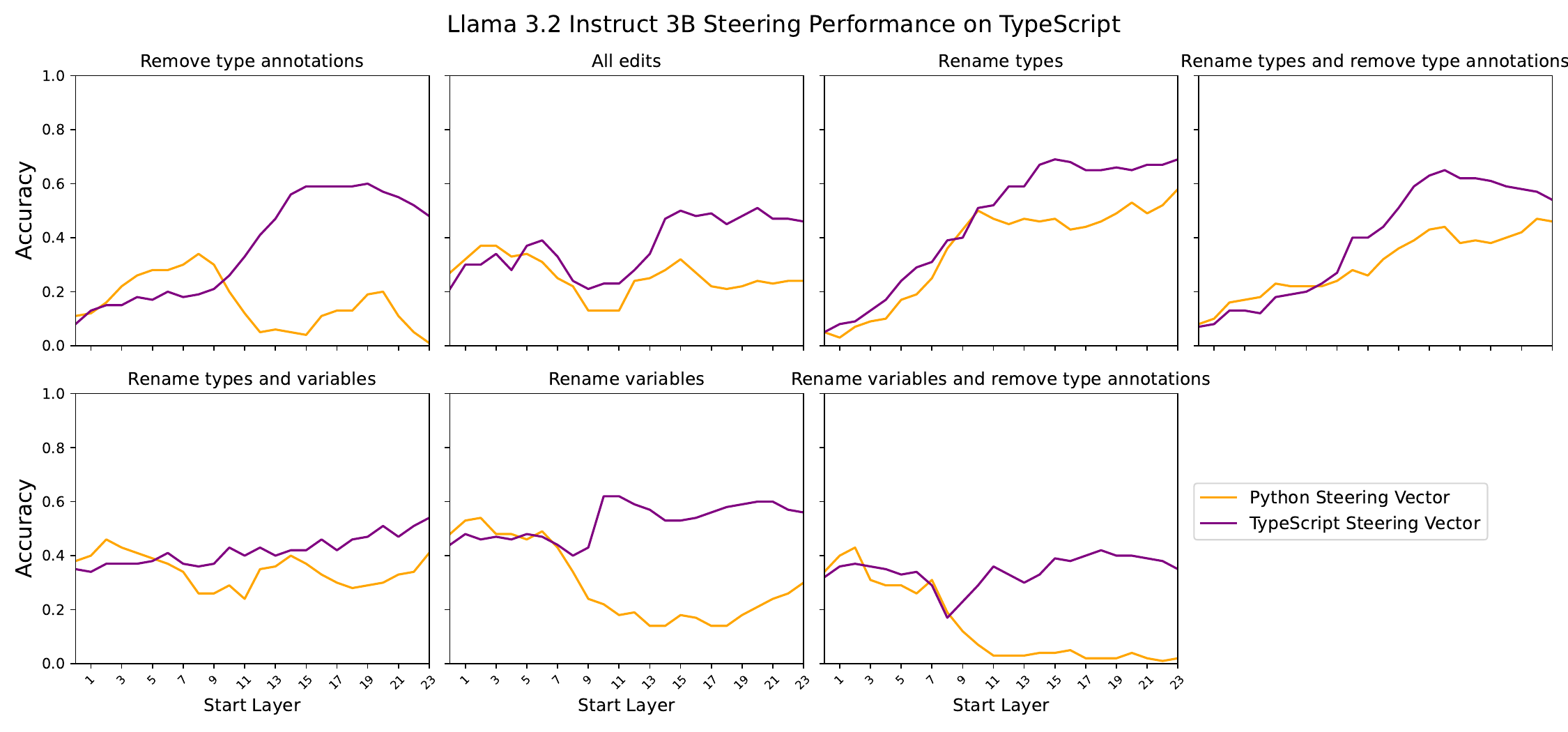}
    \caption{Steering performance for Llama 3.2 Instruct 3B on TypeScript test set using TypeScript and Python steering vectors. We steer 5 adjacent layers.}
    \label{fig:lang_transfer-Llama-3.2-3B-Instruct-py_onto_ts-interval_5}
\end{figure*}

\begin{figure*}
    \centering
    \includegraphics[width=\textwidth]{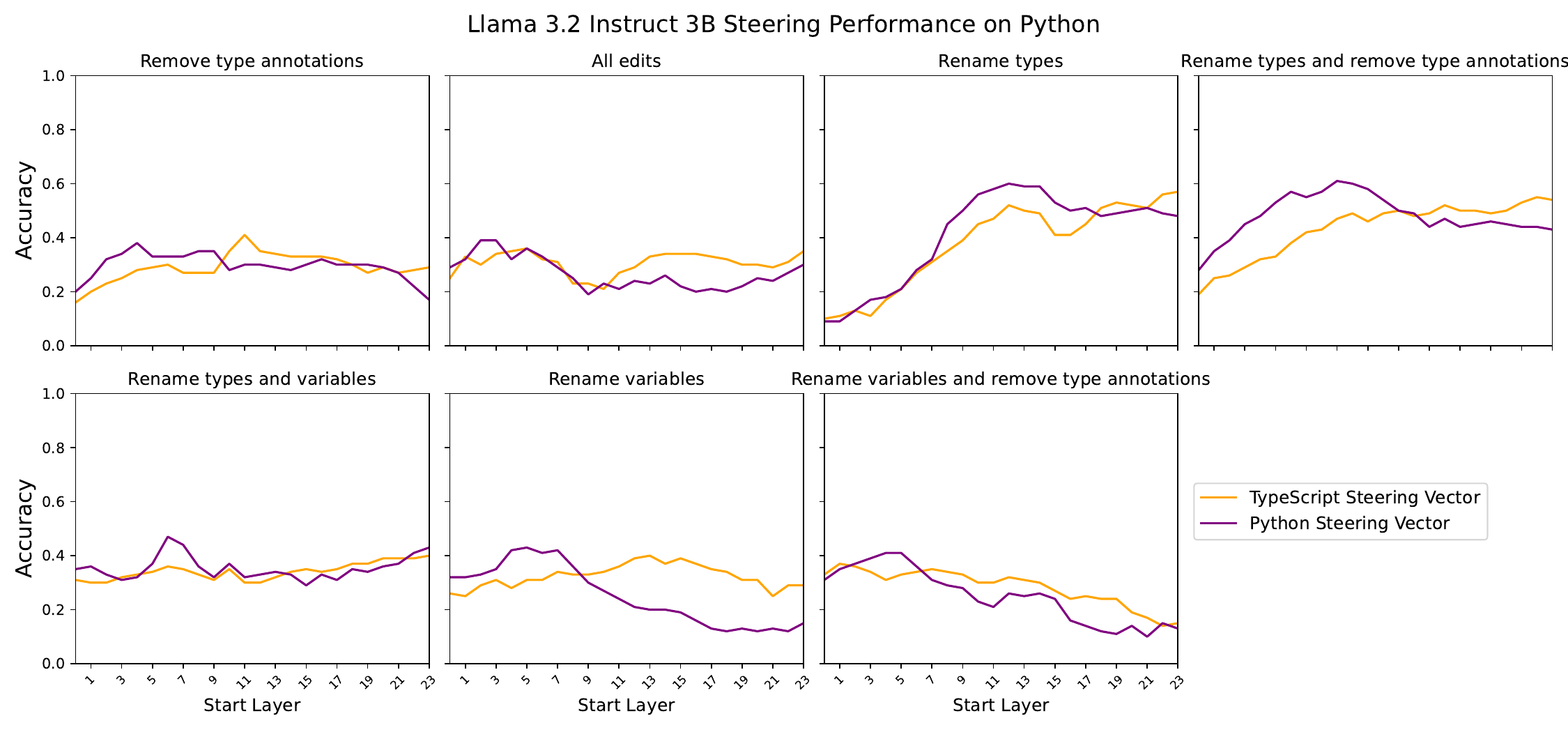}
    \caption{Steering performance for Llama 3.2 Instruct 3B on Python test set using Python and TypeScript steering vectors. We steer 5 adjacent layers.}
    \label{fig:lang_transfer-Llama-3.2-3B-Instruct-ts_onto_py-interval_5}
\end{figure*}

\begin{figure*}
    \centering
    \includegraphics[width=\textwidth]{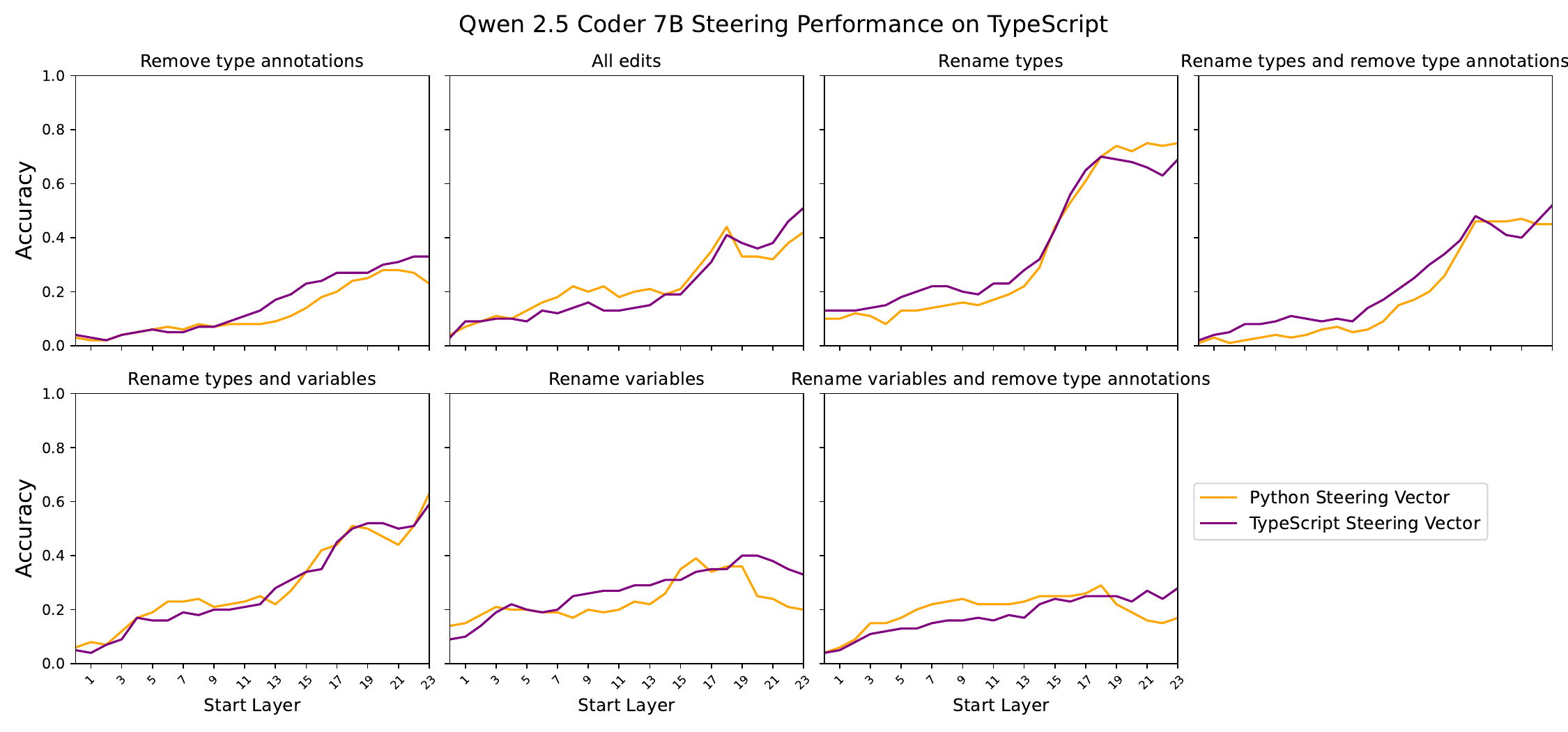}
    \caption{Steering performance for Qwen 2.5 Coder 7B on TypeScript test set using TypeScript and Python steering vectors. We steer 5 adjacent layers.}
    \label{fig:lang_transfer-qwen2p5_coder_7b_base-py_onto_ts-interval_5}
\end{figure*}

\begin{figure*}
    \centering
    \includegraphics[width=\textwidth]{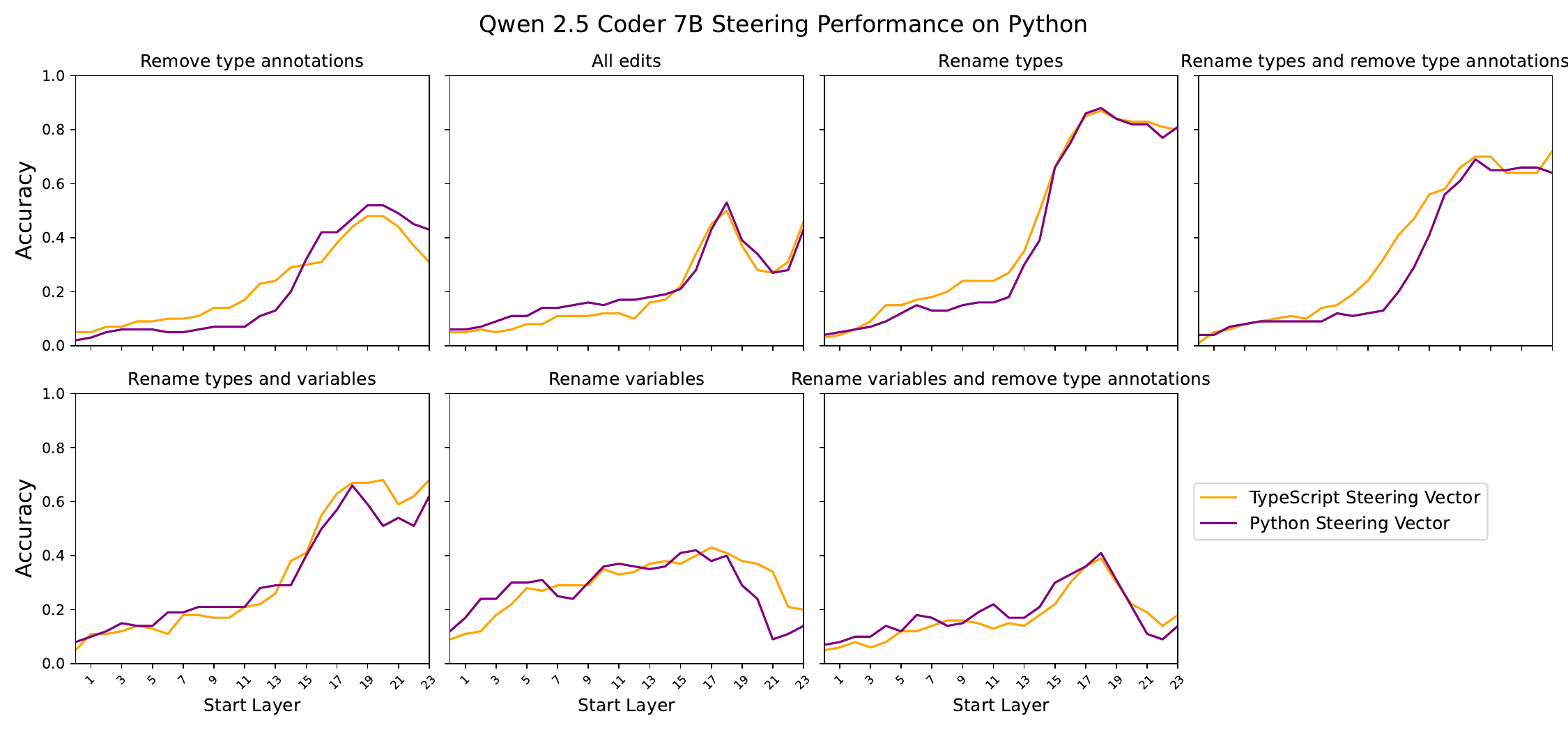}
    \caption{Steering performance for Qwen 2.5 Coder 7B on Python test set using Python and TypeScript steering vectors. We steer 5 adjacent layers.}
    \label{fig:lang_transfer-qwen2p5_coder_7b_base-ts_onto_py-interval_5}
\end{figure*}

\begin{figure*}
    \centering
    \includegraphics[width=\textwidth]{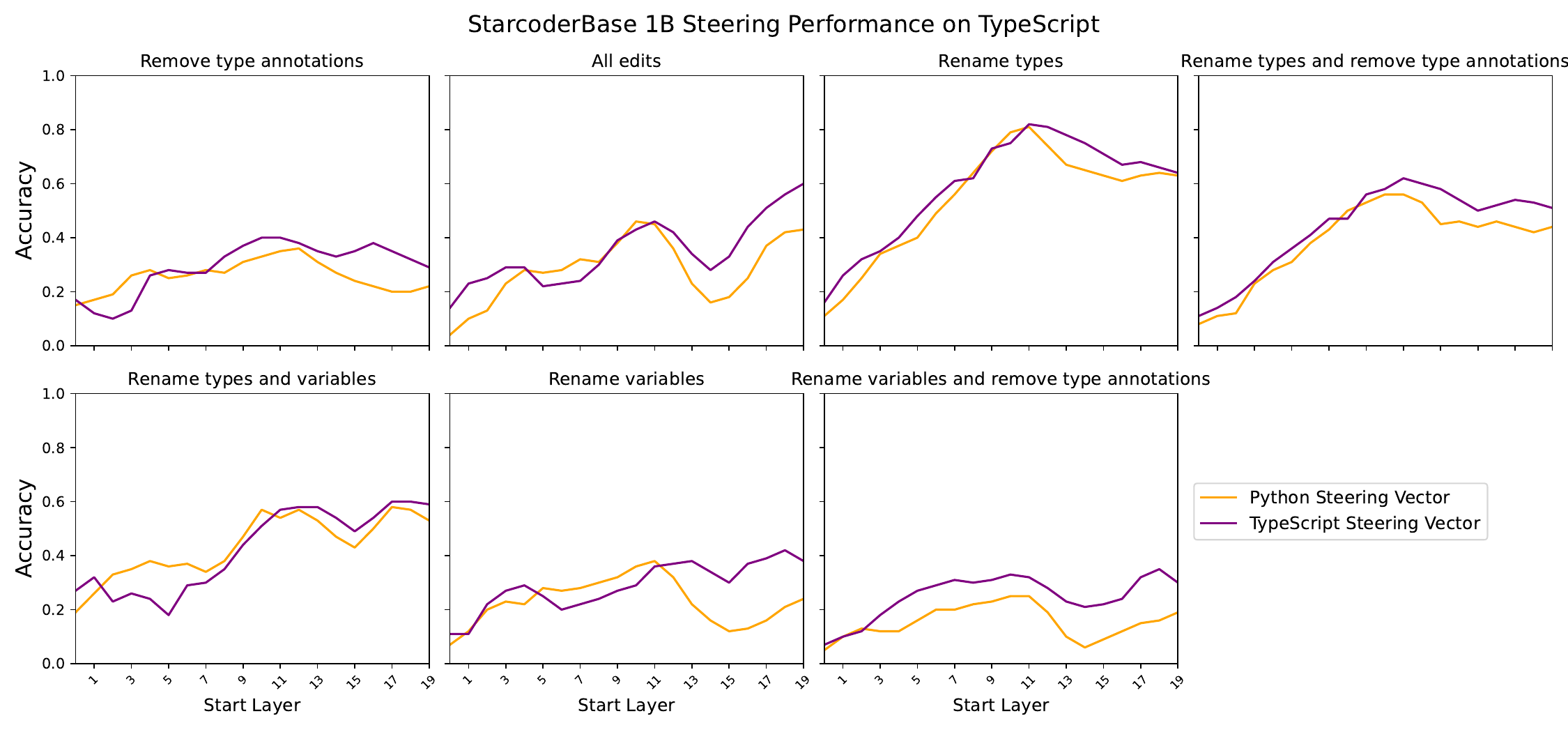}
    \caption{Steering performance for StarcoderBase 1B on TypeScript test set using TypeScript and Python steering vectors. We steer 5 adjacent layers.}
    \label{fig:lang_transfer-starcoderbase-1b-py_onto_ts-interval_5}
\end{figure*}

\begin{figure*}
    \centering
    \includegraphics[width=\textwidth]{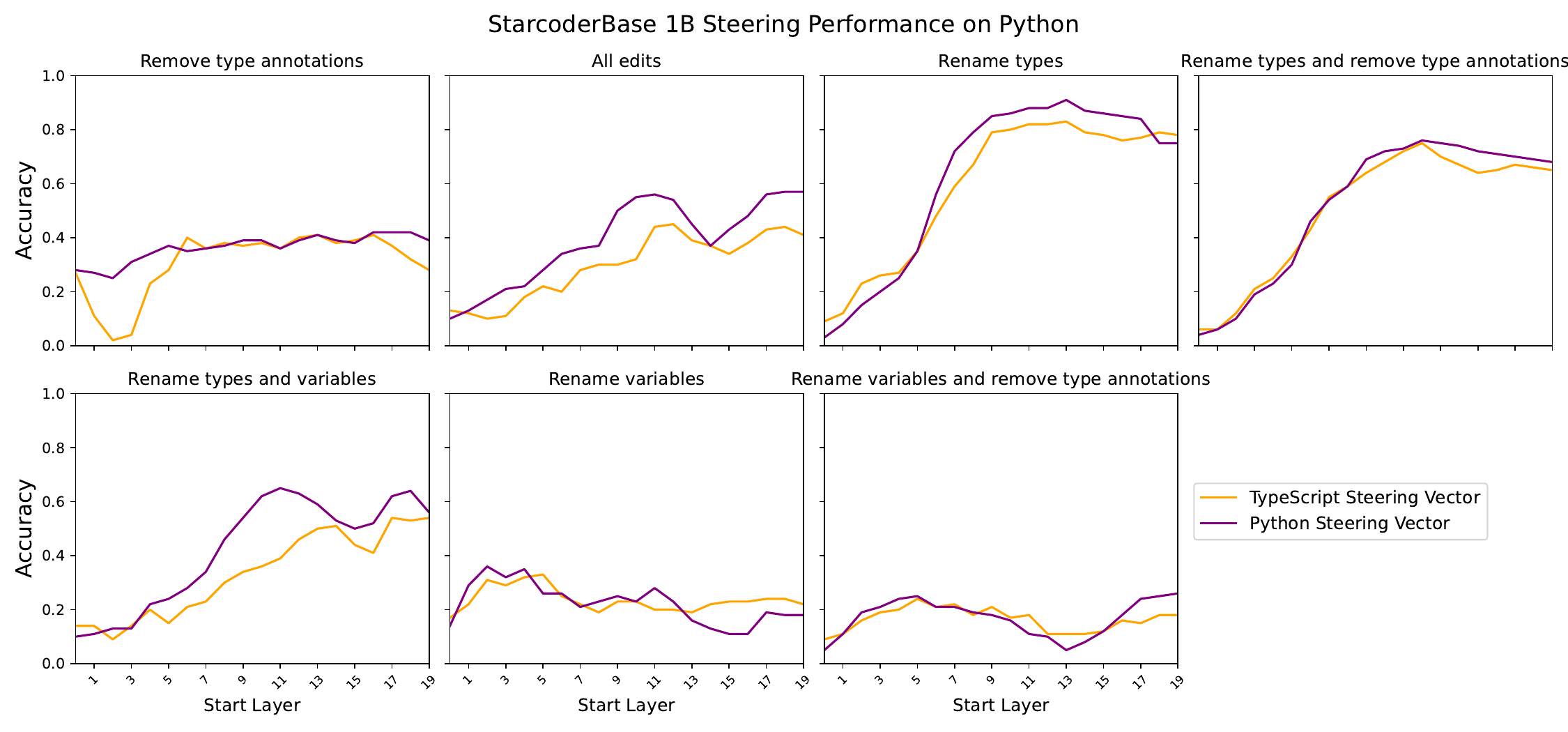}
    \caption{Steering performance for StarcoderBase 1B on Python test set using Python and TypeScript steering vectors. We steer 5 adjacent layers.}
    \label{fig:lang_transfer-starcoderbase-1b-ts_onto_py-interval_5}
\end{figure*}

\begin{figure*}
    \centering
    \includegraphics[width=\textwidth]{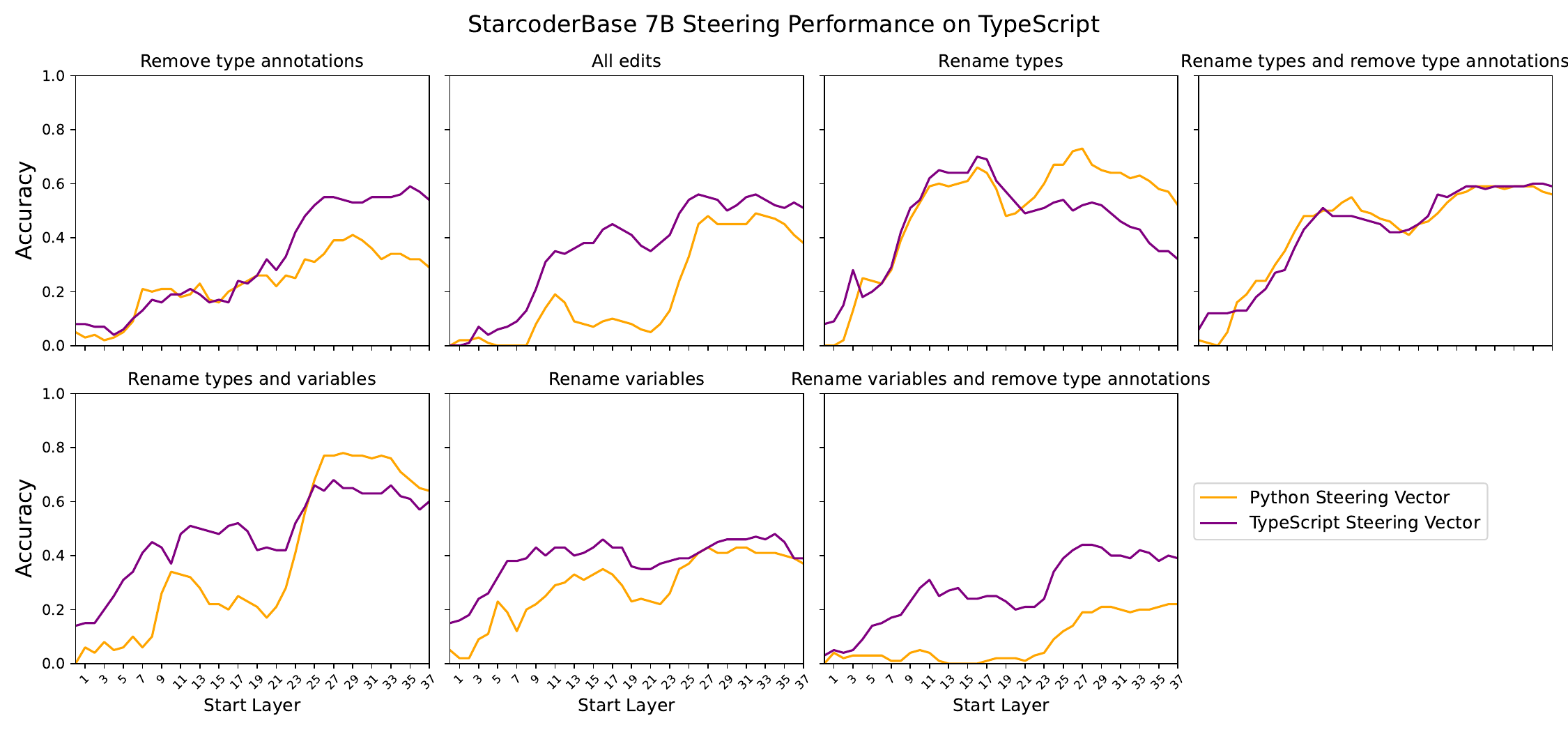}
    \caption{Steering performance for StarcoderBase 7B on TypeScript test set using TypeScript and Python steering vectors. We steer 5 adjacent layers.}
    \label{fig:lang_transfer-starcoderbase-7b-py_onto_ts-interval_5}
\end{figure*}

\begin{figure*}
    \centering
    \includegraphics[width=\textwidth]{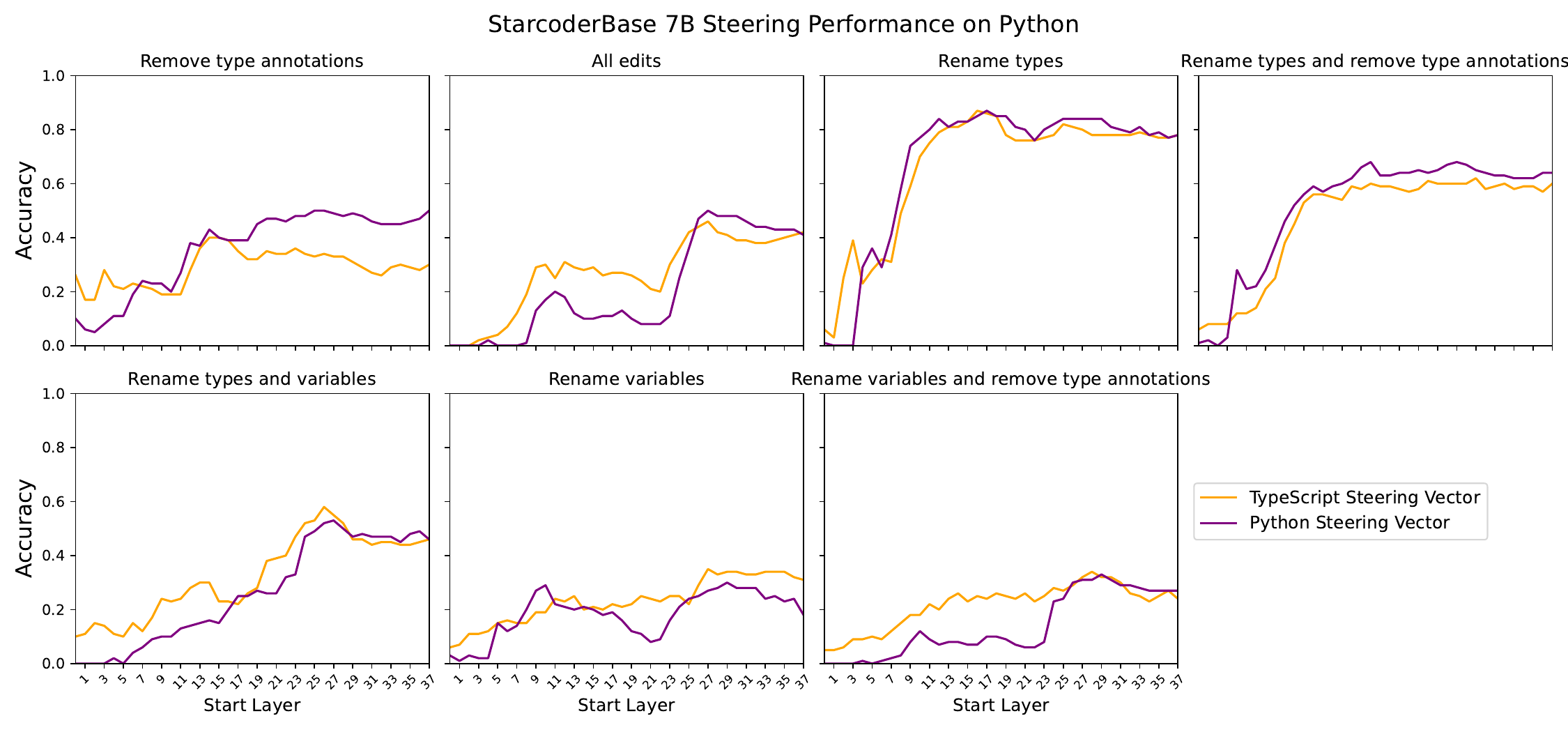}
    \caption{Steering performance for StarcoderBase 7B on Python test set using Python and TypeScript steering vectors. We steer 5 adjacent layers.}
    \label{fig:lang_transfer-starcoderbase-7b-ts_onto_py-interval_5}
\end{figure*}
\clearpage

\section{Comparing Steering Against Baselines}
\label{appendix:splits}

\begin{figure*}
    \centering
    \includegraphics[width=\textwidth]{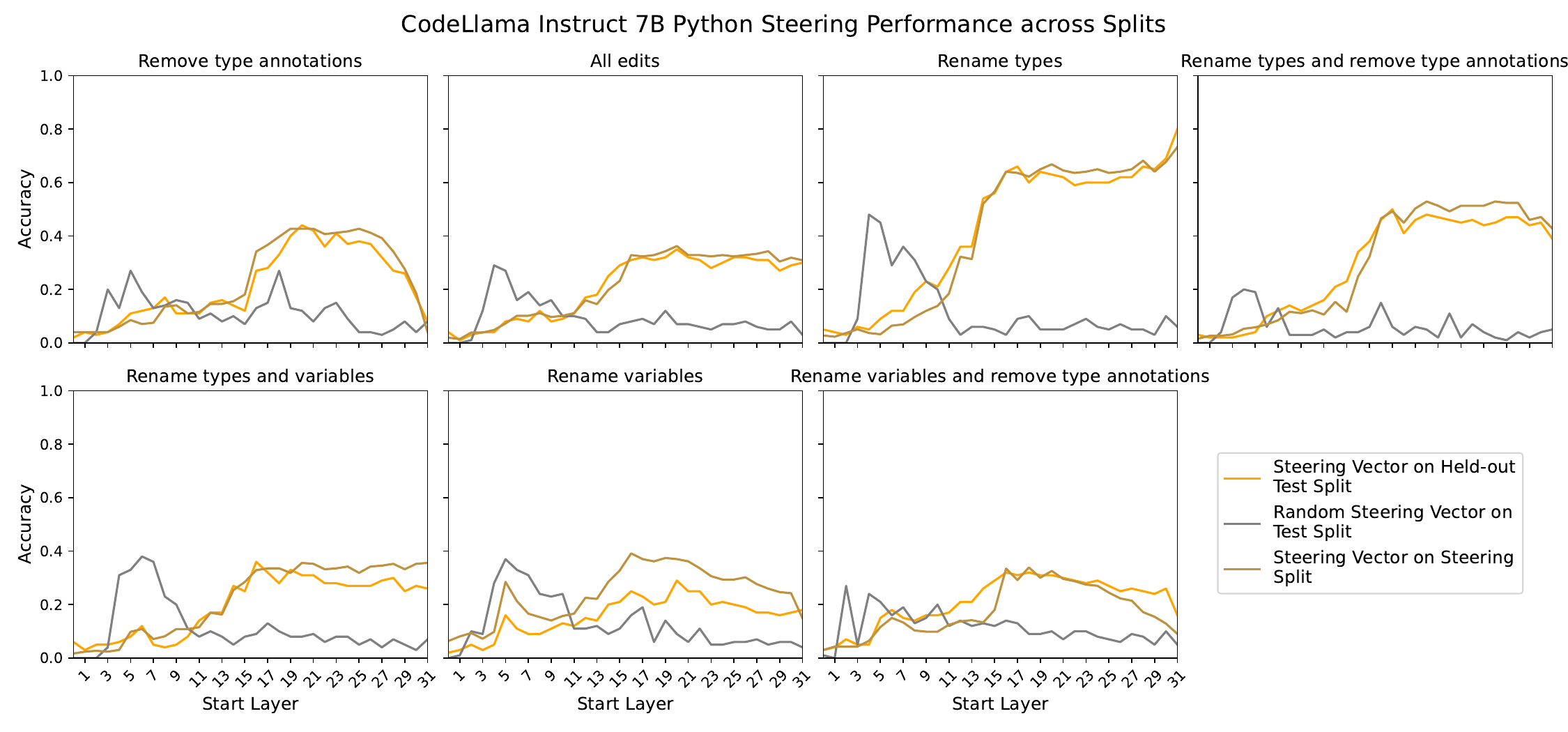}
    \caption{Python steering performance for CodeLlama Instruct 7B on test and steering datasets, compared against a random steering vector baseline. We steer 1 adjacent layers.}
    \label{fig:splits-CodeLlama-7b-Instruct-hf-py-1}
\end{figure*}

\begin{figure*}
    \centering
    \includegraphics[width=\textwidth]{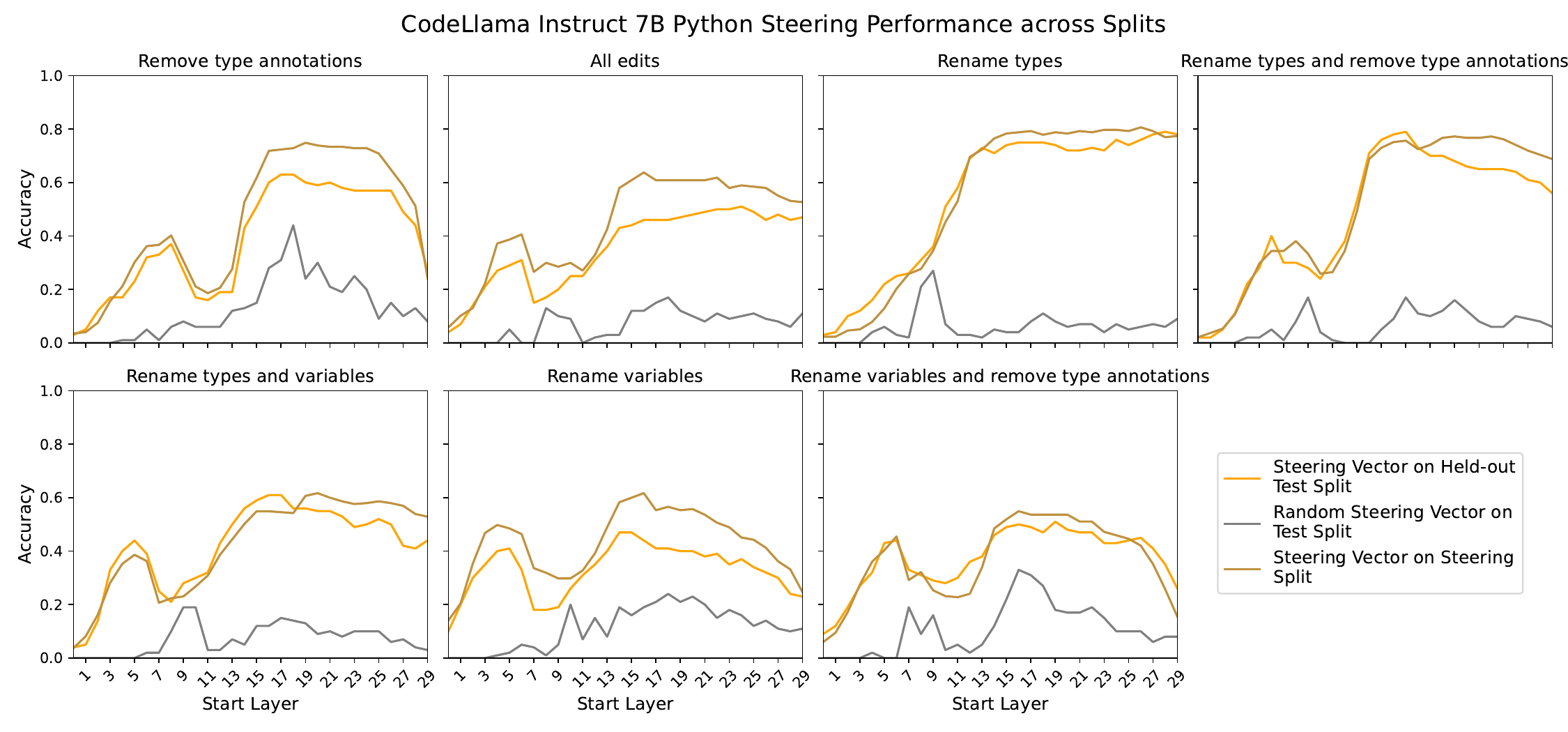}
    \caption{Python steering performance for CodeLlama Instruct 7B on test and steering datasets, compared against a random steering vector baseline. We steer 3 adjacent layers.}
    \label{fig:splits-CodeLlama-7b-Instruct-hf-py-3}
\end{figure*}

\begin{figure*}
    \centering
    \includegraphics[width=\textwidth]{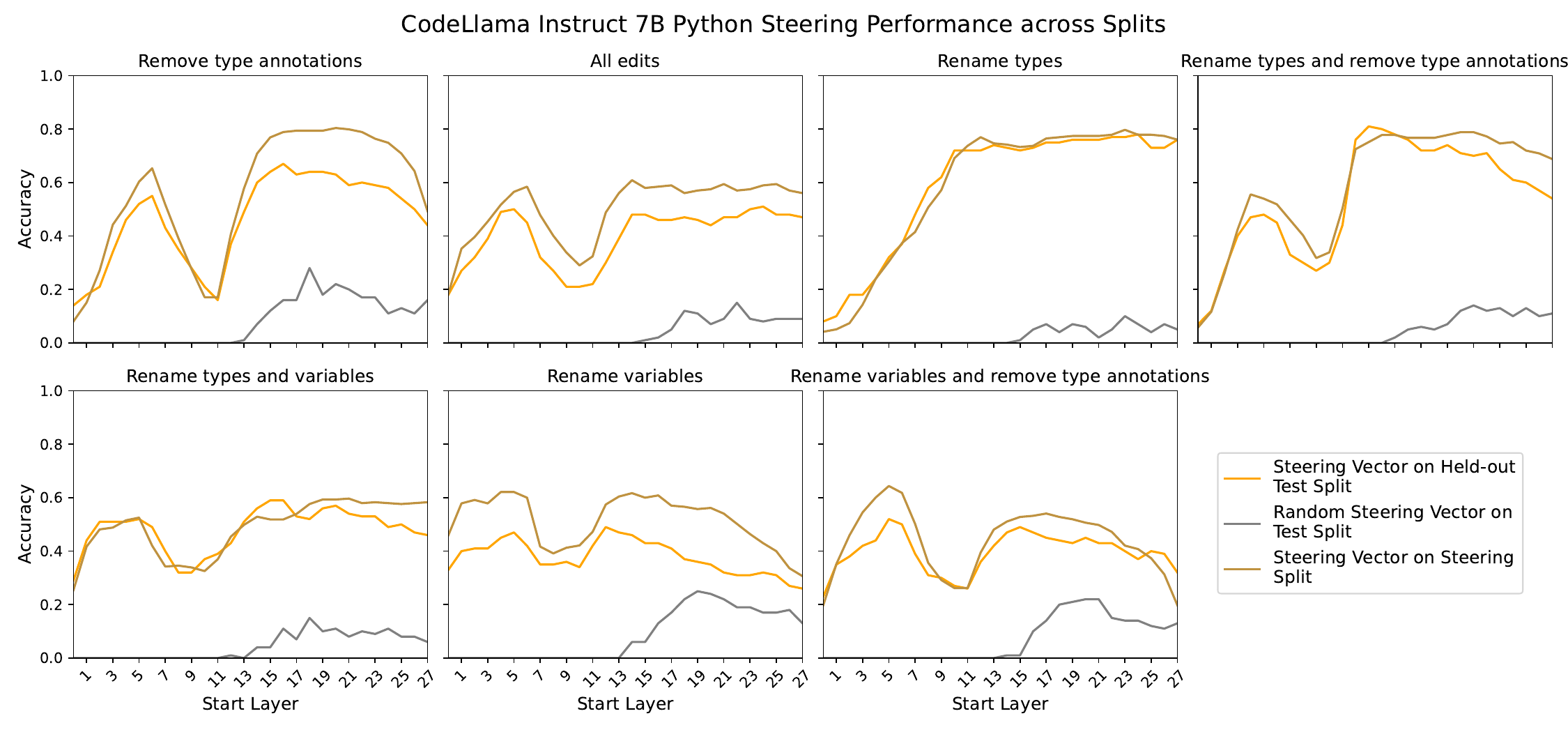}
    \caption{Python steering performance for CodeLlama Instruct 7B on test and steering datasets, compared against a random steering vector baseline. We steer 5 adjacent layers.}
    \label{fig:splits-CodeLlama-7b-Instruct-hf-py-5}
\end{figure*}

\begin{figure*}
    \centering
    \includegraphics[width=\textwidth]{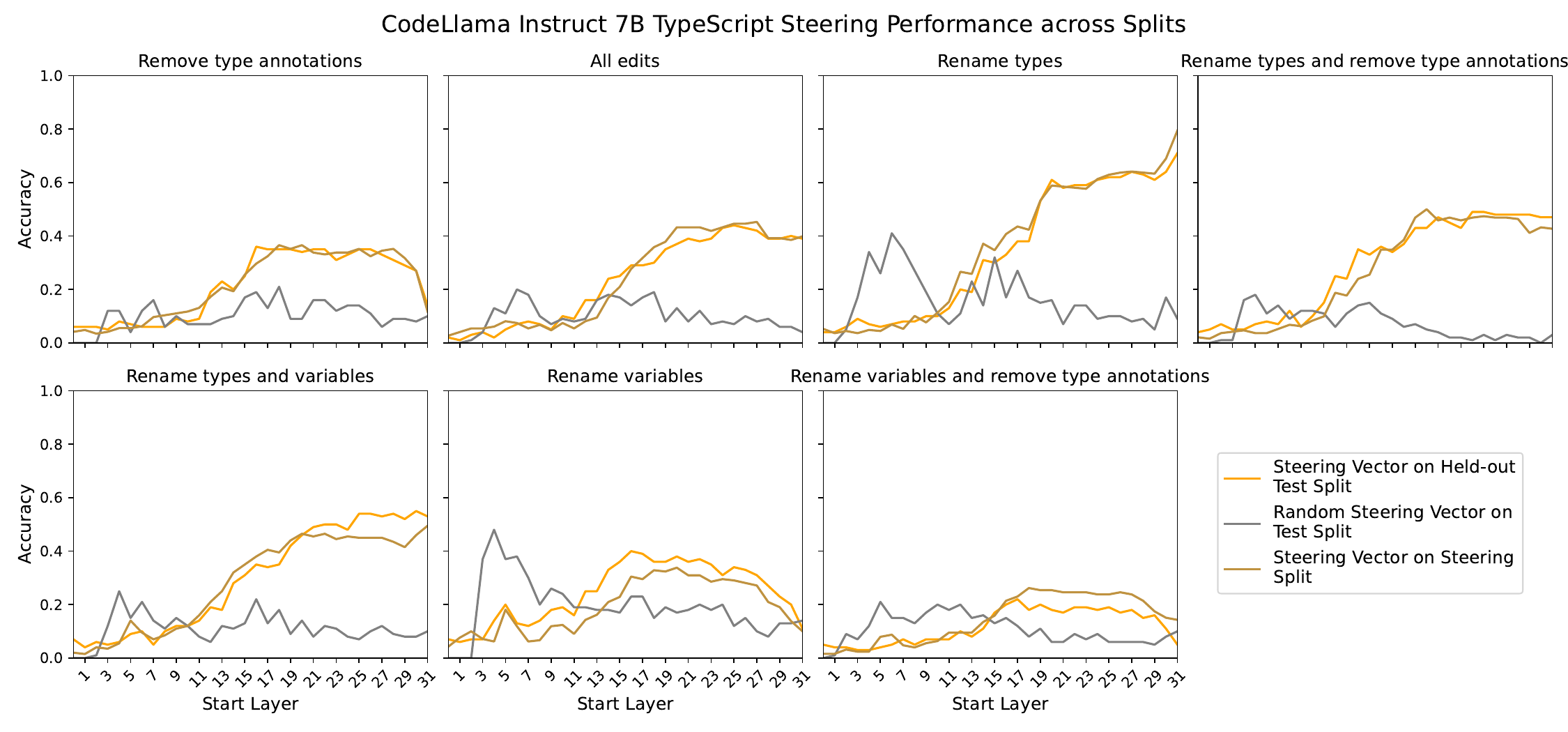}
    \caption{TypeScript steering performance for CodeLlama Instruct 7B on test and steering datasets, compared against a random steering vector baseline. We steer 1 adjacent layers.}
    \label{fig:splits-CodeLlama-7b-Instruct-hf-ts-1}
\end{figure*}

\begin{figure*}
    \centering
    \includegraphics[width=\textwidth]{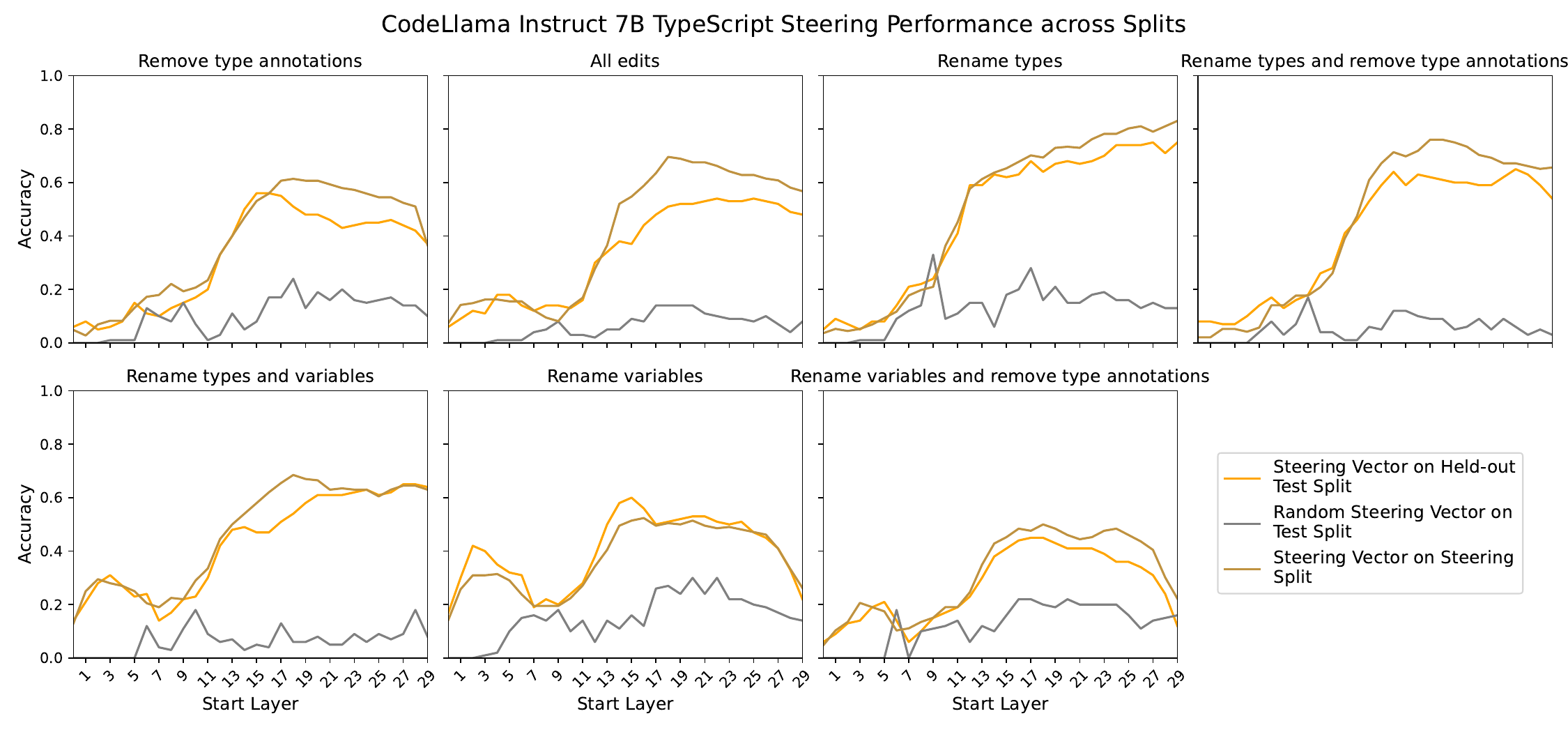}
    \caption{TypeScript steering performance for CodeLlama Instruct 7B on test and steering datasets, compared against a random steering vector baseline. We steer 3 adjacent layers.}
    \label{fig:splits-CodeLlama-7b-Instruct-hf-ts-3}
\end{figure*}

\begin{figure*}
    \centering
    \includegraphics[width=\textwidth]{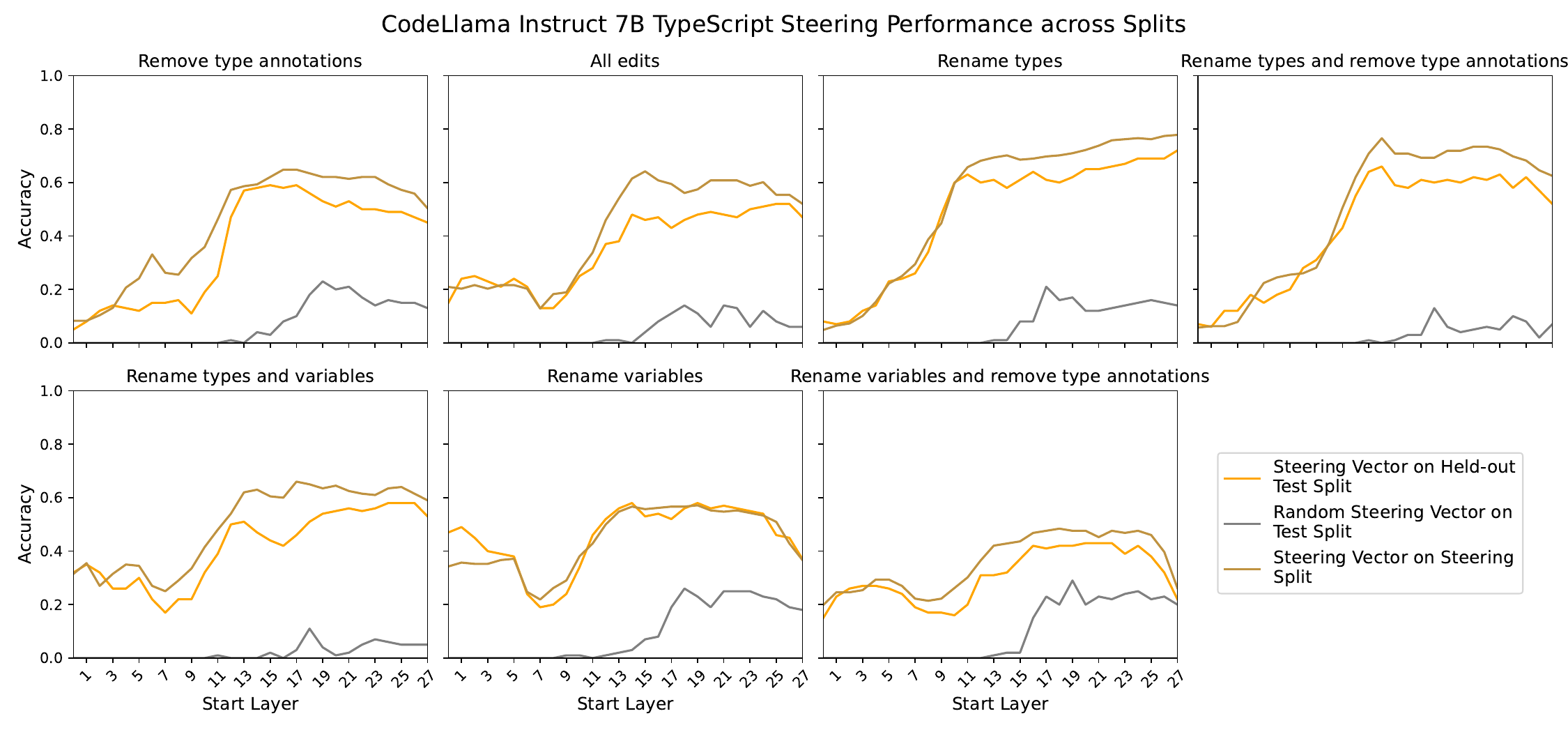}
    \caption{TypeScript steering performance for CodeLlama Instruct 7B on test and steering datasets, compared against a random steering vector baseline. We steer 5 adjacent layers.}
    \label{fig:splits-CodeLlama-7b-Instruct-hf-ts-5}
\end{figure*}

\begin{figure*}
    \centering
    \includegraphics[width=\textwidth]{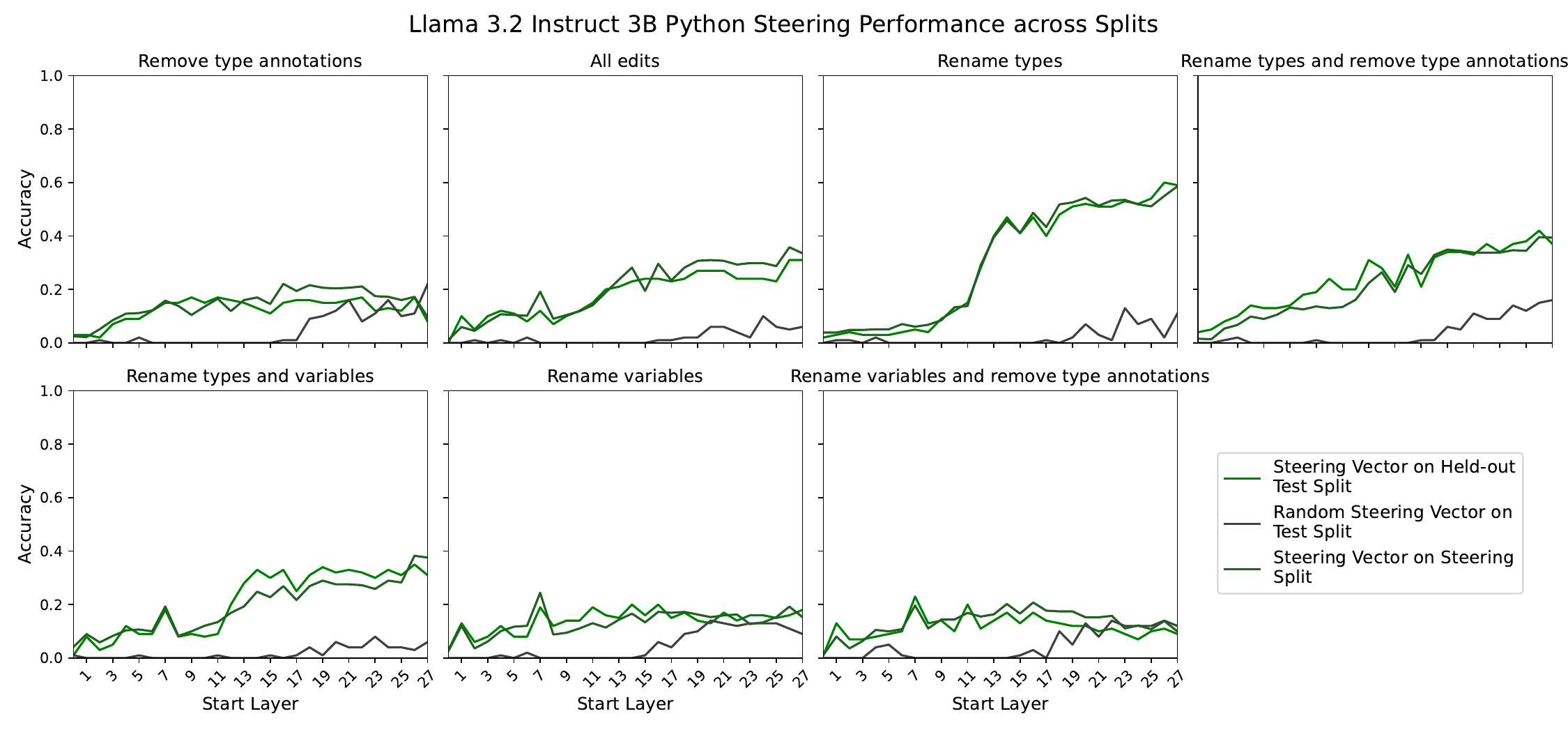}
    \caption{Python steering performance for Llama 3.2 Instruct 3B on test and steering datasets, compared against a random steering vector baseline. We steer 1 adjacent layers.}
    \label{fig:splits-Llama-3.2-3B-Instruct-py-1}
\end{figure*}

\begin{figure*}
    \centering
    \includegraphics[width=\textwidth]{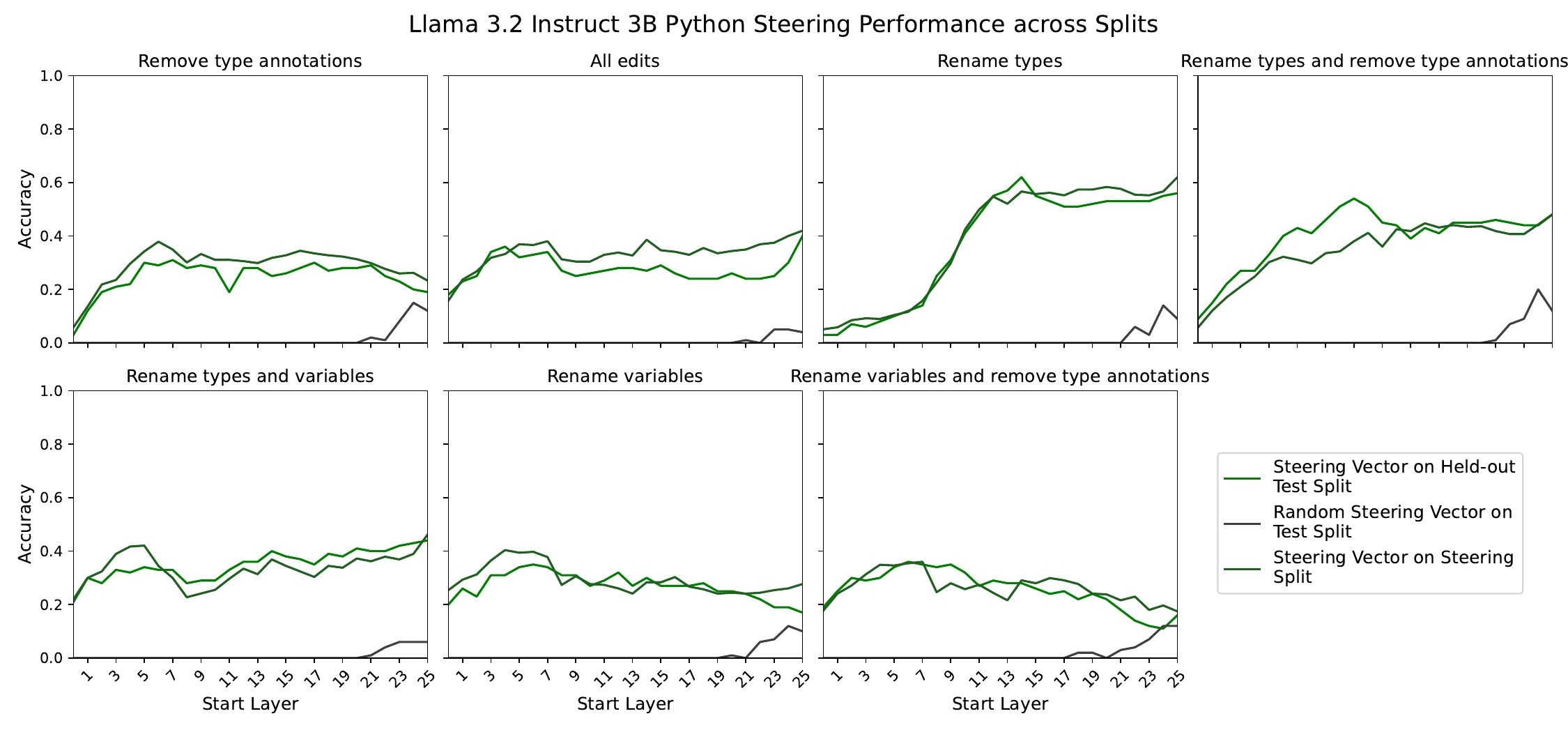}
    \caption{Python steering performance for Llama 3.2 Instruct 3B on test and steering datasets, compared against a random steering vector baseline. We steer 3 adjacent layers.}
    \label{fig:splits-Llama-3.2-3B-Instruct-py-3}
\end{figure*}

\begin{figure*}
    \centering
    \includegraphics[width=\textwidth]{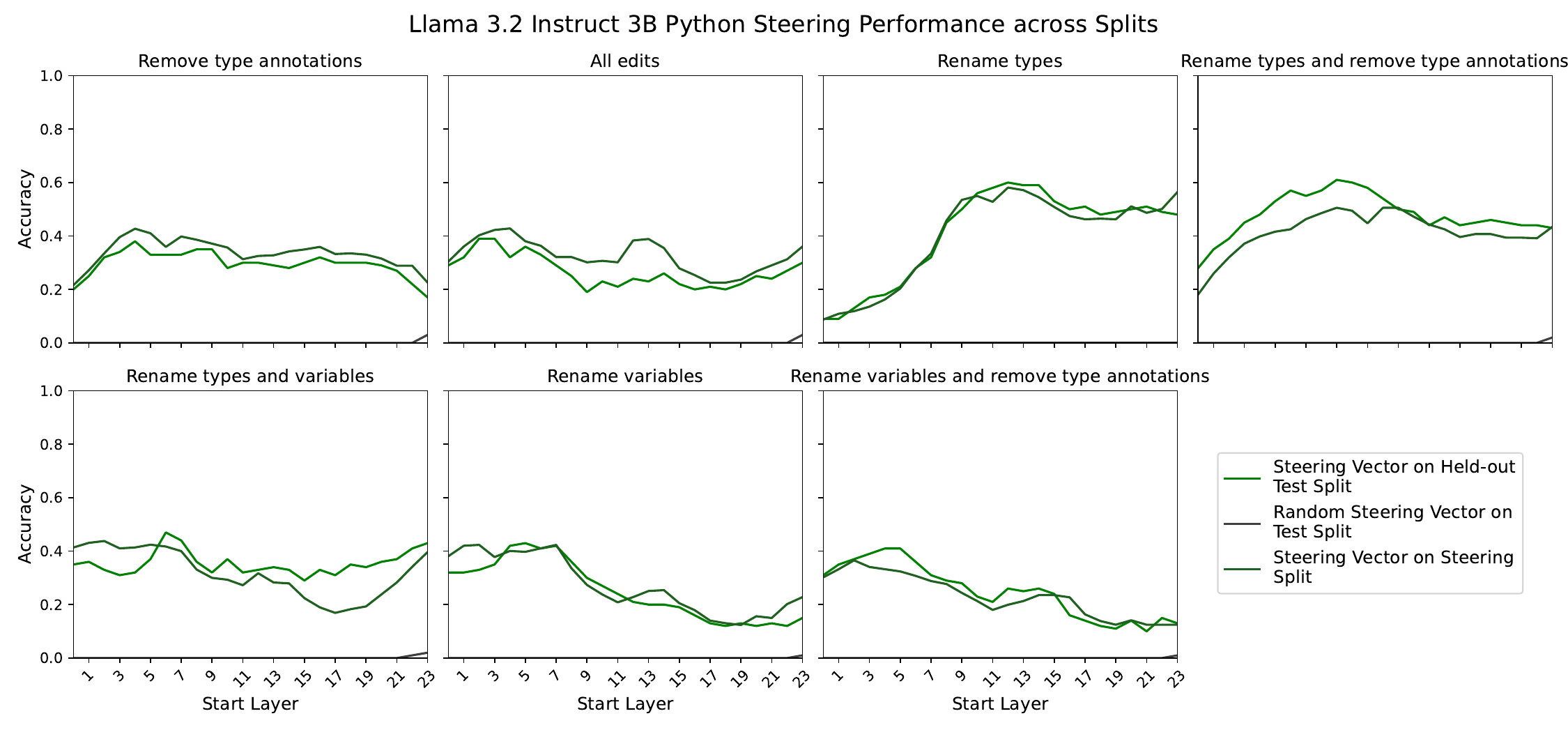}
    \caption{Python steering performance for Llama 3.2 Instruct 3B on test and steering datasets, compared against a random steering vector baseline. We steer 5 adjacent layers.}
    \label{fig:splits-Llama-3.2-3B-Instruct-py-5}
\end{figure*}

\begin{figure*}
    \centering
    \includegraphics[width=\textwidth]{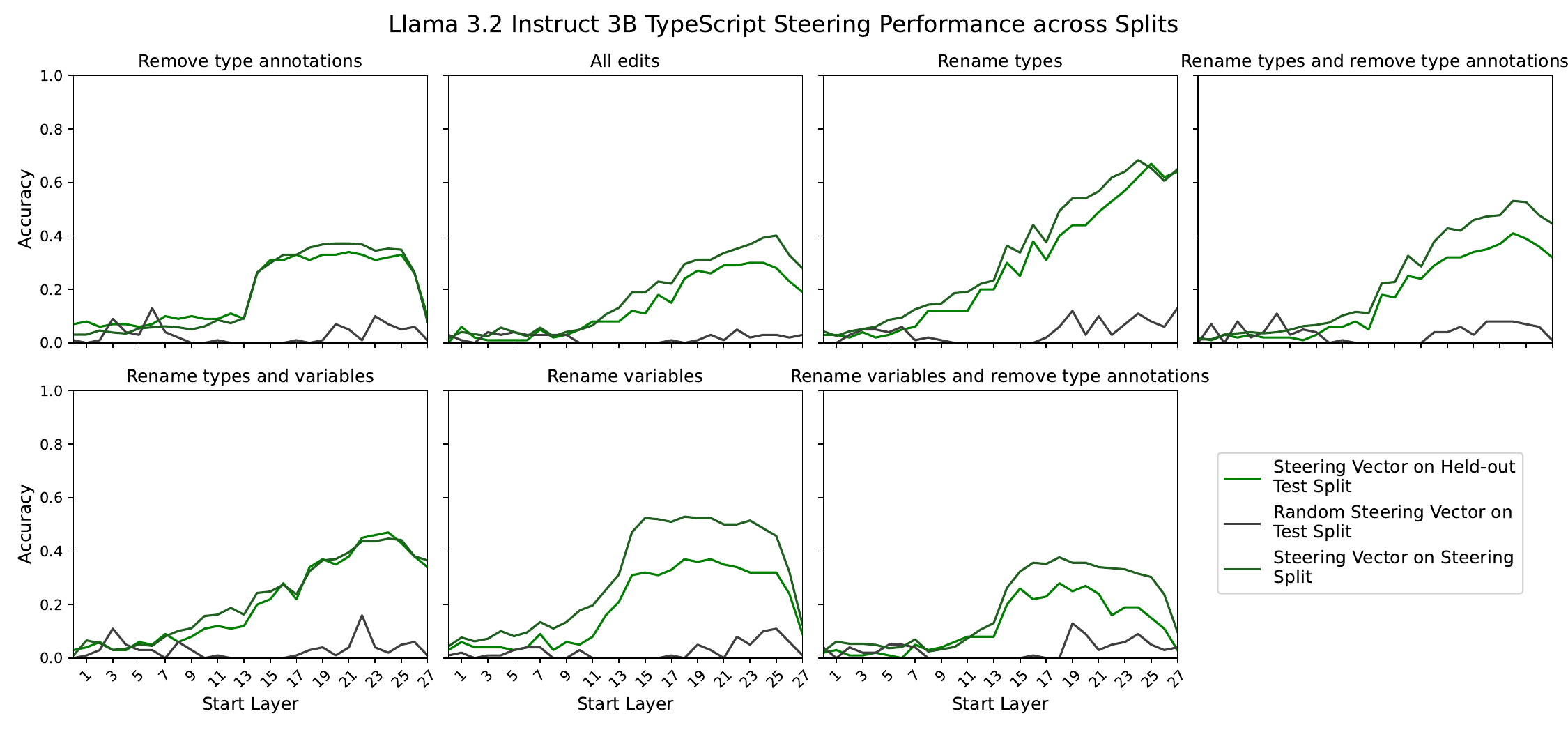}
    \caption{TypeScript steering performance for Llama 3.2 Instruct 3B on test and steering datasets, compared against a random steering vector baseline. We steer 1 adjacent layers.}
    \label{fig:splits-Llama-3.2-3B-Instruct-ts-1}
\end{figure*}

\begin{figure*}
    \centering
    \includegraphics[width=\textwidth]{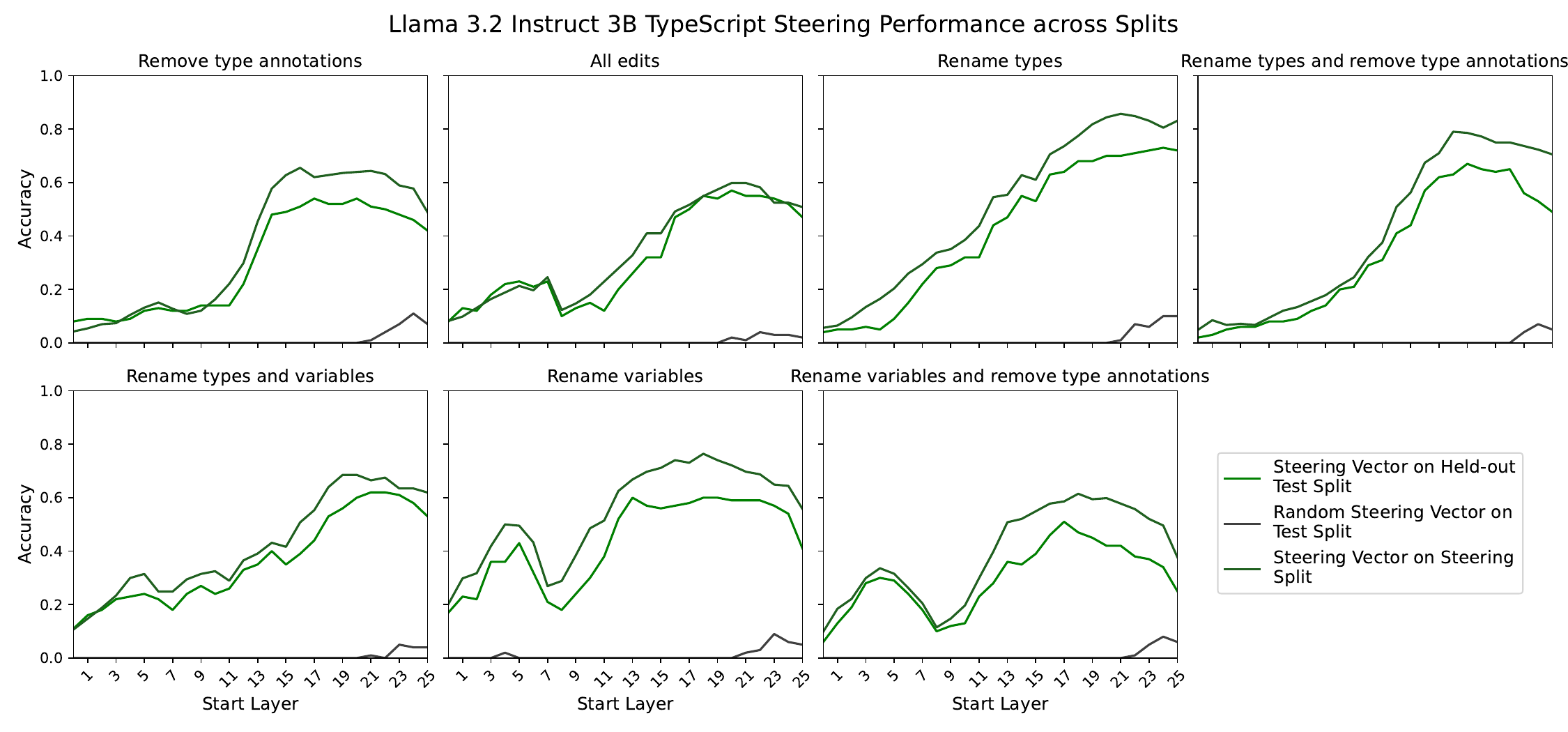}
    \caption{TypeScript steering performance for Llama 3.2 Instruct 3B on test and steering datasets, compared against a random steering vector baseline. We steer 3 adjacent layers.}
    \label{fig:splits-Llama-3.2-3B-Instruct-ts-3}
\end{figure*}

\begin{figure*}
    \centering
    \includegraphics[width=\textwidth]{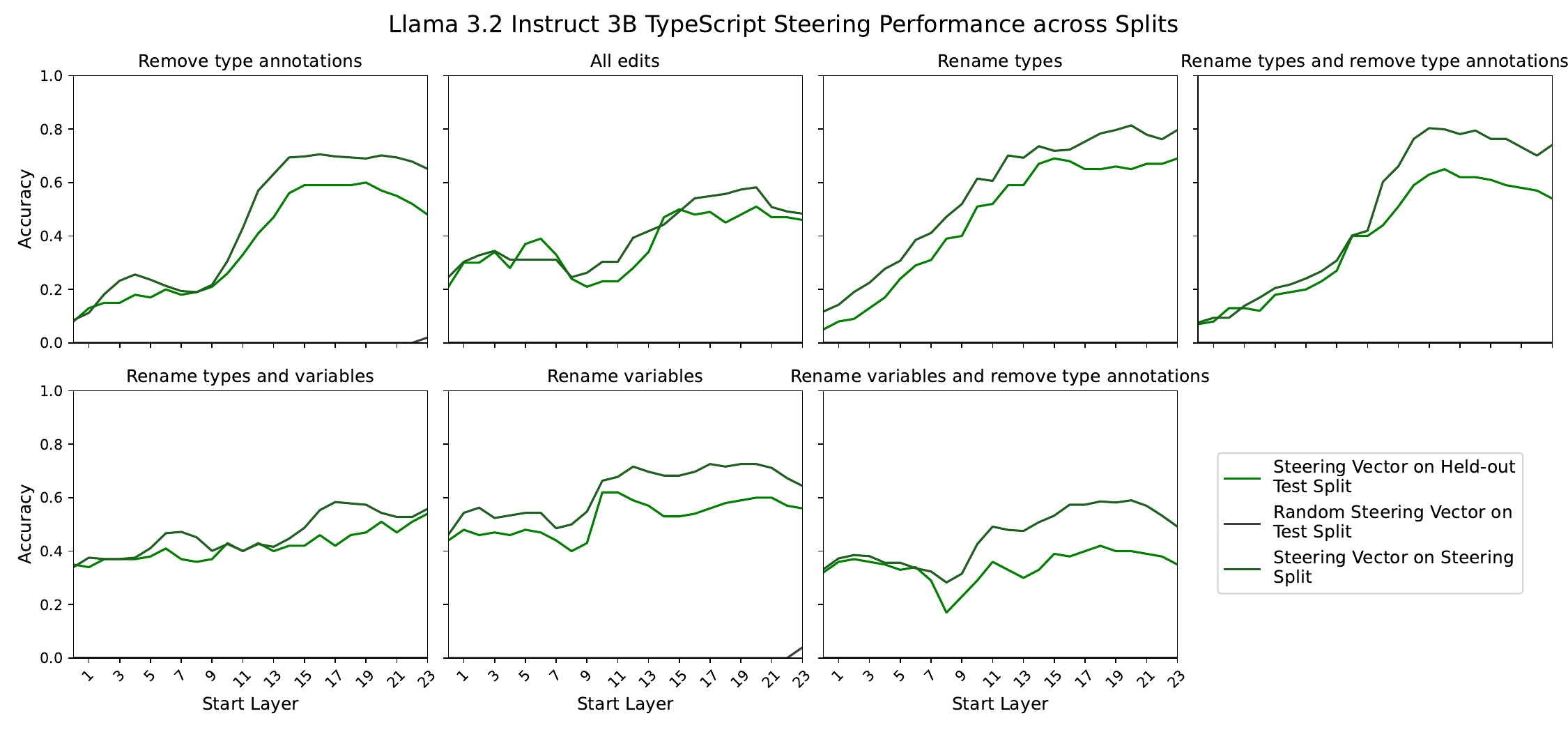}
    \caption{TypeScript steering performance for Llama 3.2 Instruct 3B on test and steering datasets, compared against a random steering vector baseline. We steer 5 adjacent layers.}
    \label{fig:splits-Llama-3.2-3B-Instruct-ts-5}
\end{figure*}

\begin{figure*}
    \centering
    \includegraphics[width=\textwidth]{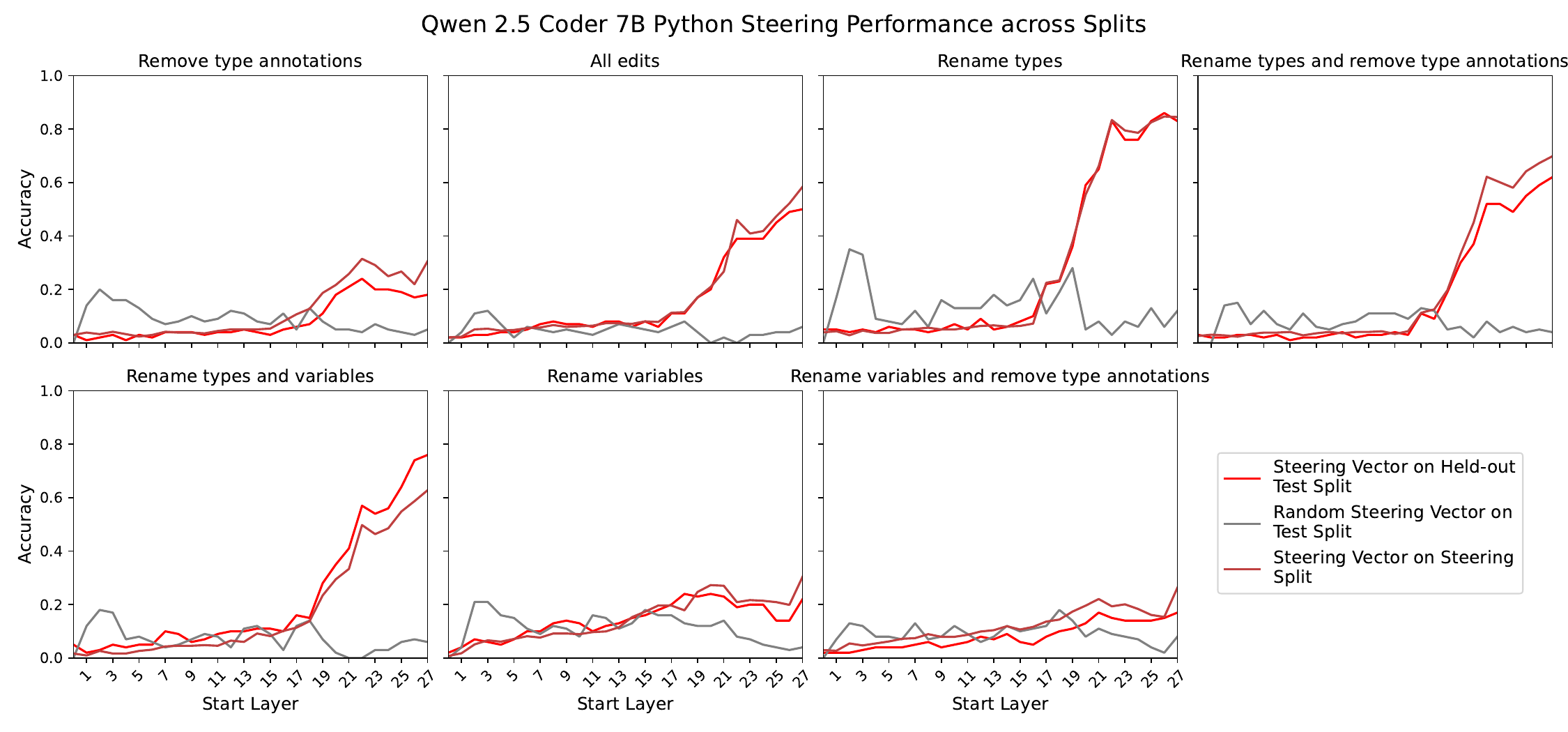}
    \caption{Python steering performance for Qwen 2.5 Coder 7B on test and steering datasets, compared against a random steering vector baseline. We steer 1 adjacent layers.}
    \label{fig:splits-qwen2p5_coder_7b_base-py-1}
\end{figure*}

\begin{figure*}
    \centering
    \includegraphics[width=\textwidth]{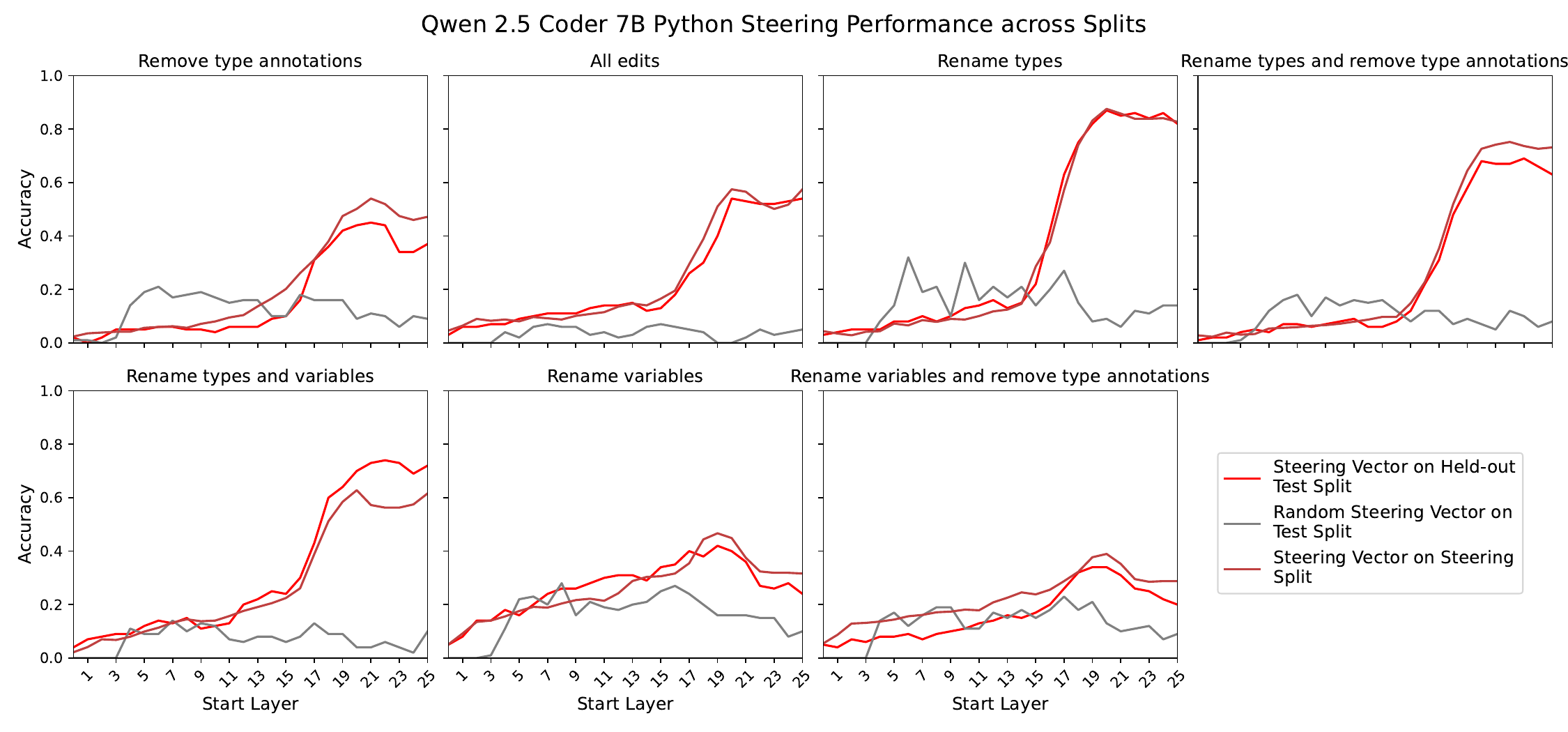}
    \caption{Python steering performance for Qwen 2.5 Coder 7B on test and steering datasets, compared against a random steering vector baseline. We steer 3 adjacent layers.}
    \label{fig:splits-qwen2p5_coder_7b_base-py-3}
\end{figure*}

\begin{figure*}
    \centering
    \includegraphics[width=\textwidth]{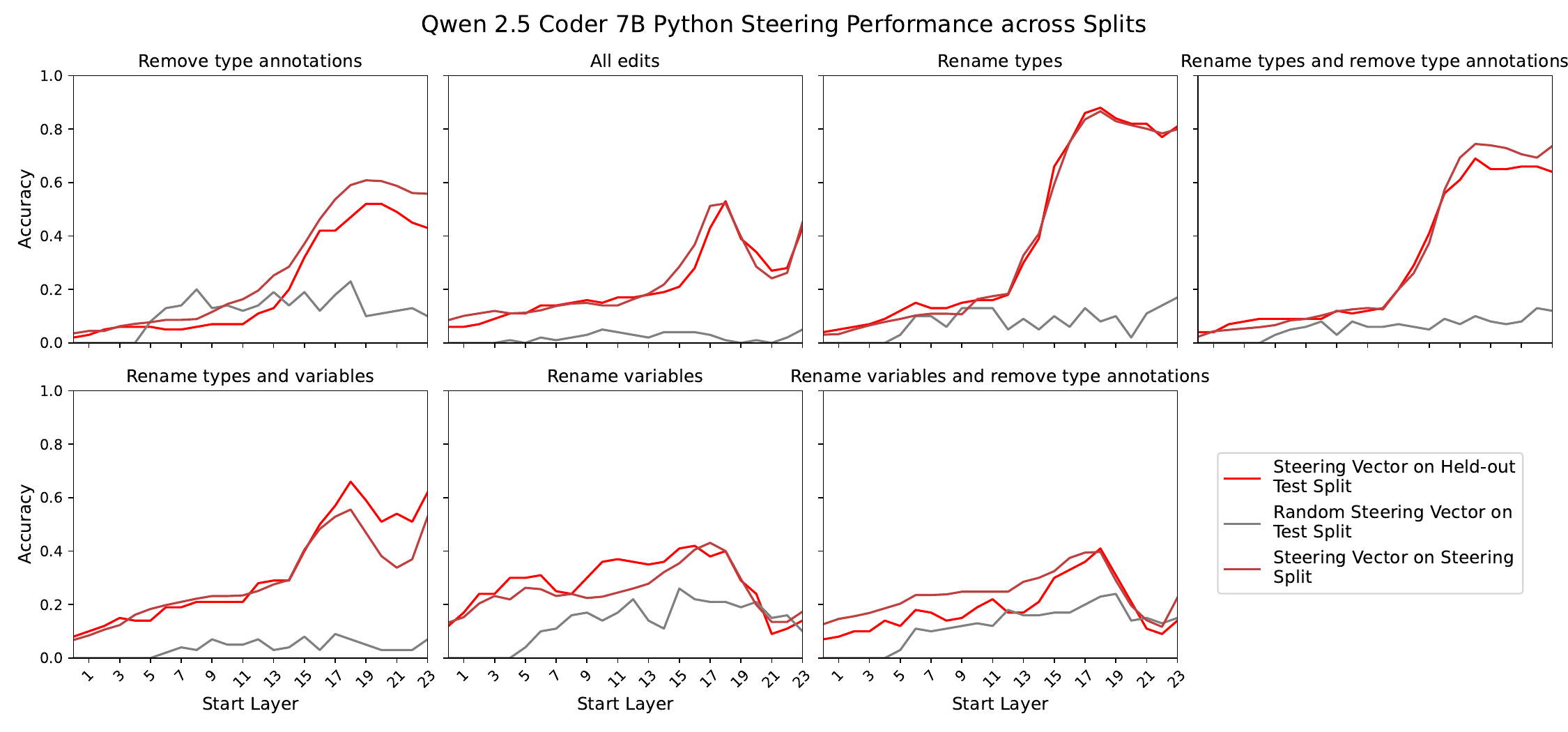}
    \caption{Python steering performance for Qwen 2.5 Coder 7B on test and steering datasets, compared against a random steering vector baseline. We steer 5 adjacent layers.}
    \label{fig:splits-qwen2p5_coder_7b_base-py-5}
\end{figure*}

\begin{figure*}
    \centering
    \includegraphics[width=\textwidth]{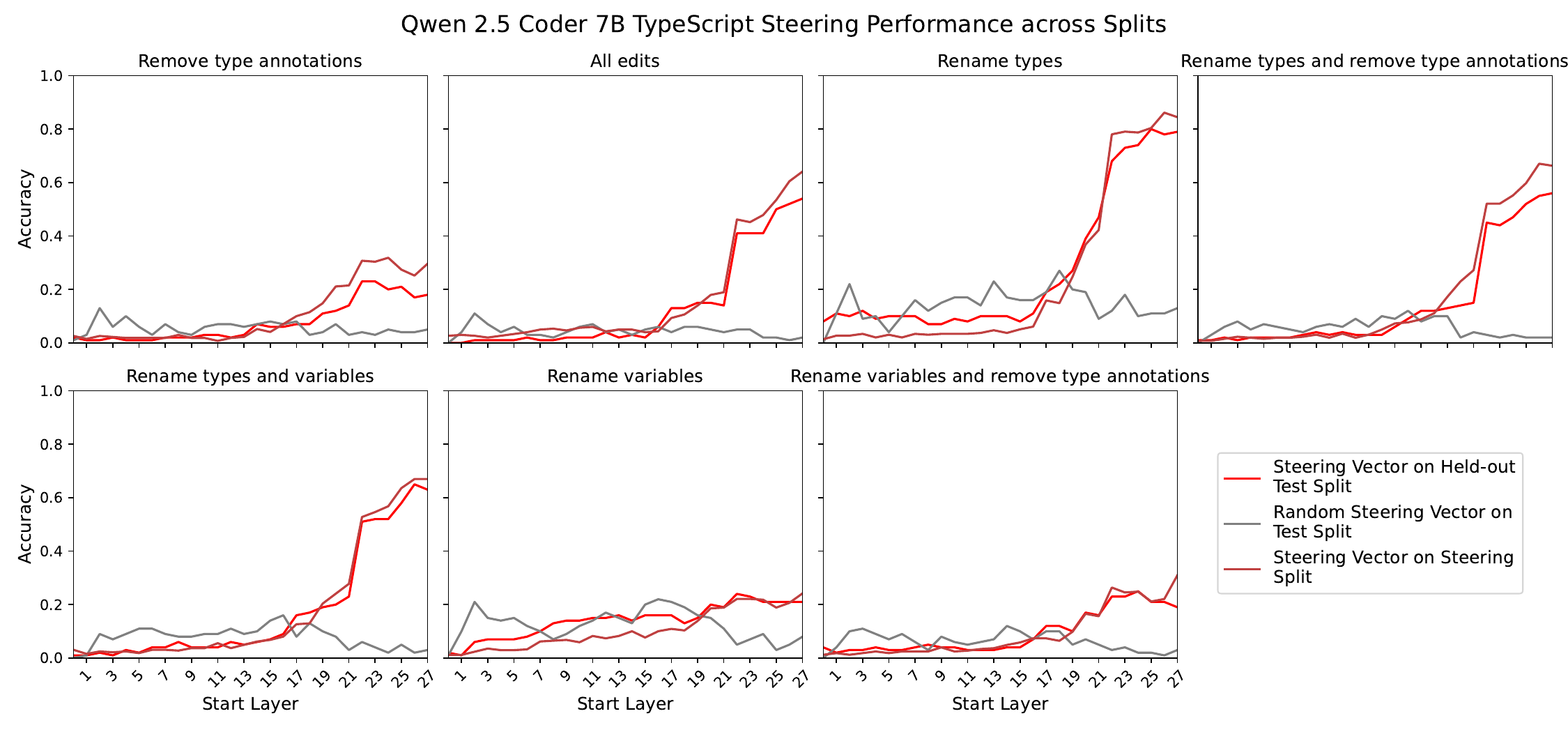}
    \caption{TypeScript steering performance for Qwen 2.5 Coder 7B on test and steering datasets, compared against a random steering vector baseline. We steer 1 adjacent layers.}
    \label{fig:splits-qwen2p5_coder_7b_base-ts-1}
\end{figure*}

\begin{figure*}
    \centering
    \includegraphics[width=\textwidth]{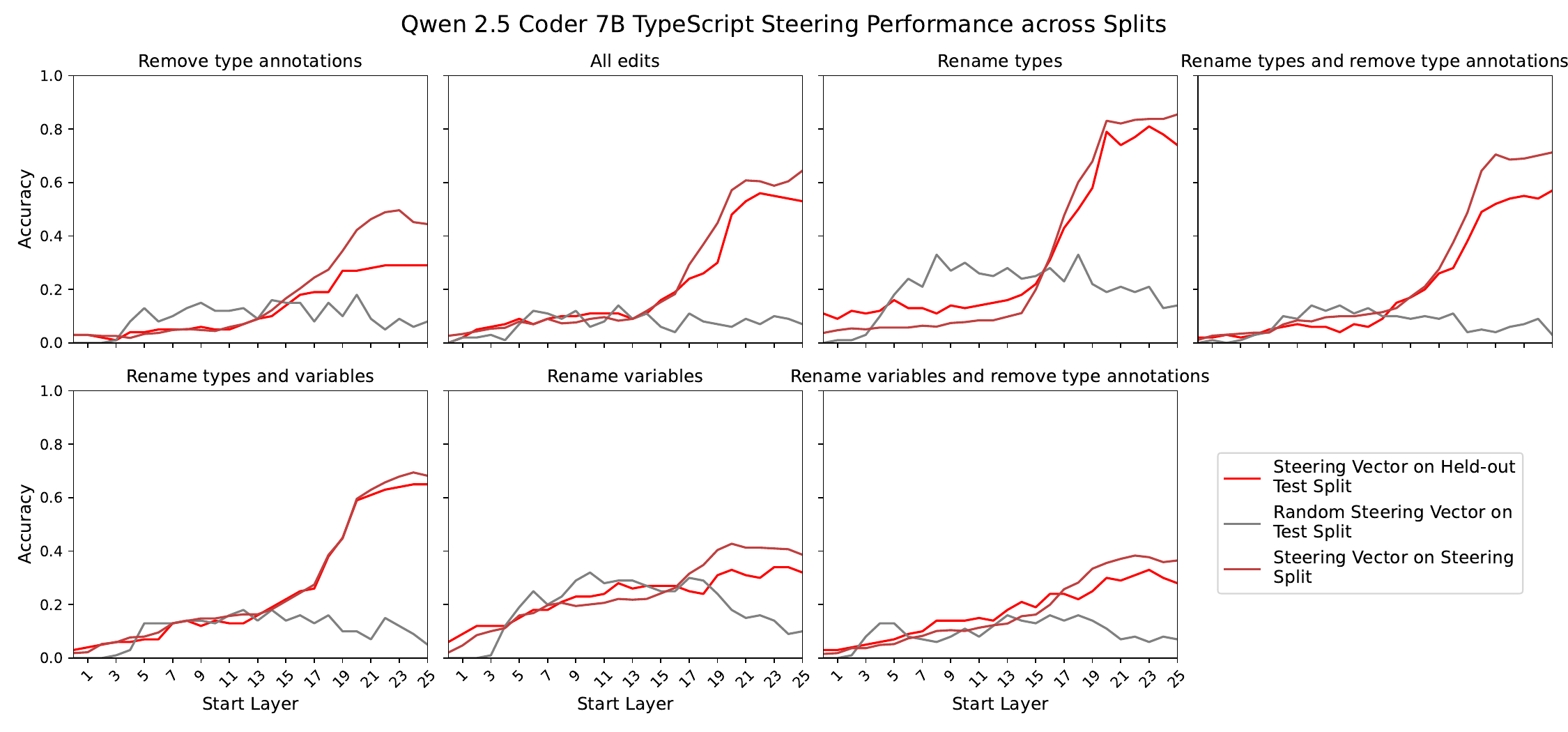}
    \caption{TypeScript steering performance for Qwen 2.5 Coder 7B on test and steering datasets, compared against a random steering vector baseline. We steer 3 adjacent layers.}
    \label{fig:splits-qwen2p5_coder_7b_base-ts-3}
\end{figure*}

\begin{figure*}
    \centering
    \includegraphics[width=\textwidth]{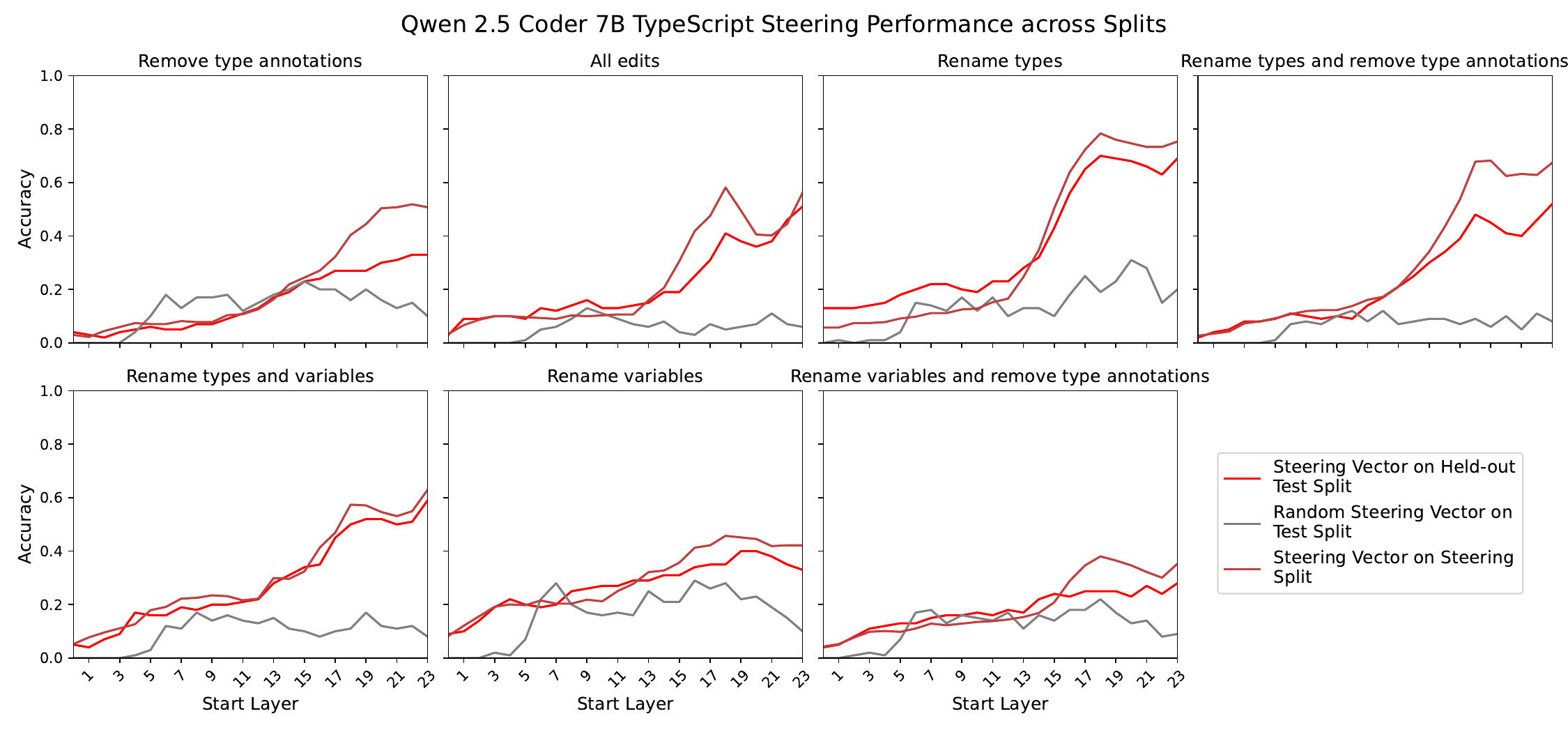}
    \caption{TypeScript steering performance for Qwen 2.5 Coder 7B on test and steering datasets, compared against a random steering vector baseline. We steer 5 adjacent layers.}
    \label{fig:splits-qwen2p5_coder_7b_base-ts-5}
\end{figure*}

\begin{figure*}
    \centering
    \includegraphics[width=\textwidth]{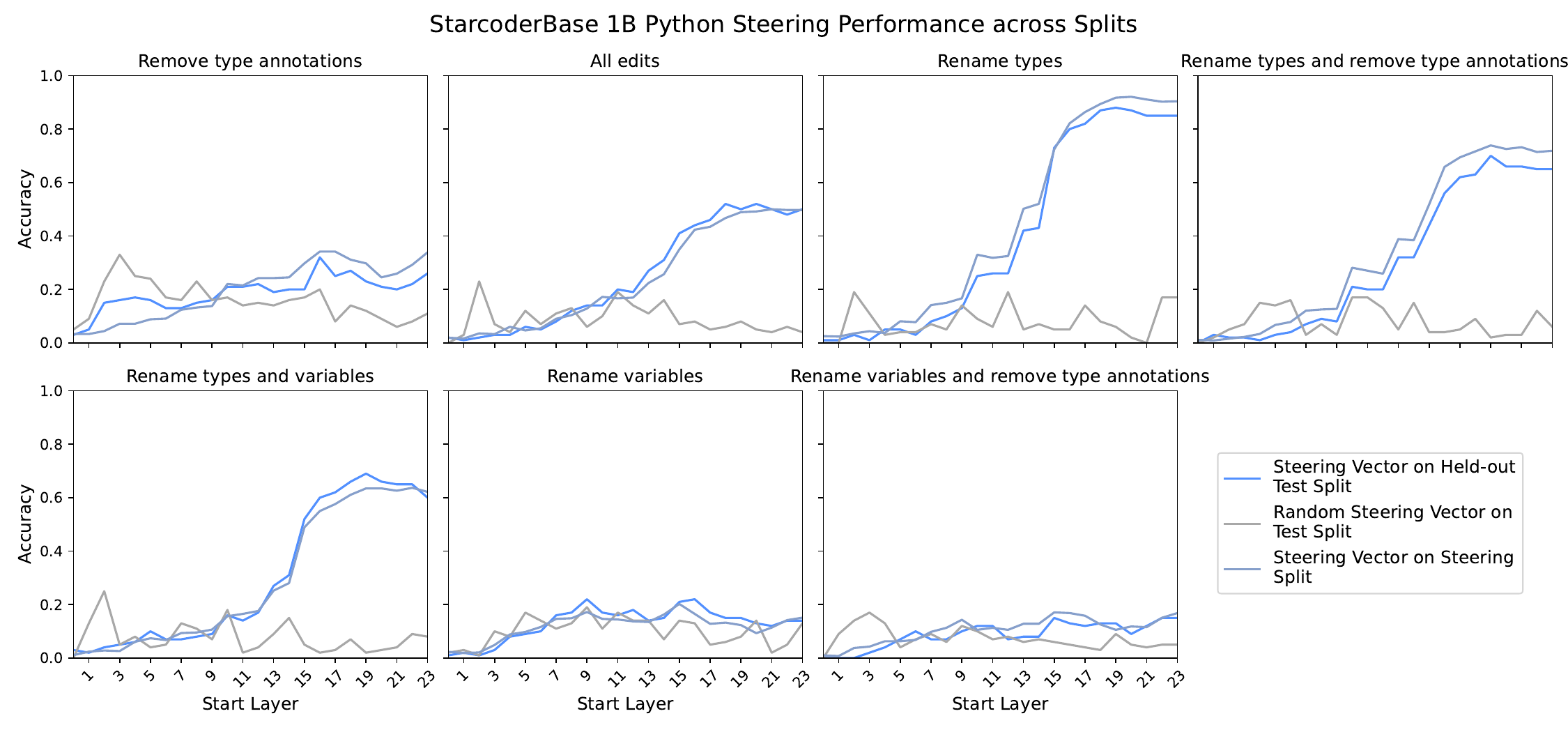}
    \caption{Python steering performance for StarcoderBase 1B on test and steering datasets, compared against a random steering vector baseline. We steer 1 adjacent layers.}
    \label{fig:splits-starcoderbase-1b-py-1}
\end{figure*}

\begin{figure*}
    \centering
    \includegraphics[width=\textwidth]{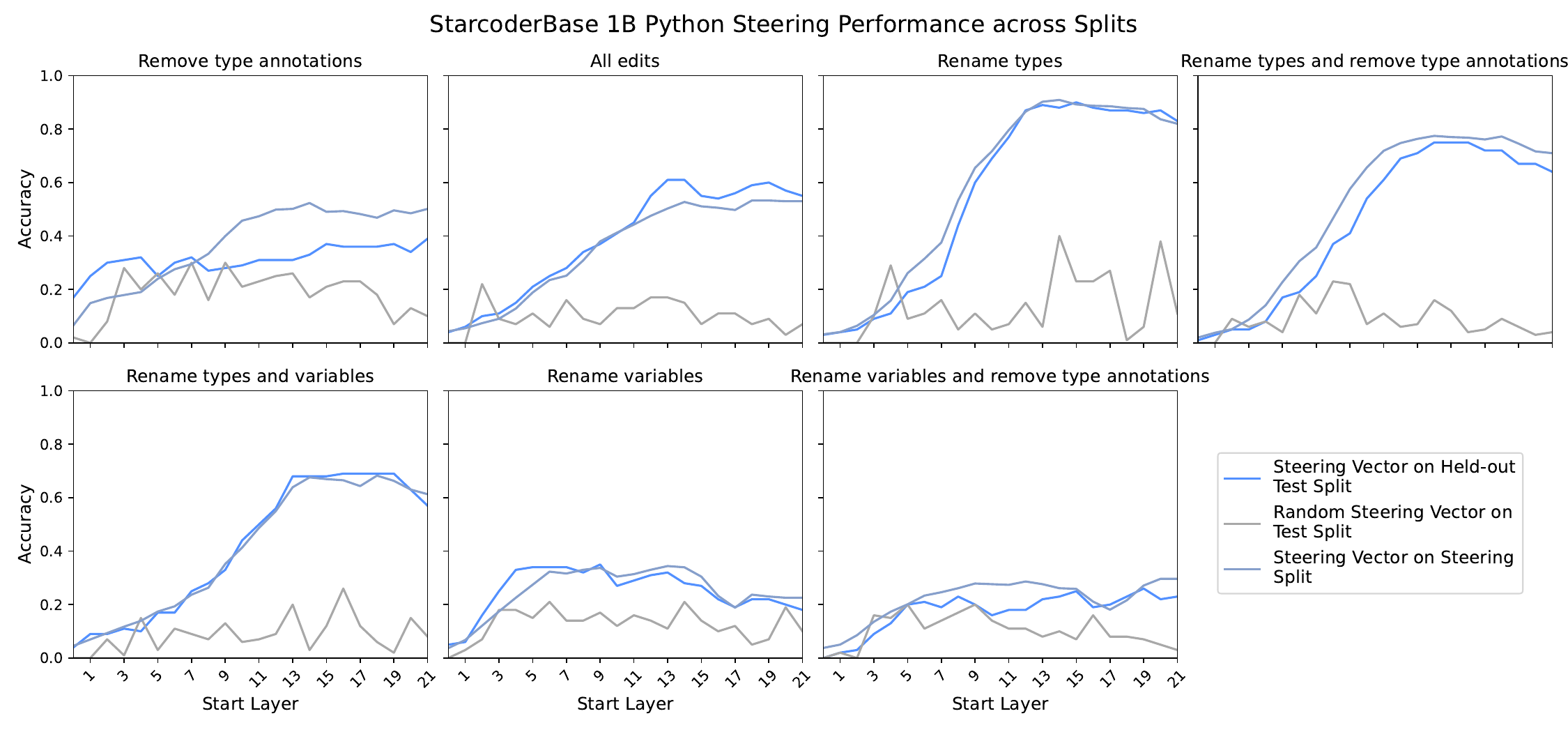}
    \caption{Python steering performance for StarcoderBase 1B on test and steering datasets, compared against a random steering vector baseline. We steer 3 adjacent layers.}
    \label{fig:splits-starcoderbase-1b-py-3}
\end{figure*}

\begin{figure*}
    \centering
    \includegraphics[width=\textwidth]{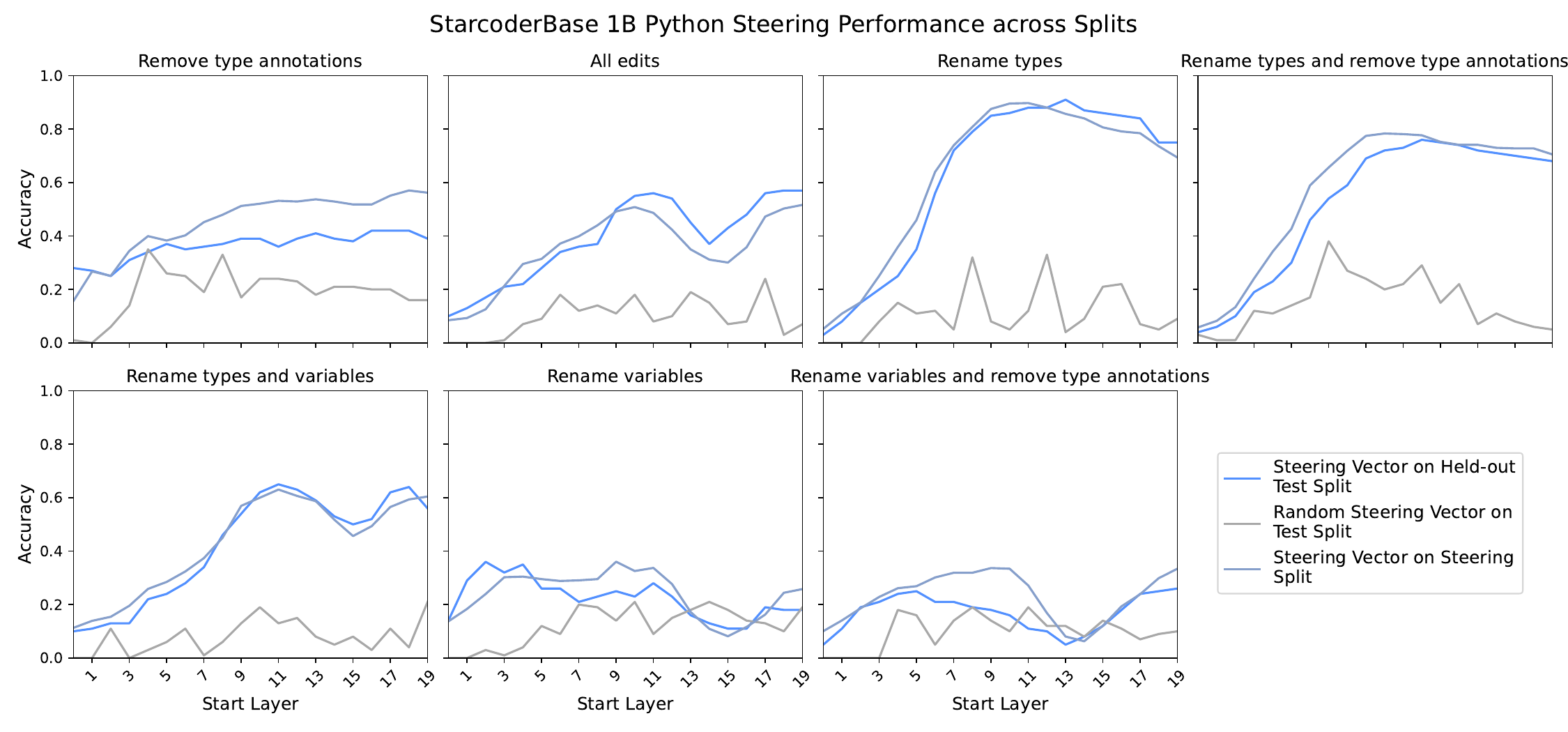}
    \caption{Python steering performance for StarcoderBase 1B on test and steering datasets, compared against a random steering vector baseline. We steer 5 adjacent layers.}
    \label{fig:splits-starcoderbase-1b-py-5}
\end{figure*}

\begin{figure*}
    \centering
    \includegraphics[width=\textwidth]{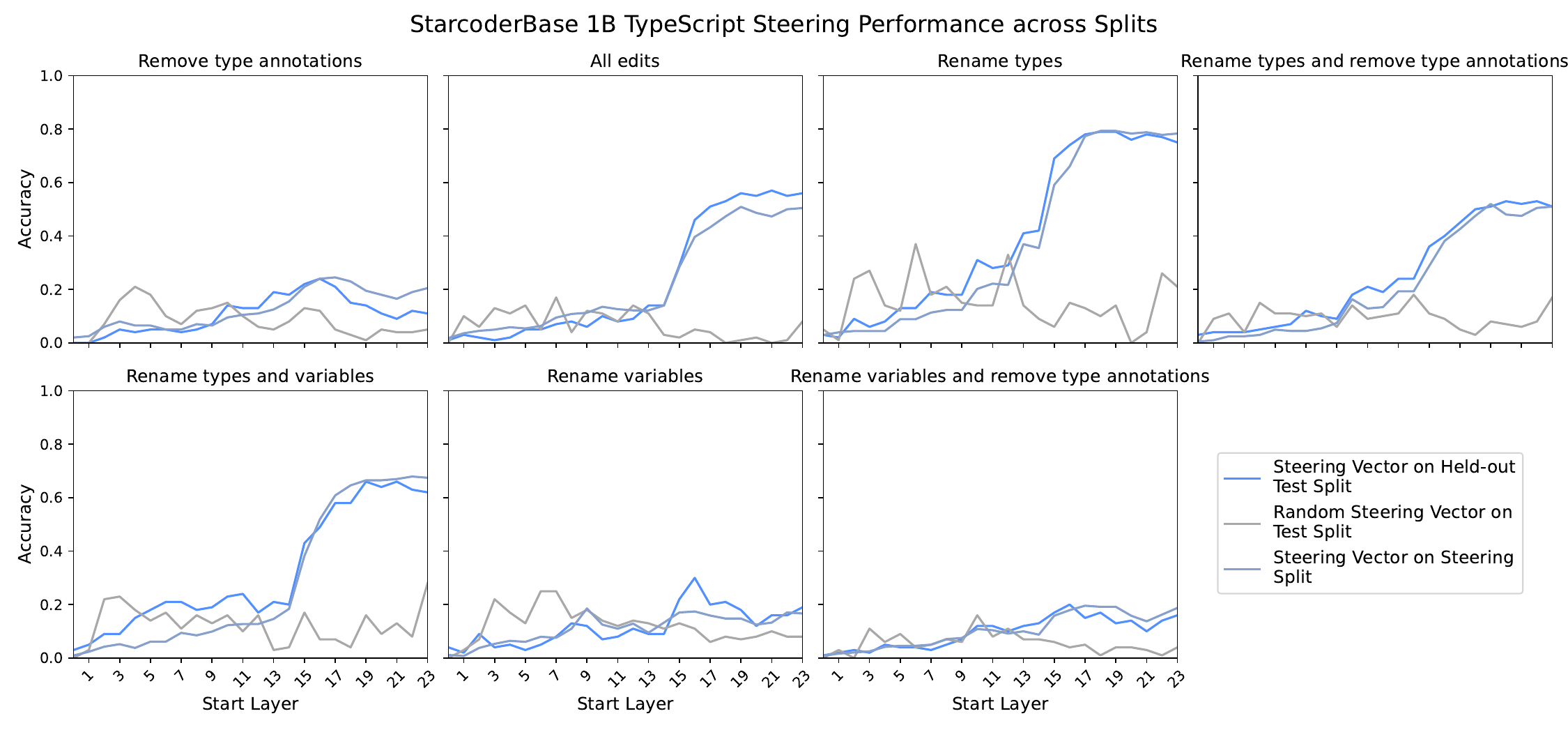}
    \caption{TypeScript steering performance for StarcoderBase 1B on test and steering datasets, compared against a random steering vector baseline. We steer 1 adjacent layers.}
    \label{fig:splits-starcoderbase-1b-ts-1}
\end{figure*}

\begin{figure*}
    \centering
    \includegraphics[width=\textwidth]{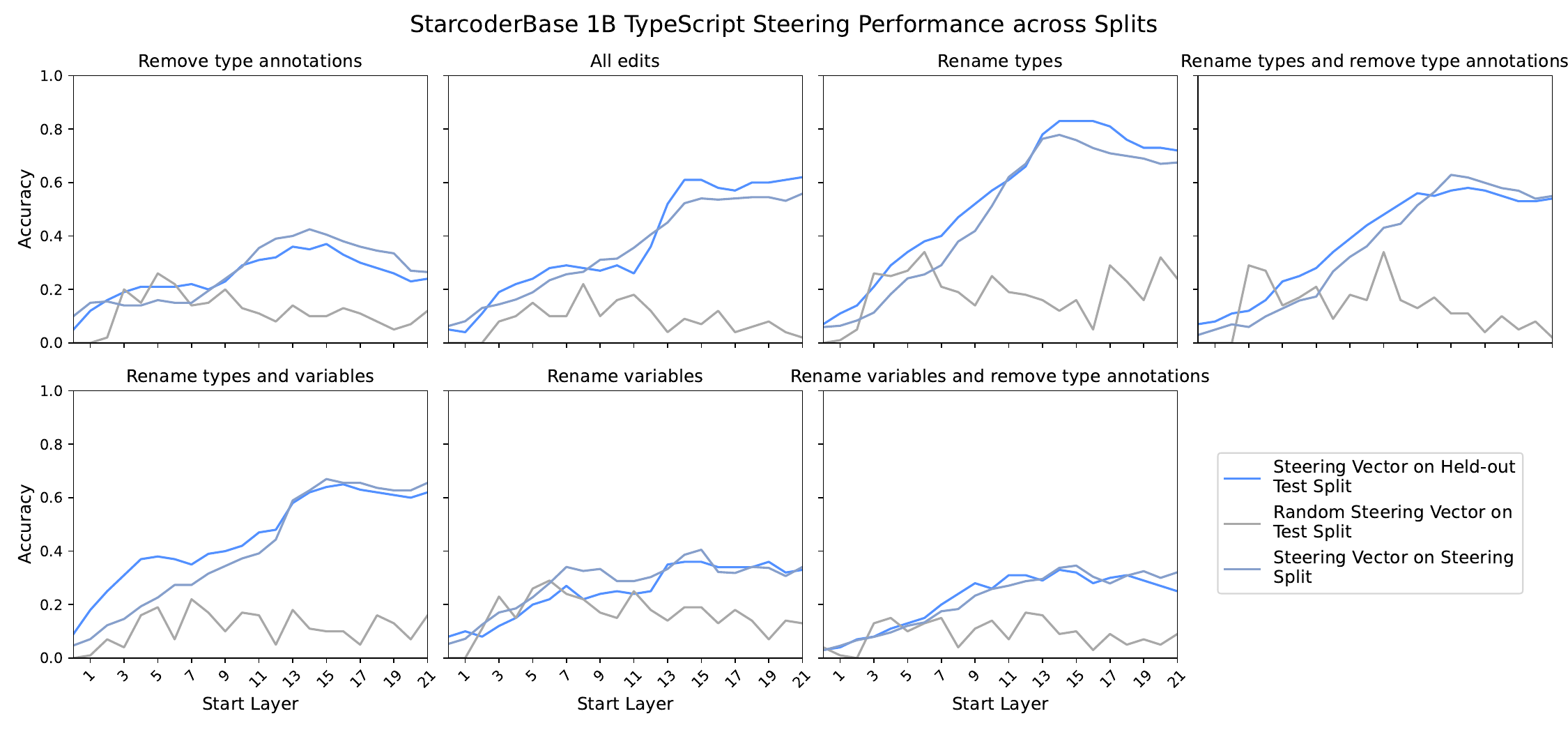}
    \caption{TypeScript steering performance for StarcoderBase 1B on test and steering datasets, compared against a random steering vector baseline. We steer 3 adjacent layers.}
    \label{fig:splits-starcoderbase-1b-ts-3}
\end{figure*}

\begin{figure*}
    \centering
    \includegraphics[width=\textwidth]{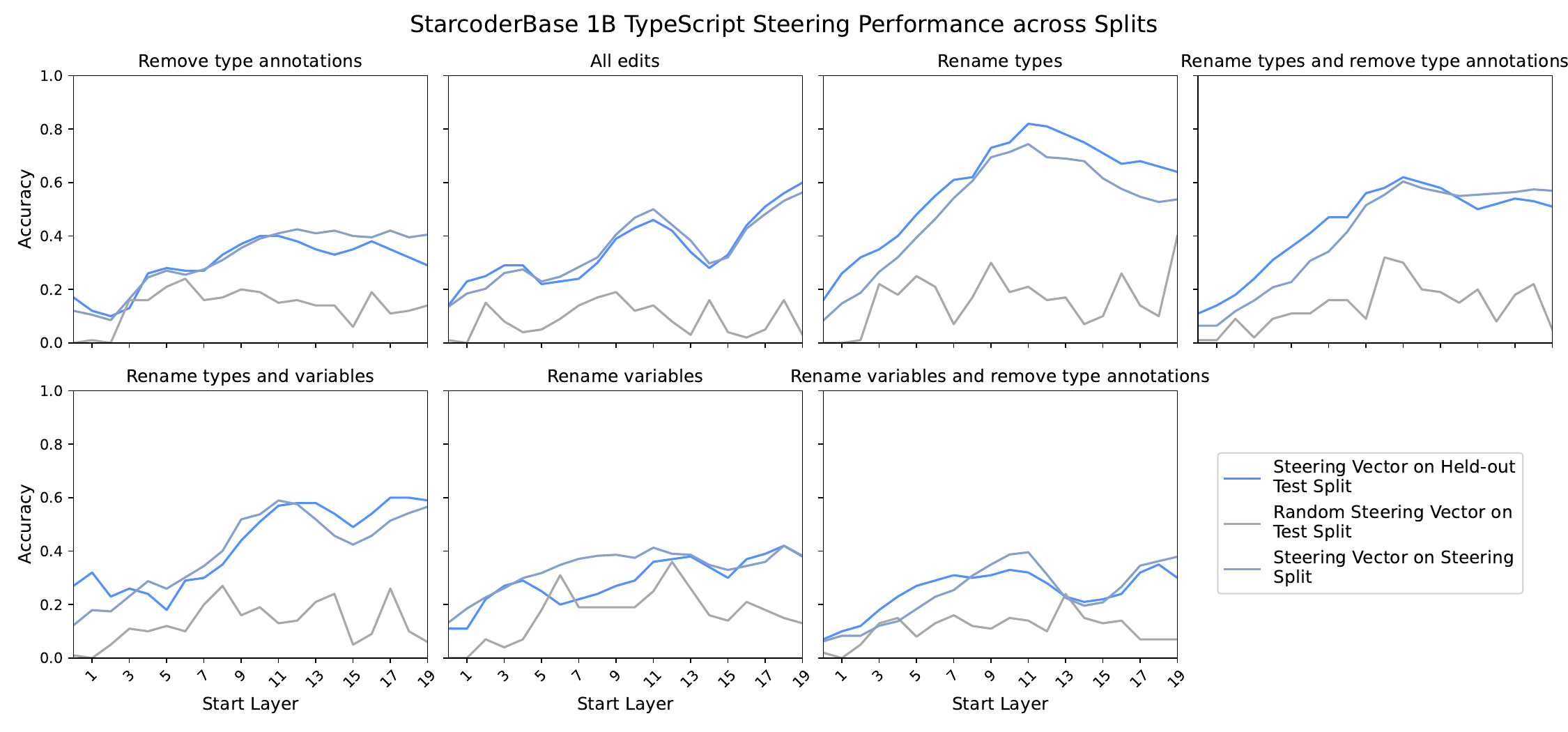}
    \caption{TypeScript steering performance for StarcoderBase 1B on test and steering datasets, compared against a random steering vector baseline. We steer 5 adjacent layers.}
    \label{fig:splits-starcoderbase-1b-ts-5}
\end{figure*}

\begin{figure*}
    \centering
    \includegraphics[width=\textwidth]{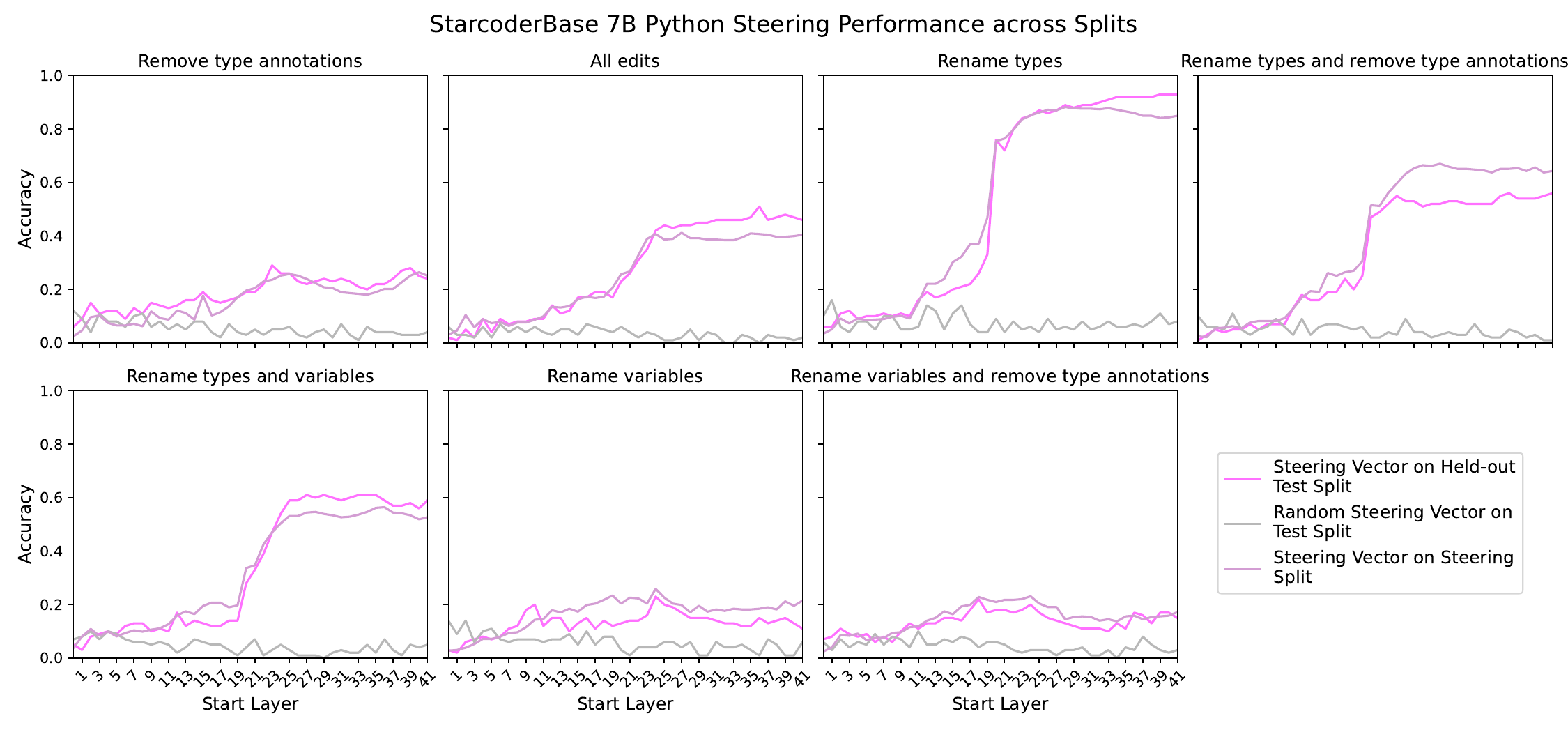}
    \caption{Python steering performance for StarcoderBase 7B on test and steering datasets, compared against a random steering vector baseline. We steer 1 adjacent layers.}
    \label{fig:splits-starcoderbase-7b-py-1}
\end{figure*}

\begin{figure*}
    \centering
    \includegraphics[width=\textwidth]{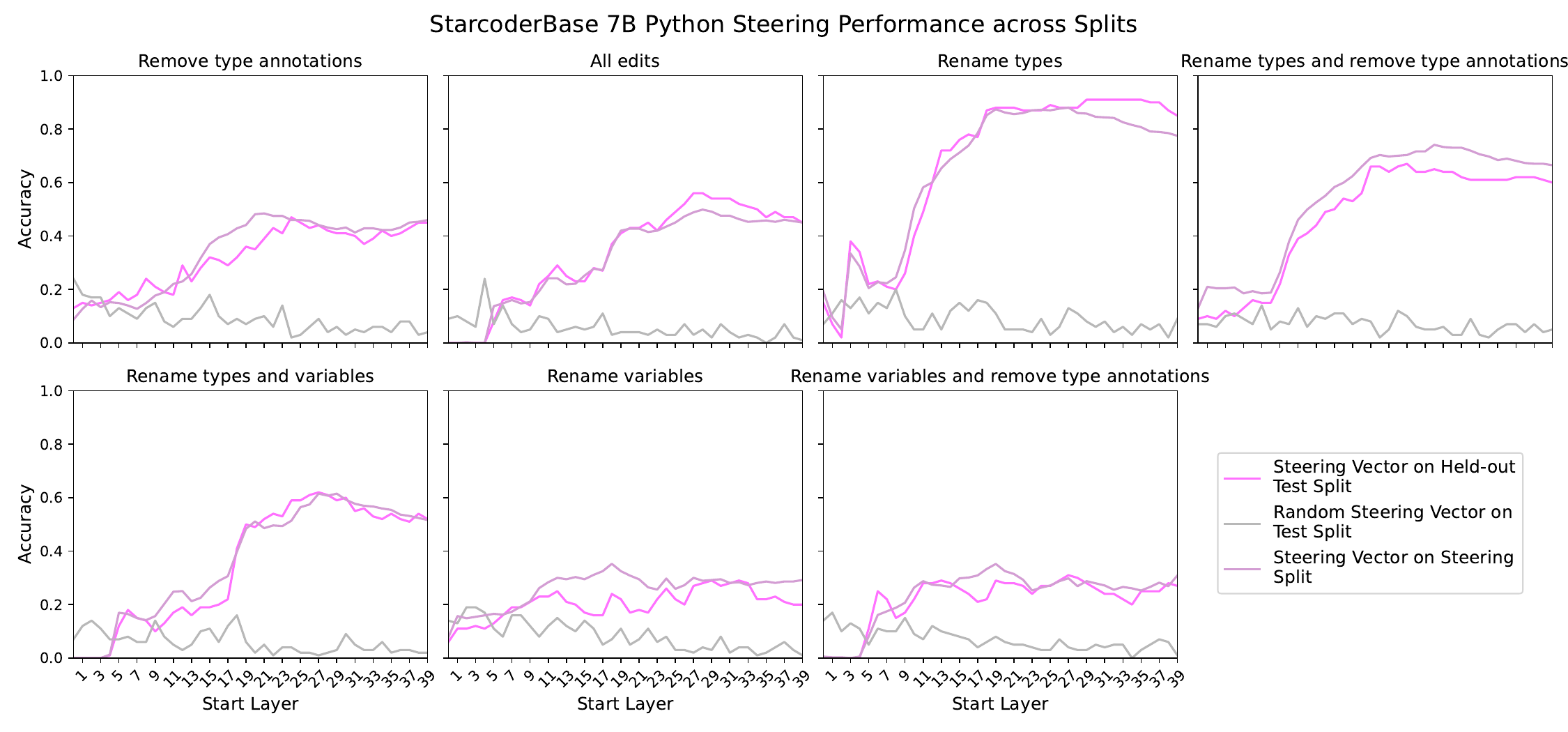}
    \caption{Python steering performance for StarcoderBase 7B on test and steering datasets, compared against a random steering vector baseline. We steer 3 adjacent layers.}
    \label{fig:splits-starcoderbase-7b-py-3}
\end{figure*}

\begin{figure*}
    \centering
    \includegraphics[width=\textwidth]{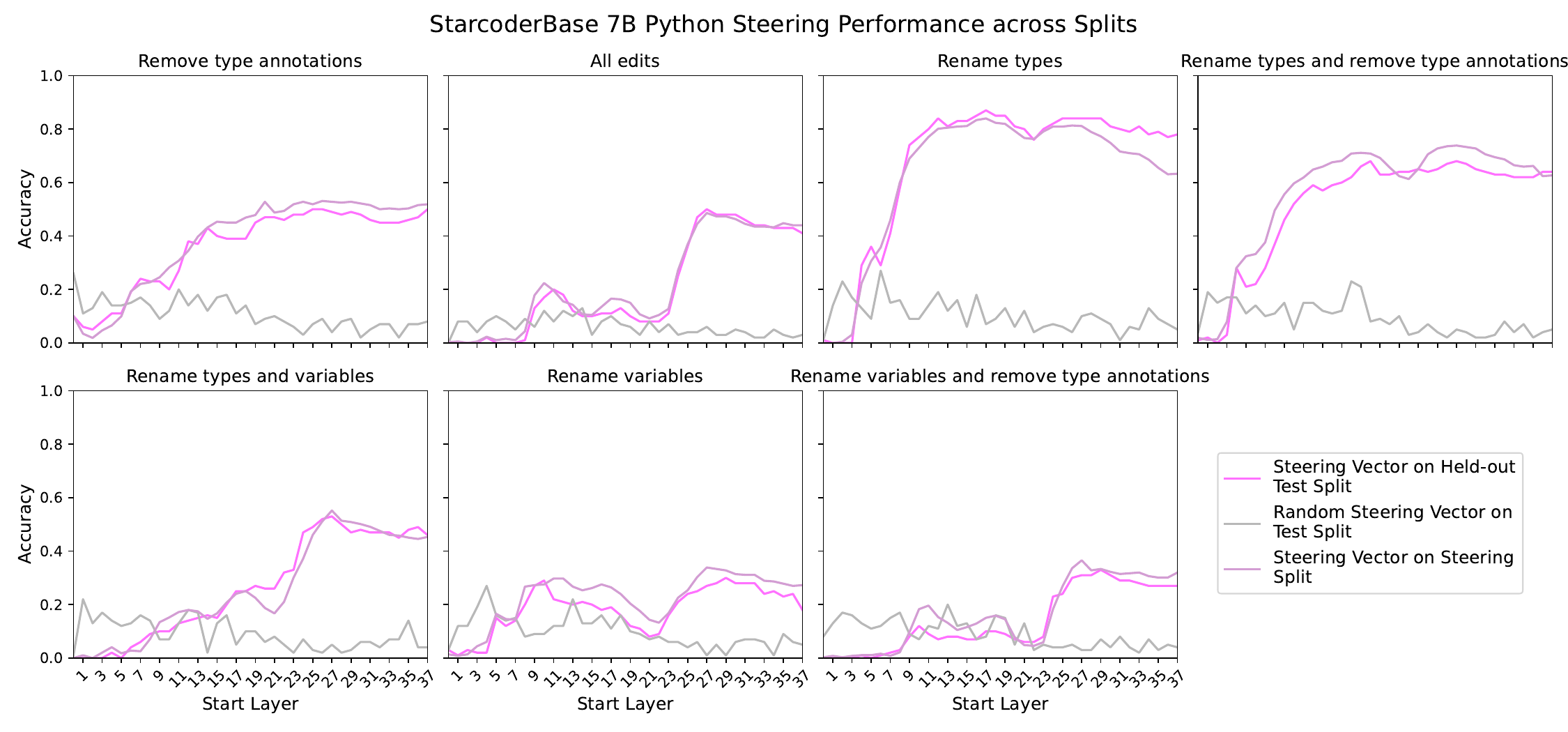}
    \caption{Python steering performance for StarcoderBase 7B on test and steering datasets, compared against a random steering vector baseline. We steer 5 adjacent layers.}
    \label{fig:splits-starcoderbase-7b-py-5}
\end{figure*}

\begin{figure*}
    \centering
    \includegraphics[width=\textwidth]{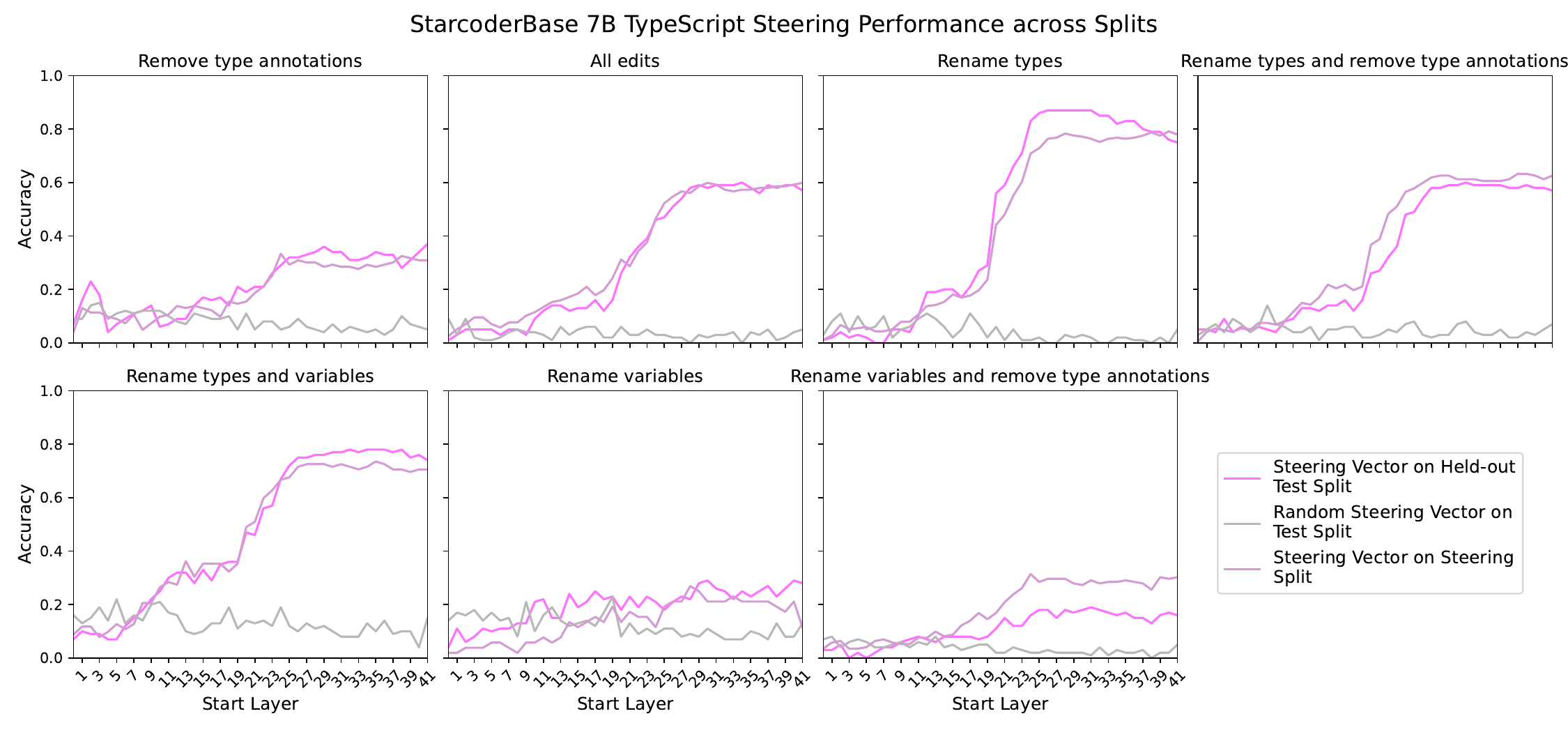}
    \caption{TypeScript steering performance for StarcoderBase 7B on test and steering datasets, compared against a random steering vector baseline. We steer 1 adjacent layers.}
    \label{fig:splits-starcoderbase-7b-ts-1}
\end{figure*}

\begin{figure*}
    \centering
    \includegraphics[width=\textwidth]{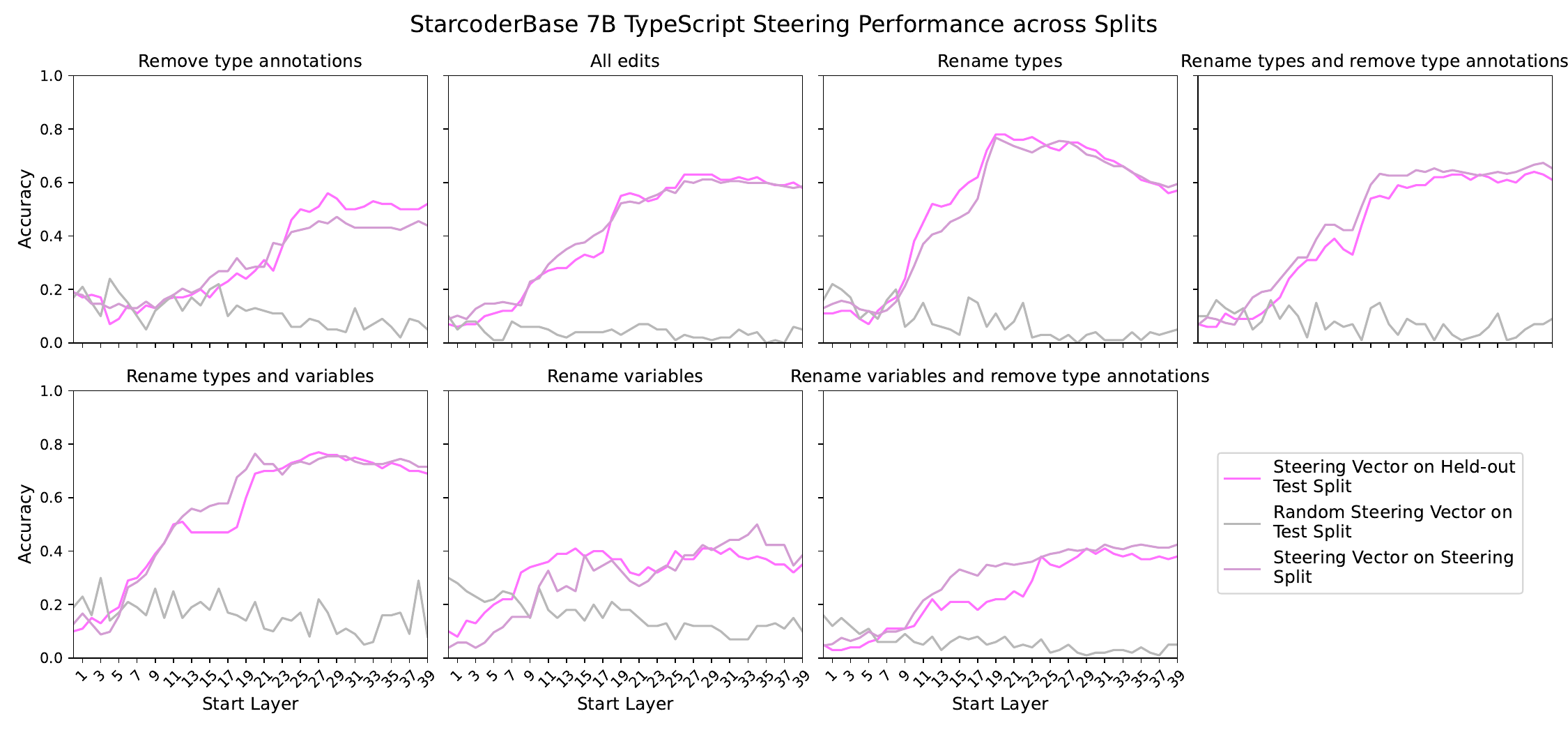}
    \caption{TypeScript steering performance for StarcoderBase 7B on test and steering datasets, compared against a random steering vector baseline. We steer 3 adjacent layers.}
    \label{fig:splits-starcoderbase-7b-ts-3}
\end{figure*}

\begin{figure*}
    \centering
    \includegraphics[width=\textwidth]{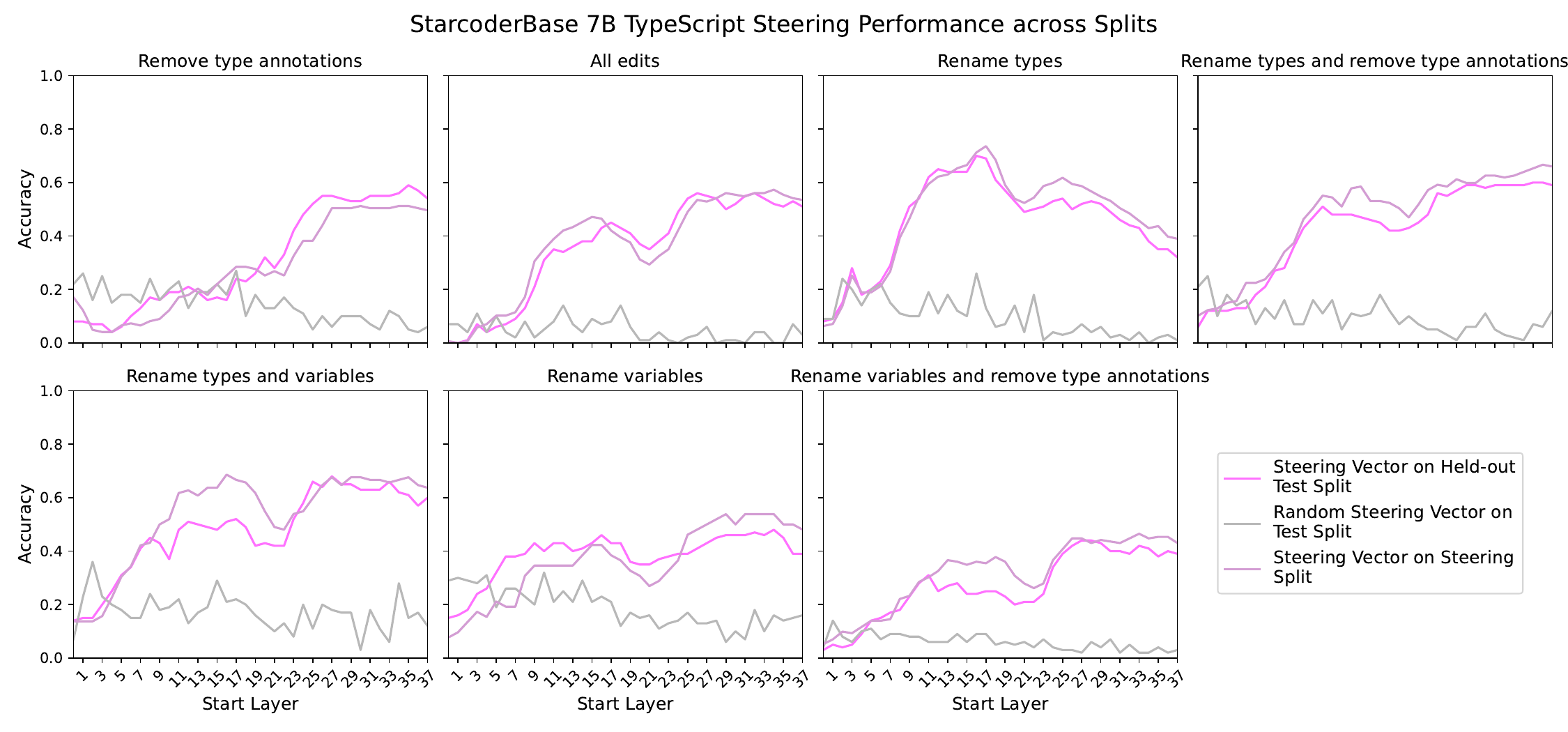}
    \caption{TypeScript steering performance for StarcoderBase 7B on test and steering datasets, compared against a random steering vector baseline. We steer 5 adjacent layers.}
    \label{fig:splits-starcoderbase-7b-ts-5}
\end{figure*}



\end{document}